\documentclass[12pt]{article}
\usepackage[nonatbib, final]{neurips_2020}

\usepackage[numbers]{natbib}
\usepackage[utf8]{inputenc} 
\usepackage[T1]{fontenc}    
\usepackage[colorlinks=true,citecolor=blue,urlcolor=blue]{hyperref}       
\usepackage{url}            
\usepackage{booktabs}       
\usepackage{amsfonts}       
\usepackage{nicefrac}       
\usepackage{microtype}      
\usepackage[bottom]{footmisc}  

\usepackage{amsmath}
\usepackage{amssymb}
\usepackage{amsthm}
\usepackage{graphicx}
\usepackage{diagbox}
\usepackage{mathtools}
\usepackage{bbm}
\usepackage[skins]{tcolorbox}
\usepackage{wrapfig}
\usepackage{arydshln}
\usepackage[font=small,labelfont=bf]{caption}
\usepackage{tabularx}

\usepackage{color, colortbl}

\newcommand{\norm}[1]{\left\|#1\right\|}

\newcommand{\inner}[1]{\left\langle#1\right\rangle}
\newcommand*{\defeq}{\stackrel{\text{def}}{=}}

\definecolor{lightgrey}{rgb}{0.9, 0.9, 0.9}

\newcommand{\myparagraph}{\textbf}

\newcommand{\Id}{\mathbbm{1}}
\def\argmax{\mathop{\rm arg\,max}}

\def\minop{\mathop{\rm min}\limits}
\def\maxop{\mathop{\rm max}\limits}
\def\sign{\mathop{\rm sign}\limits}
\def\R{\mathbb{R}}
\def\U{\mathcal{U}}

\definecolor{myorange}{RGB}{232,209,82}
\tcbset{commonstyle/.style={boxrule=0pt,sharp corners,enhanced jigsaw,nobeforeafter,boxsep=0pt,left=\fboxsep,right=\fboxsep}}
\newtcolorbox{mycolorbox}[1][]{commonstyle,#1}

\newtheorem{lemma}{Lemma}

\newtheorem{innercustomlemma}{Lemma}
\newenvironment{customlemma}[1]{\renewcommand\theinnercustomlemma{#1}\innercustomlemma}{\endinnercustomlemma}

\newcolumntype{C}[1]{>{\centering\arraybackslash}p{#1}}

\DeclareMathOperator{\E}{\mathbb{E}}
\let\l\relax
\DeclareMathOperator{\l}{\mathbf{\ell}}
\let\vec\relax
\DeclareMathOperator{\vec}{vec}

\graphicspath{ {images/} }

\title{Understanding and Improving \\ Fast Adversarial Training}

\author{
	Maksym Andriushchenko \\
	EPFL, Theory of Machine Learning Lab\\
	\texttt{maksym.andriushchenko@epfl.ch} \\
	\And
	Nicolas Flammarion \\
	EPFL, Theory of Machine Learning Lab \\
	\texttt{nicolas.flammarion@epfl.ch}
}

\date{\vspace{-5ex}}

\begin{document}
	
	\maketitle

	\begin{abstract}
		A recent line of work focused on making adversarial training computationally efficient for deep learning models. In particular, \citet{wong2020fast} showed that $\ell_\infty$-adversarial training with fast gradient sign method (FGSM) can fail due to a phenomenon called \textit{catastrophic overfitting}, when the model quickly loses its robustness over a single epoch of training. We show that adding a random step to FGSM, as proposed in \cite{wong2020fast}, does not prevent catastrophic overfitting, and that randomness is not important per se --- its main role being simply to reduce the magnitude of the perturbation. 
		Moreover, we show that catastrophic overfitting is not inherent to deep and overparametrized networks, but can occur in a single-layer convolutional network with a few filters. In an extreme case, even \textit{a single filter} can make the network highly non-linear \textit{locally}, which is the main reason why FGSM training fails. 
		Based on this observation, we propose a new regularization method, \texttt{GradAlign}, that \textit{prevents catastrophic overfitting} by explicitly maximizing the gradient alignment inside the perturbation set and improves the quality of the FGSM solution. As a result, \texttt{GradAlign} allows to successfully apply FGSM training also for larger $\ell_\infty$-perturbations and reduce the gap to multi-step adversarial training. The code of our experiments is available at \url{https://github.com/tml-epfl/understanding-fast-adv-training}.
	\end{abstract}

	\section{Introduction}
	Machine learning models based on empirical risk minimization are known to be often non-robust to small worst-case perturbations. 
	For decades, this has been the topic of active research by the statistics, optimization and machine learning communities~\cite{MR606374,MR2546839,globerson2006nightmare,biggio2018wild}. However, the recent success of deep learning~\cite{lecun2015deep,schmidhuber2015deep} has raised
	the interest in this topic.
	The lack of robustness in deep learning is clearly illustrated by the existence of \textit{adversarial examples}, i.e. tiny input perturbations that can easily fool state-of-the-art deep neural networks into making wrong predictions~\cite{szegedy2013intriguing, goodfellow2014explaining}. 
	
	Improving the robustness of machine learning models is motivated not only from the security perspective~\cite{biggio2018wild}. Adversarially robust models have better interpretability properties \cite{tsipras2018robustness,santurkar2019image} and can generalize better \cite{zhu2019freelb,bochkovskiy2020yolov4} including also improved performance under some distribution shifts \cite{xie2019adversarial} (although on some performing worse, see \cite{taori2020robustness}).
	In order to improve the robustness, two families of solutions have been developed: \emph{adversarial training} (AT) that amounts to training the model on adversarial examples~\cite{goodfellow2014explaining, madry2018towards} and \emph{provable defenses} that derive and optimize robustness certificates~\cite{wong2017provable,raghunathan2018certified,cohen2019certified}. Currently, adversarial-training based methods appear to be preferred by practitioners since they (a) achieve higher empirical robustness (although without providing a robustness certificate), (b) can be scaled to state-of-the-art deep networks without affecting the inference time (unlike smoothing-based approaches \cite{cohen2019certified}), and (c) work equally well for different threat models.
	Adversarial training can be formulated as a robust optimization problem \cite{shaham2015understanding, madry2018towards} which takes the form of a non-convex non-concave min-max problem. However, computing the optimal adversarial examples is an NP-hard problem \cite{katz2017reluplex,weng2018towards}. Thus adversarial training can only rely on approximate methods to solve the inner maximization problem.
	
	\begin{figure}[t]
		\centering
		\includegraphics[width=0.48\textwidth]{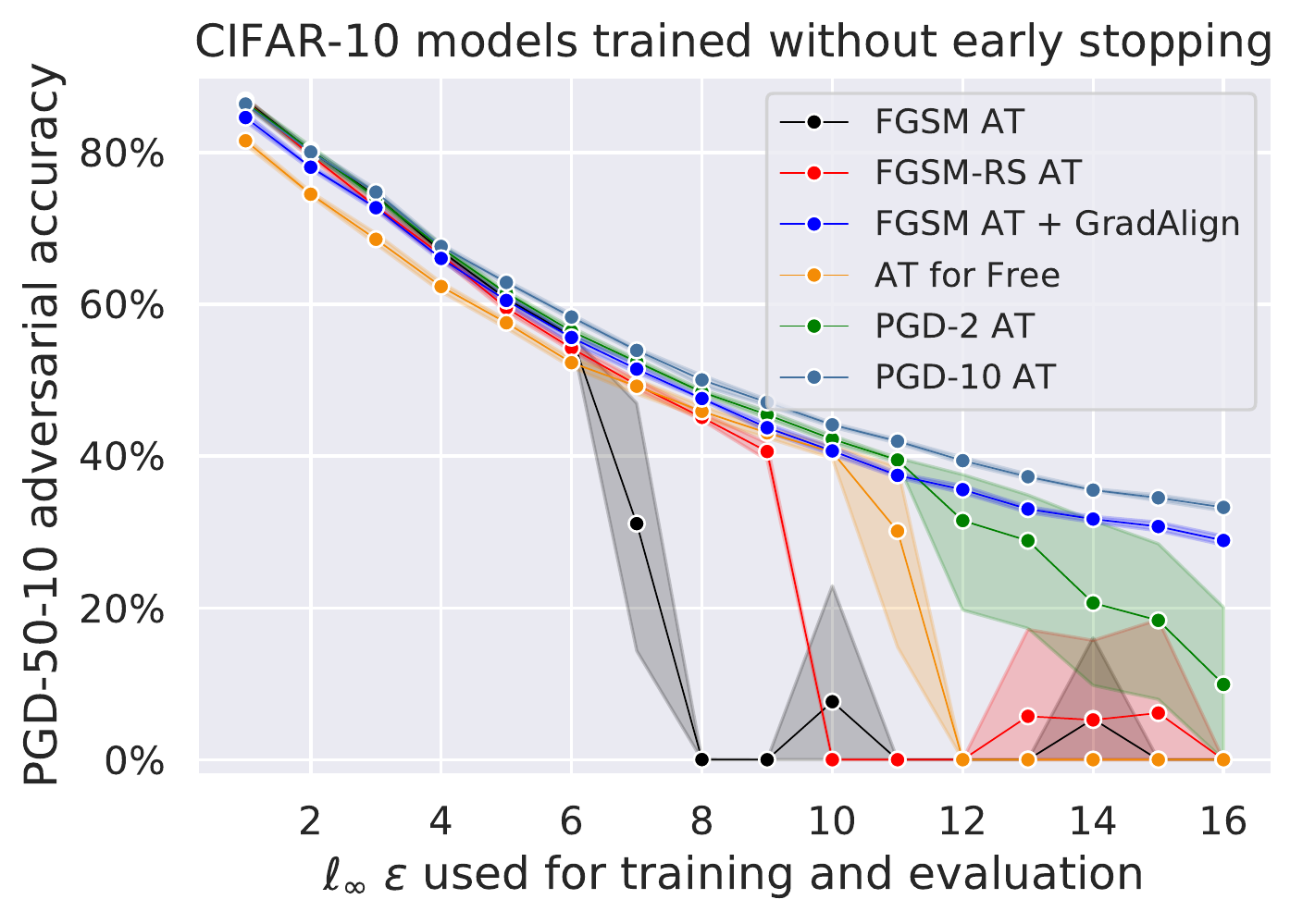} \hspace{2mm}
		\includegraphics[width=0.48\textwidth]{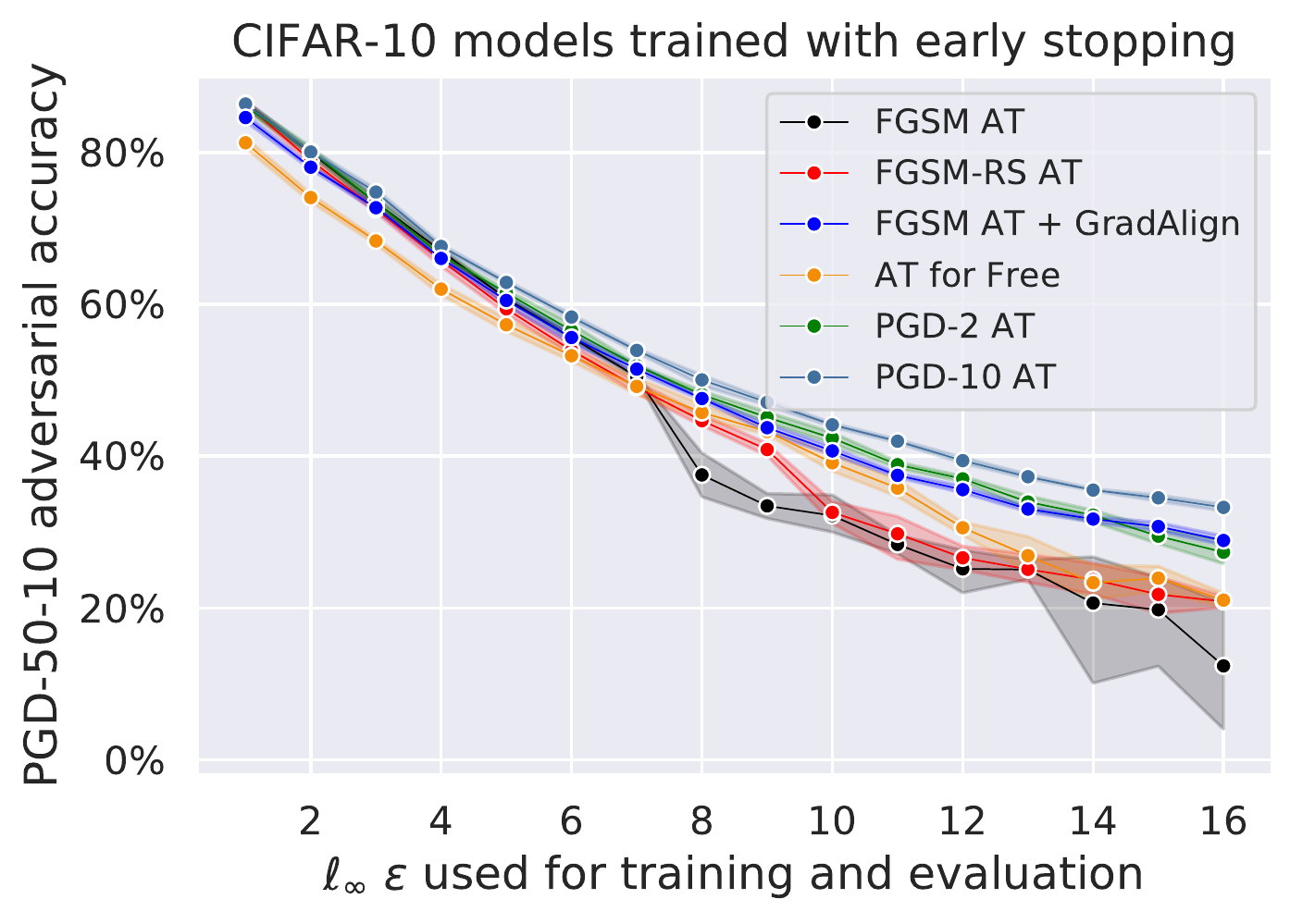}
		\caption{Robustness of different adversarial training (AT) methods on CIFAR-10 with ResNet-18 trained and evaluated with different $l_\infty$-radii. The results are averaged over 5 random seeds used for training and reported with the standard deviation.
			\textbf{FGSM AT}: standard FGSM AT,
			\textbf{FGSM-RS AT}: FGSM AT with a random step \cite{wong2020fast},
			\textbf{FGSM AT + GradAlign}: FGSM AT combined with our proposed regularizer \texttt{GradAlign},
			\textbf{AT for Free}: recently proposed method for fast PGD AT \cite{shafahi2019adversarial},
			\textbf{PGD-2/PGD-10 AT}: AT with a 2-/10-step PGD-attack.
			Our proposed regularizer \texttt{GradAlign} prevents \textit{catastrophic overfitting} in FGSM training and leads to significantly better results which are close to the computationally demanding PGD-10 AT.}
		\label{fig:teaser_plot}
	\end{figure}
	One popular approximation method successfully used in adversarial training is the PGD attack \cite{madry2018towards} where multiple steps of projected gradient descent are performed.
	It is now widely believed that models adversarially trained via the PGD attack~\cite{madry2018towards,zhang19theoretically} are robust since small adversarially trained networks can be formally verified~\cite{carlini2017provably,tjeng2017evaluating,wong2020fast}, and larger models could not be broken on public challenges~\cite{madry2018towards,zhang19theoretically}.
	Recently, \cite{croce2020reliable} evaluated the majority of recently published defenses to conclude that the standard $\ell_\infty$ PGD training achieves the best empirical robustness; a result which can only be improved using semi-supervised approaches~\cite{hendrycks2019using,alayrac2019arelabels,carmon2019unlabeled}. 
	In contrast, other empirical defenses that were claiming improvements over standard PGD training had overestimated the robustness of their reported models~\cite{croce2020reliable}. 
	These experiments imply 
	that adversarial training in general is the key algorithm for robust deep learning, and thus that performing it efficiently is of paramount importance. 
	
	Another approximation method for adversarial training is the \textit{Fast Gradient Sign Method} (FGSM) \cite{goodfellow2014explaining} which is based on the linear approximation of the neural network loss function. 
	However, the literature is still ambiguous about the performance of FGSM training, i.e. it remains unclear whether FGSM training can \textit{consistently} lead to robust models. For example, \cite{madry2018towards} and \cite{tramer2018ensemble} claim that FGSM training works only for small $\ell_\infty$-perturbations, while \cite{wong2020fast} suggest that FGSM training can lead to robust models for arbitrary $\ell_\infty$-perturbations if one adds uniformly random initialization before the FGSM step. 
	Related to this, \cite{wong2020fast} further identified a phenomenon called \textit{catastrophic overfitting} where FGSM training first leads to \textit{some} robustness at the beginning of training, but then suddenly becomes non-robust within a single training epoch. 
	However, the reasons for such a 
	failure remain unknown. This motivates us to consider the following question as the main theme of the paper:
	\begin{center}
		\textit{When and why does fast adversarial training with FGSM lead to robust models?}
	\end{center}

	\myparagraph{Contributions.}
	We first show that not only FGSM training is prone to \textit{catastrophic overfitting}, but the recently proposed fast adversarial training methods~\cite{shafahi2019adversarial,wong2020fast} as well (see Fig.~\ref{fig:teaser_plot}). 
	We then analyze the reasons why using a random step in FGSM \cite{wong2020fast} helps to slightly mitigate catastrophic overfitting and show it simply boils down to reducing the average magnitude of the perturbations.
	Then we discuss the connection behind catastrophic overfitting and local linearity in deep networks and in single-layer convolutional networks where we show that even \textit{a single filter} can make the network non-linear \textit{locally}, and  causes the failure of FGSM training.
	We additionally provide
	for this case a theoretical explanation
	which helps to explain why FGSM AT is successful at the beginning of the training.
	Finally, we propose a regularization method, \texttt{GradAlign}, that \textit{prevents catastrophic overfitting} by explicitly maximizing the gradient alignment inside the perturbation set and therefore improves the quality of the FGSM solution. 
	We compare  \texttt{GradAlign} to other adversarial training schemes in Fig.~\ref{fig:teaser_plot} and point out that among all fast adversarial training methods considered only FGSM~+~\texttt{GradAlign} does not suffer from catastrophic overfitting and leads to high robustness even for large $\ell_\infty$-perturbations.

	\section{Problem overview and related work}
	
	Let $\l(x,y;\theta)$ denote the loss of a ReLU-network  parametrized by $\theta \in \R^m $ on the example $(x,y)\sim D$ where  $D$  is the data generating distribution.\footnote{In practice we use training samples with random data augmentation.}
	Previous works \cite{shaham2015understanding,madry2018towards} formalized the goal of training adversarially robust models as the following robust optimization problem: 
	\begin{align}
	\minop_{\theta} \E_{(x, y) \sim D} \big[ \maxop_{\delta\in\Delta} \l(x+\delta, y; \theta) \big].
	\label{eq:rob_opt_general}
	\end{align}
	We focus here on the $\ell_\infty$ threat model, i.e. $\Delta = \{\delta\in\R^d, \norm{\delta}_\infty \leq \varepsilon\}$, where the adversary can change  each input coordinate $x_i$  by at most $\varepsilon$.
	Unlike classical stochastic saddle point problems 
	of the form $\min_{\theta} \max_{\delta}  \E [\l(\theta, \delta)]$~\cite{juditsky2011solving}, the inner maximization problem here is inside the expectation. Therefore the solution
	of each subproblem $\max_{\delta\in\Delta} \l(x+\delta, y; \theta)$ depends on the particular example $(x,y)$ and standard algorithms such as gradient descent-ascent which alternate gradient descent in $\theta$ and gradient ascent in $\delta$ cannot be used. 
	Instead each of these \textit{non-concave} maximization problems has to be solved independently. Thus, an inherent trade-off appears between computationally efficient approaches which aim at solving this inner problem in as few iterations as possible and approaches which aim at solving the problem more accurately but with more iterations. In an extreme case, the PGD attack~\cite{madry2018towards} uses multiple steps of projected gradient ascent (PGD), which is accurate but computationally expensive. At the other end of the spectrum, Fast Gradient Sign Method (FGSM)~\cite{goodfellow2014explaining} performs \textit{only} one iteration of gradient ascent with respect to the $\ell_\infty$-norm:
	\begin{align} \label{eq:fgsm-def}
	\delta_{FGSM} \defeq \varepsilon \sign(\nabla_x \l(x, y; \theta)),
	\end{align}
	followed by a projection of $x+\delta_{FGSM}$ onto the $[0, 1]^d$ to ensure it is a valid input.\footnote{Throughout the paper we will focus on image classification, i.e. inputs $x$ will be images.} This leads to a fast algorithm which, however, does not always lead to robust models as observed in \cite{madry2018towards,tramer2018ensemble}. 
	A~closer look at the evolution of the robustness during FGSM AT reveals that using FGSM can lead to a model with some degree of robustness but only until a point where the robustness suddenly drops. This phenomenon is called \textit{catastrophic overfitting} in \cite{wong2020fast}.
	As a partial solution, the training can be stopped just before that point which leads to non-trivial but suboptimal robustness as illustrated in Fig.~\ref{fig:teaser_plot}. 
	\citet{wong2020fast} further notice that initializing FGSM from a random starting point $\eta \sim \U([-\varepsilon, \varepsilon]^d)$, i.e. using the following perturbation where $\Pi_{[-\varepsilon, \varepsilon]^d}$ denotes the projection:
	\begin{align} \label{eq:fgsmrs-def}
	\delta_{FGSM-RS} \defeq \Pi_{[-\varepsilon, \varepsilon]^d}[\eta + \alpha \sign(\nabla_x \l(x+\eta, y; \theta))],
	\end{align}
	helps to mitigate catastrophic overfitting and leads to better robustness for the considered $\varepsilon$ values (e.g. $\varepsilon=\nicefrac{8}{255}$ on CIFAR-10 in \cite{wong2020fast}).  Along the same lines, \cite{babu2020single} observe that using dropout on all layers (including convolutional) also helps to stabilize FGSM AT. 
	
	An alternative solution is to interpolate between FGSM and PGD AT. For example, \cite{wang2019convergence} suggest to first use FGSM AT, and later to switch to multi-step PGD AT which is motivated by their analysis suggesting that the inner maximization problem has to be solved more accurately at the end of training. 
	\cite{shafahi2019adversarial} propose to run PGD with step size $\alpha=\varepsilon$ and simultaneously update the weights of the network. 
	On a related note, \cite{zhang2019propagate} collect the weight updates during PGD, but apply them after PGD is completed. Additionally, \cite{zhang2019propagate} update the gradients of the first layer multiple times. 
	However, none of these approaches are conclusive, either leading to comparable robustness to FGSM-RS training \cite{wong2020fast} and still failing for higher $\ell_\infty$-radii (see Fig.~\ref{fig:teaser_plot} for \cite{shafahi2019adversarial} and \cite{wong2020fast}) or being in the worst case as expensive as multi-step PGD AT \cite{wang2019convergence}. 
	Additionally, some previous works deviate from the robust optimization formulation stated in Eq.~\eqref{eq:rob_opt_general} and instead regularize the model to improve robustness \cite{simon2019first,moosavi2019robustness,qin2019adversarial}, however this does not lead to higher robustness compared to standard adversarial training.
	We focus next on analyzing the FGSM-RS training \cite{wong2020fast} as the other recent variations of fast adversarial training \cite{shafahi2019adversarial,zhang2019propagate,babu2020single} lead to models with similar robustness.

	\myparagraph{Experimental setup.}
	Unless mentioned otherwise, we perform training on PreAct ResNet-18~\cite{he2016identity} with the cyclic learning rates \cite{smith2017cyclical} and half-precision training \cite{micikevicius2018mixed} following the setup of \cite{wong2020fast}.
	We evaluate adversarial robustness using the PGD-50-10 attack, i.e. with 50 iterations and 10 restarts with step size $\alpha=\nicefrac{\varepsilon}{4}$ following \cite{wong2020fast}.
	More experimental details are specified in Appendix~\ref{app:exp_details}.

	\section{The role and limitations of using random initialization in FGSM training} \label{sec:role_limitation_of_rs}
	First, we show that FGSM with a random step fails to resolve catastrophic overfitting for larger $\varepsilon$. Then we provide evidence against the explanation given by~\cite{wong2020fast} on the benefit of randomness for FGSM AT, and propose a new explanation based on the linear approximation quality of FGSM.
	
	\myparagraph{FGSM with random step does not resolve catastrophic overfitting.}
	Crucially, \cite{wong2020fast} observed that adding an initial random step to FGSM as in Eq.~\eqref{eq:fgsmrs-def} helps to avoid catastrophic overfitting. 
	However, this holds only if the step size is not too large (as illustrated in Fig.~3 of \cite{wong2020fast} for $\varepsilon=\nicefrac{8}{255}$) and, more importantly, only for small enough $\varepsilon$ as we show in Fig.~\ref{fig:teaser_plot}.
	Indeed, using the step size $\alpha=1.25\varepsilon$ recommended by~\cite{wong2020fast} \textit{extends} the working regime of FGSM but only from $\varepsilon=\nicefrac{6}{255}$ to $\varepsilon=\nicefrac{9}{255}$, with $0\%$ adversarial accuracy for $\varepsilon=\nicefrac{10}{255}$.
	When early stopping is applied (Fig.~\ref{fig:teaser_plot}, right), there is still a significant gap compared to PGD-10 training, particularly for large $\ell_\infty$-radii. For example, for $\varepsilon=\nicefrac{16}{255}$, FGSM-RS AT leads to 22.24\% PGD-50-10 accuracy while PGD-10 AT obtains a much better accuracy of 30.65\%.

	\myparagraph{Previous explanation: randomness diversifies the threat model.} 
	A hypothesis stated in \cite{wong2020fast} was that FGSM-RS helps to avoid catastrophic overfitting by diversifying the threat model.
	Indeed, the random step allows to have perturbations not only at the corners $\{-\varepsilon, \varepsilon\}^d$ like the FGSM-attack\footnote{For simplicity, we ignore the projection of $x + \delta$ onto $[0, 1]^d$ in this section.}, but rather in the whole $\ell_\infty$-ball,  $[-\varepsilon, \varepsilon]^d$.
	Here we refute this hypothesis by modifying the usual PGD training by projecting onto $\{-\varepsilon, \varepsilon\}^d$ the perturbation obtained via the PGD attack.
	We perform experiments on CIFAR-10 with ResNet-18 with $\ell_\infty$-perturbations of radius $\varepsilon=\nicefrac{8}{255}$ 
	over 5 random seeds. 
	FGSM AT 
	leads to catastrophic overfitting achieving $0.00\pm0.00\%$ adversarial accuracy if early stopping is not applied, while the standard PGD-10 AT and our modified PGD-10 AT schemes  achieve $50.48\pm0.20\%$ and $50.64\pm0.23\%$ adversarial accuracy respectively. Thereby similar robustness as the original PGD AT can still be achieved  without training on pertubations from the interior of the $\ell_\infty$-ball. 
	We conclude that \textit{diversity} of adversarial examples is not crucial here.
	What makes the difference is rather having an iterative instead of a single-step procedure to find a corner of the $\ell_\infty$-ball that sufficiently maximizes the loss.

	\myparagraph{New explanation: a random step improves the linear approximation quality.}
	Using a random step in FGSM is \textit{guaranteed} to decrease the expected magnitude of the perturbation. 
	This simple observation is formalized in the following lemma.
	
	\begin{lemma}{\normalfont \textbf{(Effect of the random step)}} \label{lem:norm_fgsm_rs} 
		Let $\eta \sim \U([-\varepsilon, \varepsilon]^d)$ be a random starting point, and  $\alpha \in [0, 2\varepsilon]$ be the step size of FGSM-RS defined in Eq.~\eqref{eq:fgsmrs-def}, then
		\begin{align}
		\E_\eta \left[ \norm{\delta_{FGSM-RS}(\eta)}_2 \right] \leq 
		\sqrt{\E_\eta \left[ \norm{\delta_{FGSM-RS}(\eta)}_2^2 \right]} =
		\sqrt{d} \sqrt{-\frac{1}{6\varepsilon} \alpha^3 + \frac{1}{2} \alpha^2 + \frac{1}{3} \varepsilon^2}.
		\end{align}
	\end{lemma}
	\begin{figure}[b]
		\centering
		\begin{minipage}{.47\textwidth}
			\centering
			\includegraphics[width=0.9\linewidth]{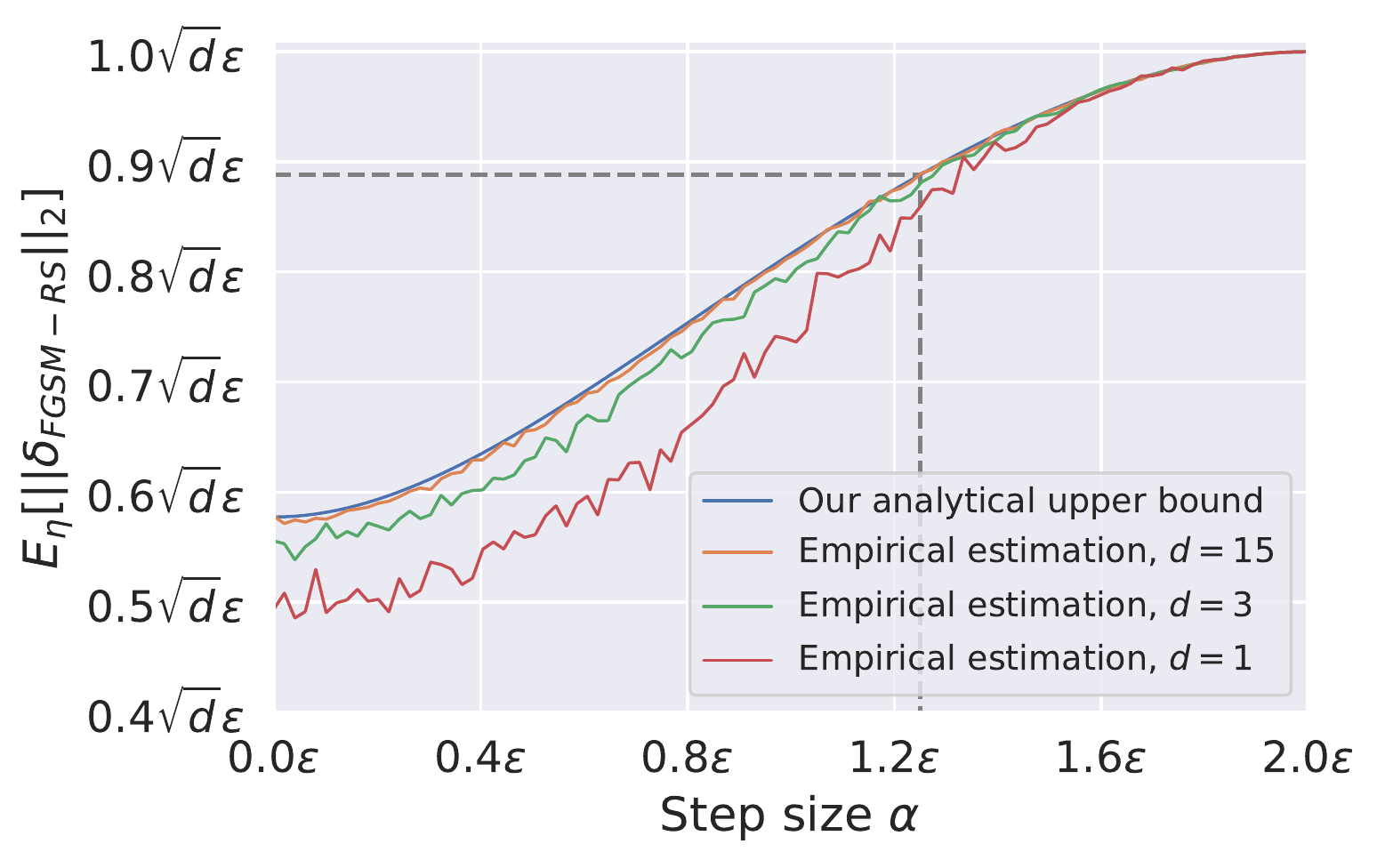} 
			\caption{Visualization of our upper bound on $\E_\eta[\norm{\delta_{FGSM-RS}}_2]$. The dashed line corresponds to the step size $\alpha=1.25\varepsilon$ recommended in \cite{wong2020fast}.}
			\label{fig:norm_fgsm_rs}
		\end{minipage}%
		\hspace{3.5mm}
		\begin{minipage}{.47\textwidth}
			\centering
			\includegraphics[width=0.82\linewidth]{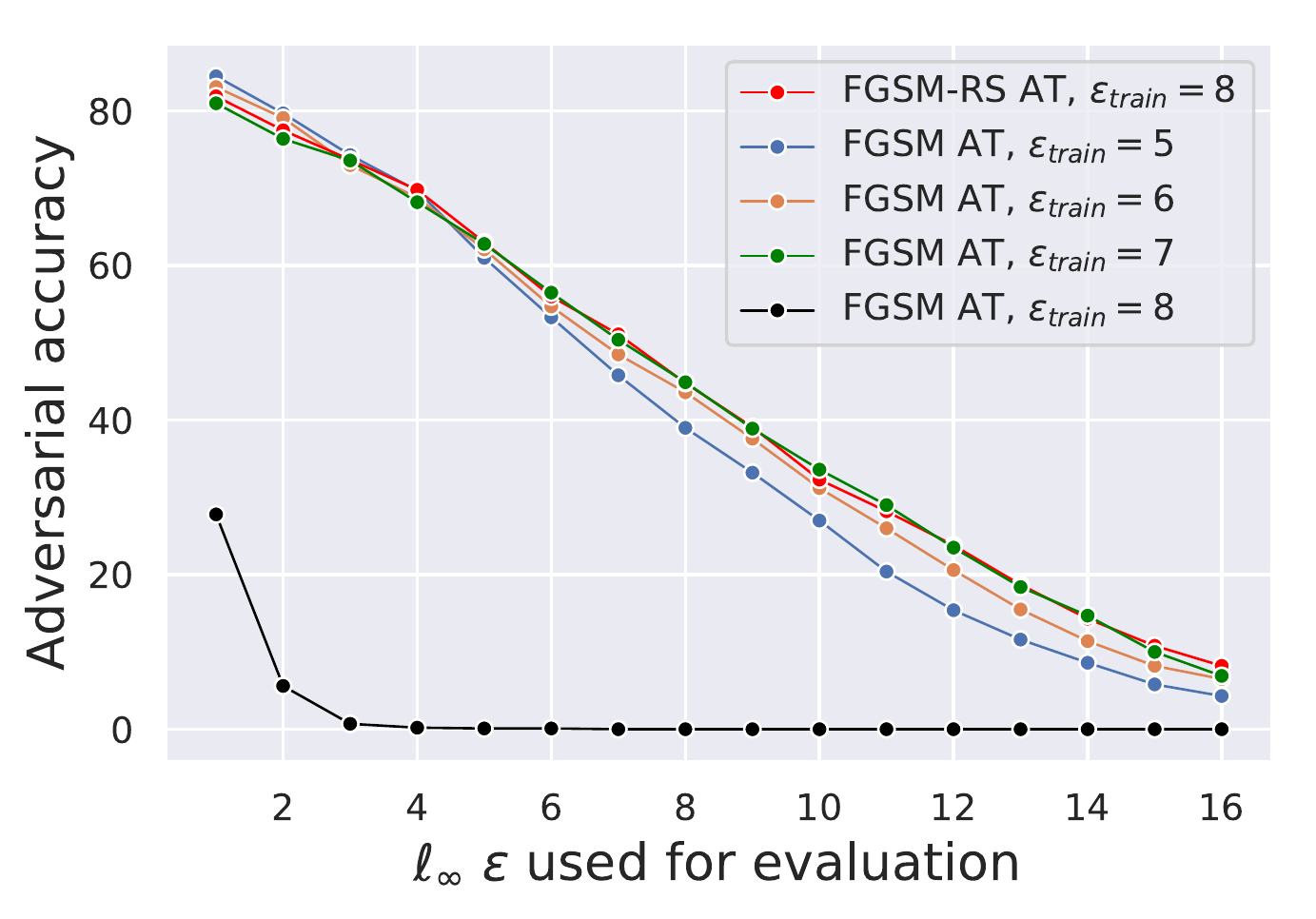}
			\caption{Robustness of FGSM-trained ResNet-18 on CIFAR-10 with different $\varepsilon_{train}$ used for training compared to FGSM-RS AT with $\varepsilon_{train}=\nicefrac{8}{255}$.}
			\label{fig:fgsm_diff_eps_eval}
		\end{minipage}
	\end{figure}
	The proof is deferred to Appendix~\ref{sec:proof_norm_fgsm_rs}. We first remark that the upper bound is in the range $[\nicefrac{1}{\sqrt{3}} \sqrt{d} \varepsilon,  \sqrt{d} \varepsilon]$, and therefore always less or equal than $\norm{\delta_{FGSM}}_2 = \sqrt{d} \varepsilon.$
	We visualize our bound in Fig.~\ref{fig:norm_fgsm_rs} where the expectation is approximated by Monte-Carlo sampling over $1{,}000$ samples of $\eta$, and note that the bound becomes increasingly tight for high-dimensional inputs. 
	
	The key observation here is that among all possible perturbations of $\ell_\infty$-norm $\varepsilon$, perturbations with a smaller $\ell_2$-norm benefit from a better linear approximation. 
	This statement follows from the second-order Taylor expansion for twice differentiable functions: 
	\begin{equation*}
	f(x+\delta) \approx f(x) + \inner{\nabla_x f(x), \delta} + \inner{\delta, \nabla^2_{xx} f(x) \delta},
	\end{equation*}
	i.e. a smaller value of $\norm{\delta}_2^2$ implies a smaller linear approximation error $|f(x+\delta) - f(x) - \inner{\nabla_x f(x), \delta}|$.
	Moreover, the same property still holds empirically for the non-differentiable ReLU networks (see Appendix~\ref{app:sec:quality_lin_approx}). 
	We conclude that by reducing in expectation the length of the perturbation $\norm{\delta}_2$, the FGSM-RS approach of \cite{wong2020fast} takes advantage  of a better linear approximation. 
	This is supported by the fact that FGSM-RS AT also leads to catastrophic overfitting if the step size $\alpha$ is chosen to be too large 
	(see Fig.~3 in \cite{wong2020fast}), thus providing no benefits over FGSM AT even when combined with early stopping.
	We argue this is the main improvement over the standard FGSM AT.

	\myparagraph{Successful FGSM AT does not require randomness.}
	If having perturbation with a too large $\ell_2$-norm is indeed the key factor in catastrophic overfitting, we can expect that just reducing the step size of the standard FGSM  should work equally well as FGSM-RS. For  $\varepsilon=\nicefrac{8}{255}$ on CIFAR-10, \cite{wong2020fast} recommend to use FGSM-RS with step size $\alpha=1.25\varepsilon$ which induces a perturbation of expected $\ell_2$-norm  $\norm{\delta_{FGSM-RS}}_2 \approx \nicefrac{7}{255} \sqrt{d}$. 
	This corresponds to using standard FGSM with a step size $\alpha \approx \nicefrac{7}{255}$ instead of $\alpha=\varepsilon=\nicefrac{8}{255}$  (see the dashed line in Fig.~\ref{fig:norm_fgsm_rs}).
	We report the results in Table~\ref{tab:fgsm_reduced_step_size_eps8}
	and observe that simply reducing the step size of  FGSM (without \textit{any} randomness) leads to the same level of robustness.
	We show further in Fig.~\ref{fig:fgsm_diff_eps_eval} that when used with a smaller step size, the robustness of \textit{standard} FGSM training even without early stopping can generalize to much higher~$\varepsilon$. This contrasts with the previous literature~\cite{madry2018towards,tramer2018ensemble}.
	We conclude from these experiments that a more direct way to improve FGSM AT and to prevent it from catastrophic overfitting is to simply reduce the step size. Note that this still leads to suboptimal robustness compared to PGD AT (see Fig.~\ref{fig:teaser_plot}) for $\varepsilon$ larger than the one used during training, since in this case adversarial examples can only be generated inside the smaller $\ell_\infty$-ball. 
	This motivates us to take a closer look on \textit{how} and \textit{why} catastrophic overfitting occurs to be able to prevent it without reducing the FGSM step size. 
	\begin{table}[t]
		\caption{Robustness of FGSM AT with a reduced step size ($\alpha=\nicefrac{7}{255}$) compared to the FGSM-RS AT proposed in \cite{wong2020fast} ($\alpha=\nicefrac{10}{255}$) for $\varepsilon=8/255$ on CIFAR-10 for ResNet-18 trained with early stopping. The results are averaged over 5 random seeds used for training. 
		}
		\label{tab:fgsm_reduced_step_size_eps8}
		\centering
		{\small
			\begin{tabular}{lccc}
				\toprule
				\diagbox{\textbf{Accuracy}}{\textbf{Model}}                       & FGSM AT & FGSM $\alpha=\nicefrac{7}{255}$ AT & FGSM-RS AT \\
				\midrule
				PGD-50-10   & $36.35\pm1.74\%$ & $45.35\pm0.48\%$ & $45.60\pm0.19\%$ \\
				\bottomrule
			\end{tabular}
		}
	\end{table}

	\section{Understanding catastrophic overfitting via gradient alignment} \label{sec:understanding_co_via_ga}
	First, we establish a connection between catastrophic overfitting and local linearity of the model. Then we show that catastrophic overfitting also occurs in a single-layer convolutional network, for which we analyze local linearity both empirically and theoretically.

	\myparagraph{When can the inner maximization problem be accurately solved with FGSM?}
	Recall that the FGSM attack \cite{goodfellow2014explaining} is obtained as a closed-form solution of the following optimization problem:
	$\delta_{FGSM} = \argmax_{\norm{\delta}_\infty \leq \varepsilon} \inner{\nabla_x \l(x, y; \theta)), \delta}$. 
	Thus, the FGSM attack is guaranteed to find the optimal adversarial perturbation if $\nabla_x \l(x, y; \theta)$ is constant inside the $\ell_\infty$-ball around the input $x$, i.e. the loss function is \textit{locally linear}.  
	This motivates us to study the evolution of local linearity during FGSM training and its connection to catastrophic overfitting. With this aim, we define the following local linearity metric of the loss function $\ell$:
	\begin{align} \label{eq:grad_alignment}
	\E_{(x, y) \sim D, \ \eta \sim \U([-\varepsilon, \varepsilon]^d)} \left[\cos \left( \nabla_x \l(x, y; \theta), \nabla_x \l(x+\eta, y; \theta) \right) \right], 
	\end{align}
	which we refer to as \textit{gradient alignment}. This quantity is easily interpretable: 
	it is equal to one for models linear inside the $\ell_\infty$-ball of radius $\varepsilon$, and it is approximately zero when the input gradients are nearly orthogonal to each other.
	Previous works also considered local linearity of deep networks \cite{moosavi2019robustness,qin2019adversarial}, however rather with the goal of introducing regularization methods that improve robustness as an \textit{alternative} to adversarial training.
	More precisely, \cite{moosavi2019robustness} propose to use a curvature regularization method that uses the FGSM point, and \cite{qin2019adversarial} find the input point where local linearity is maximally violated using an iterative method, leading to comparable computational cost as PGD AT.
	In contrast, we analyze here gradient alignment to improve FGSM training without seeking an alternative to it. 
	
\begin{figure}[t]
		\centering
		\includegraphics[width=0.32\textwidth]{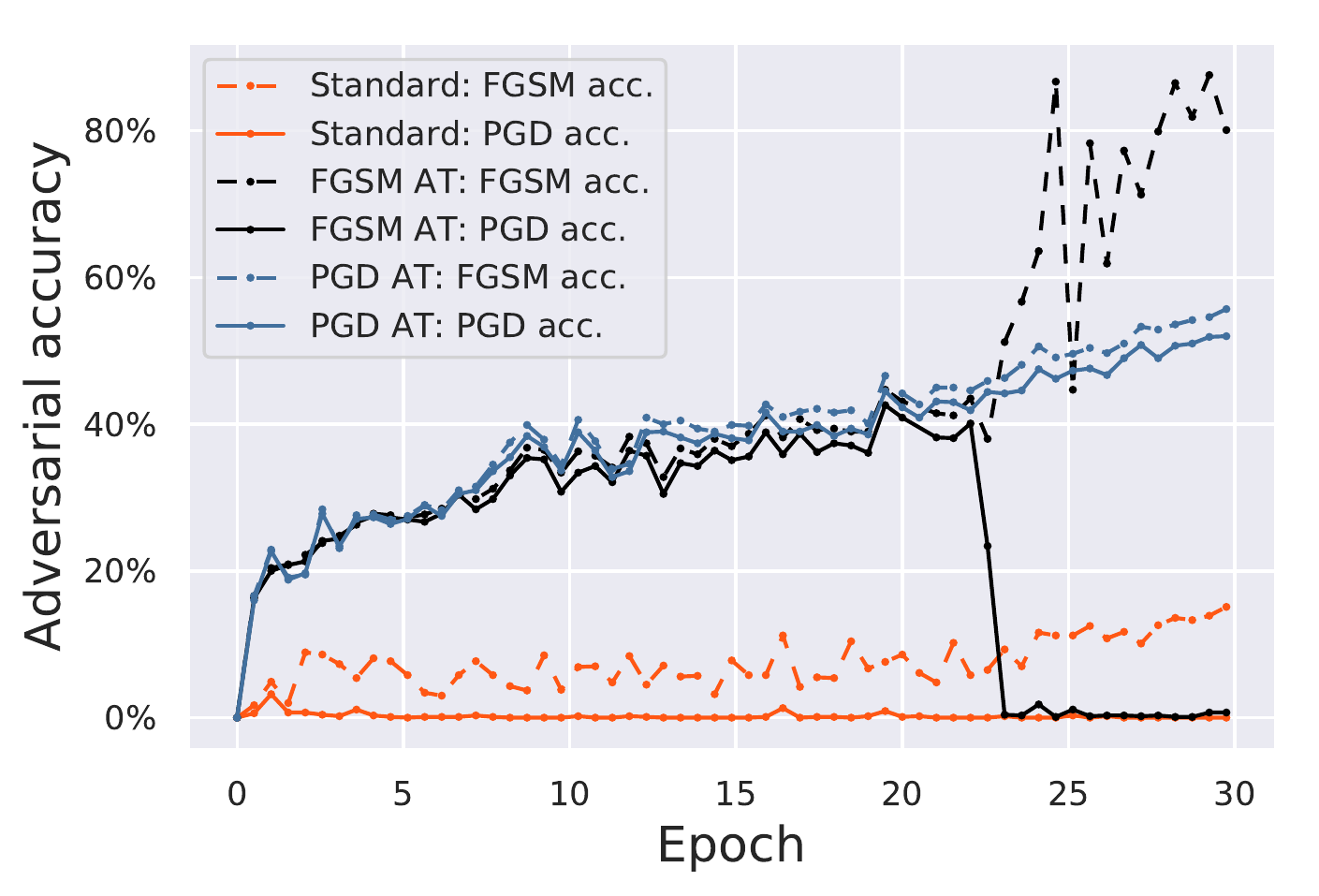}
		\includegraphics[width=0.32\textwidth]{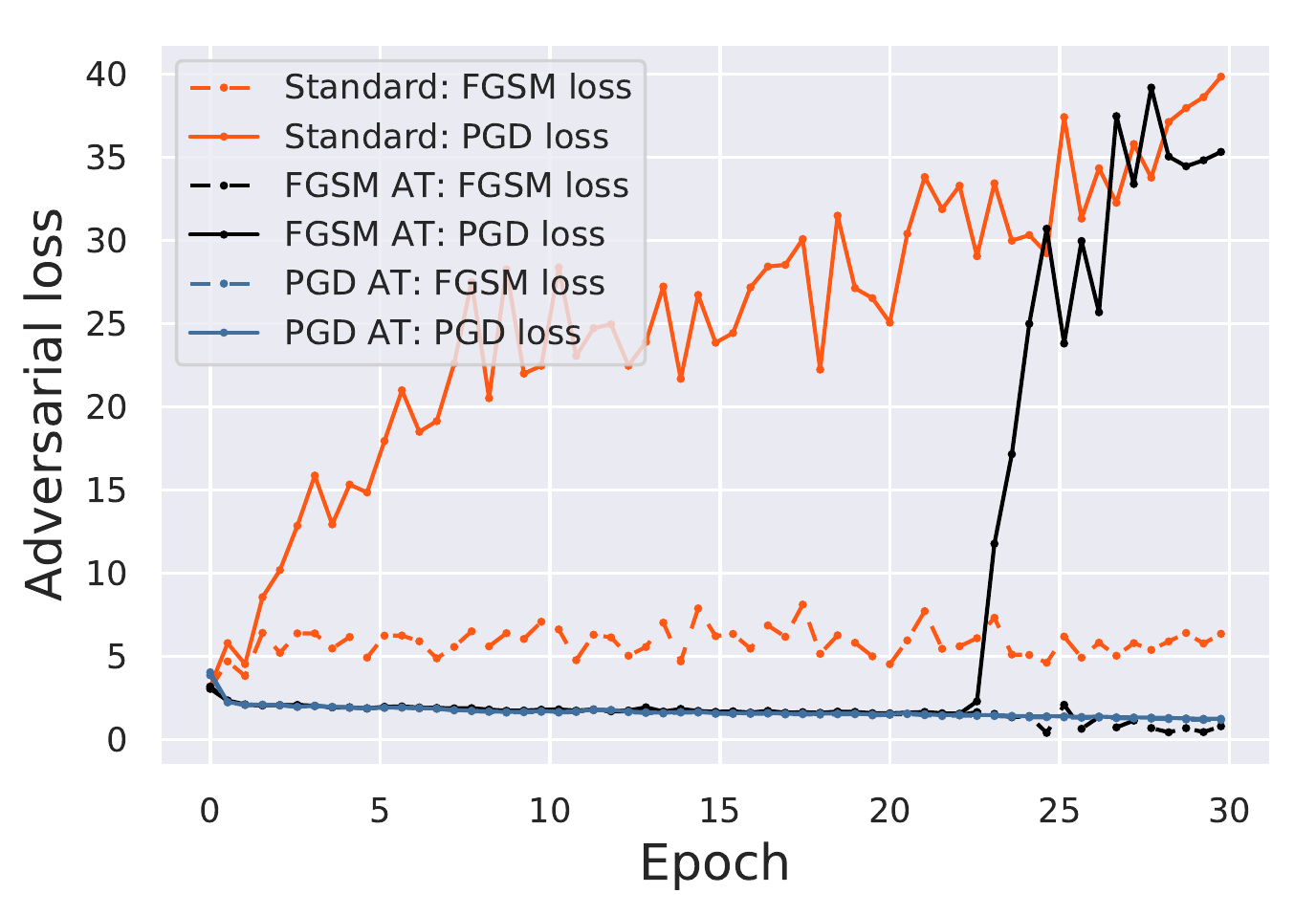}
		\includegraphics[width=0.32\textwidth]{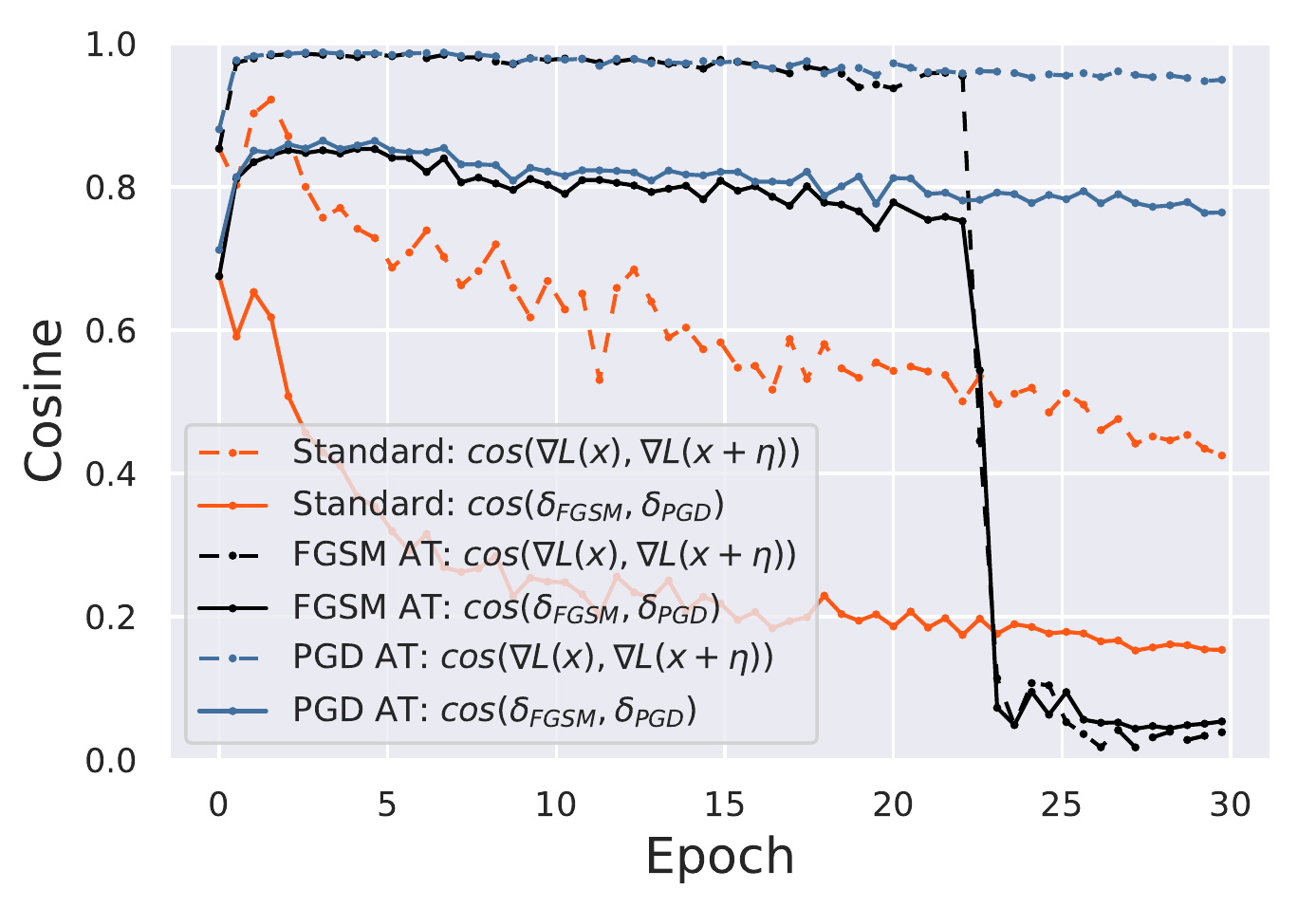}
		\caption{Visualization of the training process of standardly trained, FGSM trained, and PGD-10 trained ResNet-18 on CIFAR-10 with $\varepsilon = \nicefrac{8}{255}$. All the statistics are calculated on the test set. Catastrophic overfitting for the FGSM AT model occurs around epoch 23 and is characterized by a sudden drop in the PGD accuracy, a gap between the FGSM and PGD losses, and a dramatic decrease of \textit{local linearity}.}
		\label{fig:training_curves_resnet}
	\end{figure}

	\myparagraph{Catastrophic overfitting in deep networks.}
	To understand the link between catastrophic overfitting and local linearity, we plot in Fig.~\ref{fig:training_curves_resnet}  the adversarial accuracies and the loss values obtained by FGSM and PGD AT on CIFAR-10 using ResNet-18, together with the gradient alignment (see Eq.~\ref{eq:grad_alignment}) and the cosine between FGSM and PGD perturbations. 
	We compute these statistics on the test set.
	Catastrophic overfitting  occurs for FGSM AT around epoch 23, and is characterized by the following intertwined events: (a)
	There is a \textit{sudden drop} in the PGD accuracy from $40.1\%$ to $0.0\%$, along with an \textit{abrupt jump}  of the FGSM accuracy from $43.5\%$ to $86.7\%$. In contrast, before the catastrophic overfitting, the ratio between the average PGD and FGSM losses never exceeded~$1.05$. This suggests that FGSM cannot anymore accurately solve the inner maximization problem.
	(b) Concurrently, after catastrophic overfitting, the gradient alignment of the FGSM model experiences a \textit{phase transition} and drops significantly from $0.95$ to $0.05$ within an epoch of training, i.e. \textit{the input gradients become nearly orthogonal inside the $\ell_\infty$-ball}. We observe the same drop also for $\cos(\delta_{FGSM}, \delta_{PGD})$ which means that the FGSM and PGD directions are not aligned anymore (as also observed in \cite{tramer2018ensemble}).
	This echoes the observation made in \cite{Nakkiran2019SGD} that SGD on the standard loss of a neural network learns models of increasing complexity. We observe qualitatively the same phenomenon for FGSM AT, where the complexity is captured by the degree of local non-linearity.
	The connection between local linearity and catastrophic overfitting sparks interest for a further analysis in a simpler setting. 

\begin{figure}[t]
		\centering
		\begin{minipage}{.36\textwidth}
			\scriptsize
			\setlength{\tabcolsep}{2pt}
			\begin{tabular}{cc}
				\hspace{-15pt} Epoch 5 (before CO) & \hspace{-12pt} Epoch 6 (after CO) \\ 
				\includegraphics[width=0.49\columnwidth]{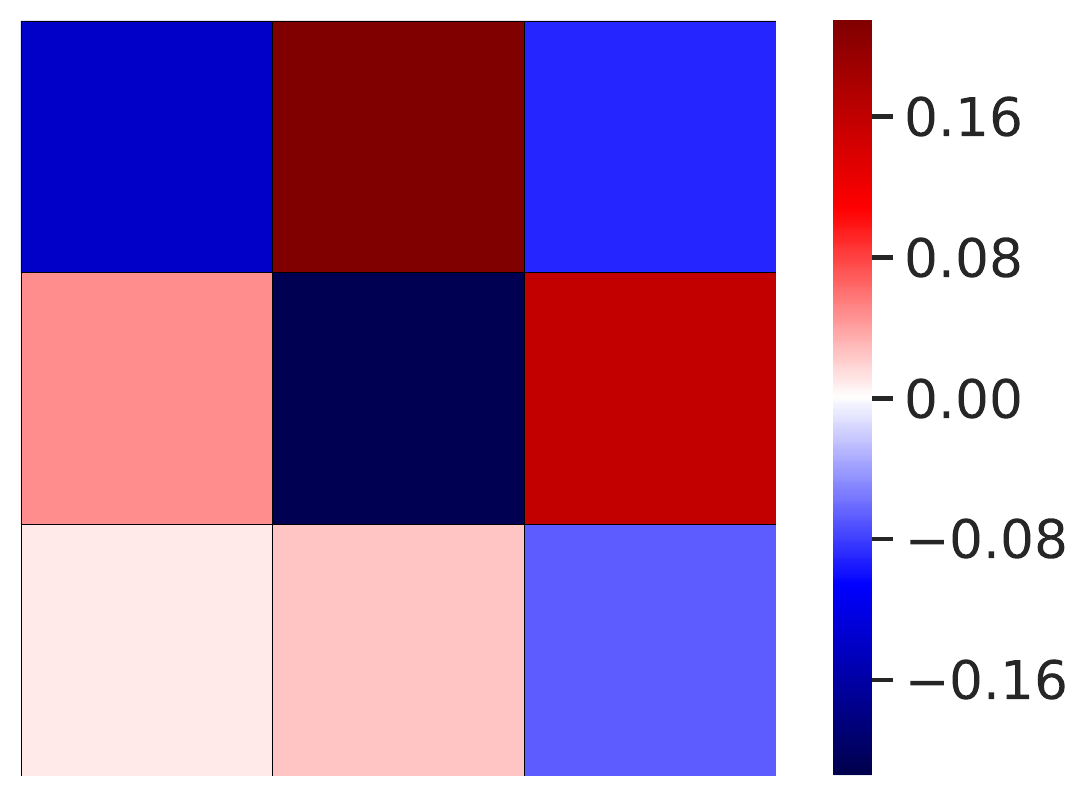} &
				\includegraphics[width=0.49\columnwidth]{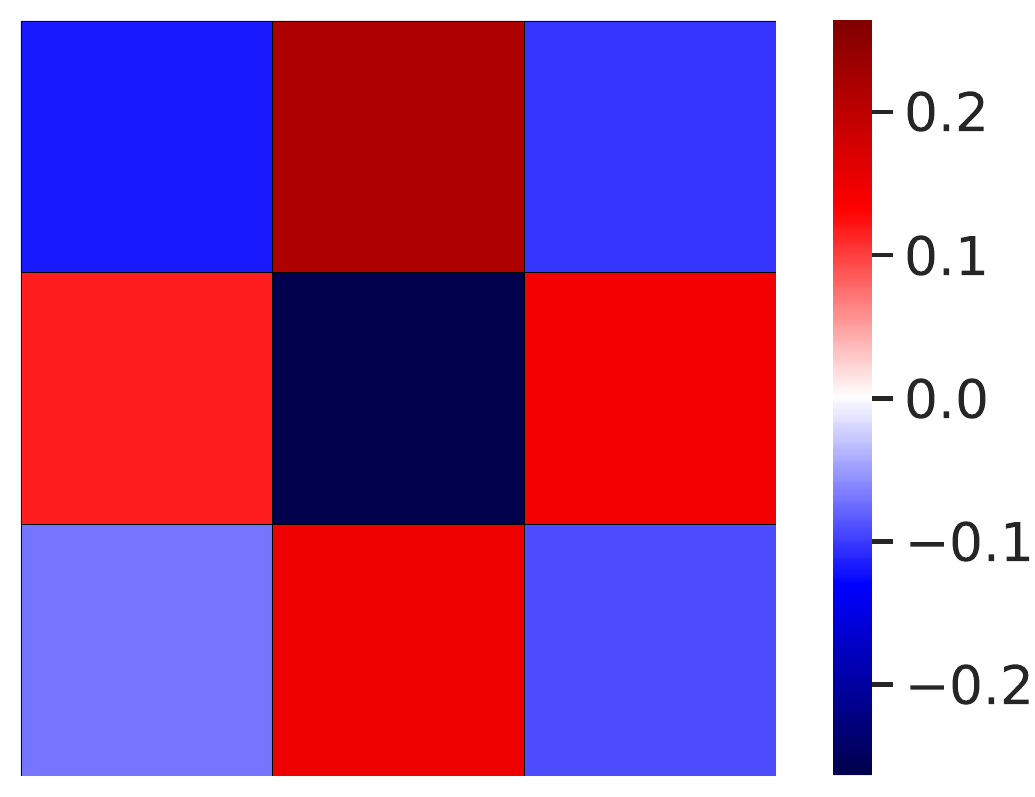} \\
			\end{tabular}
			\caption{Filter $w_4$ (green channel) in a single-layer CNN before and after catastrophic overfitting (CO). 
			}
			\label{fig:cnn4_filters}
		\end{minipage}
		\hspace{1mm}
		\begin{minipage}{.59\textwidth}
			\centering
			\scriptsize
			\includegraphics[width=0.46\columnwidth]{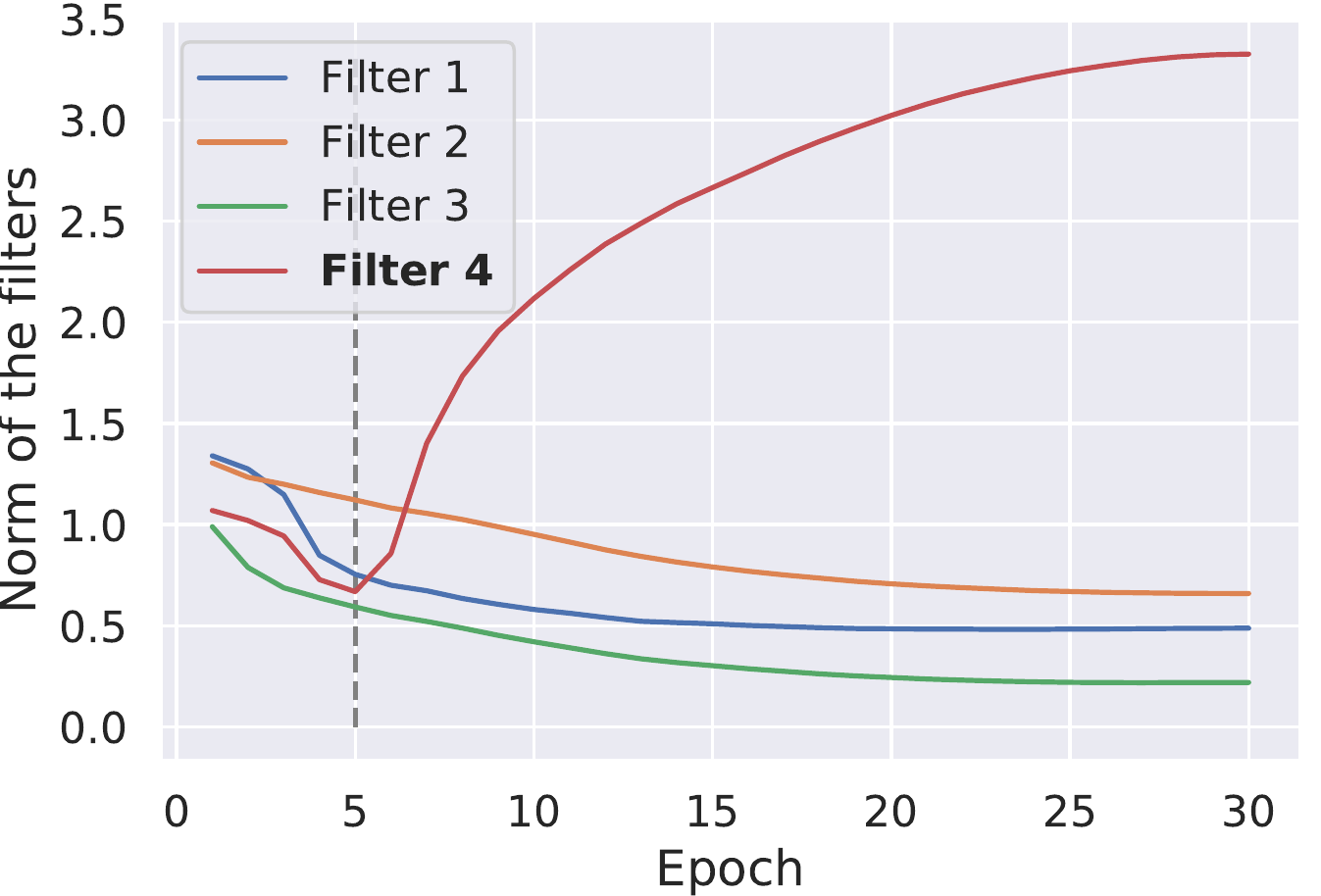} \hspace{2mm}
			\includegraphics[width=0.46\columnwidth]{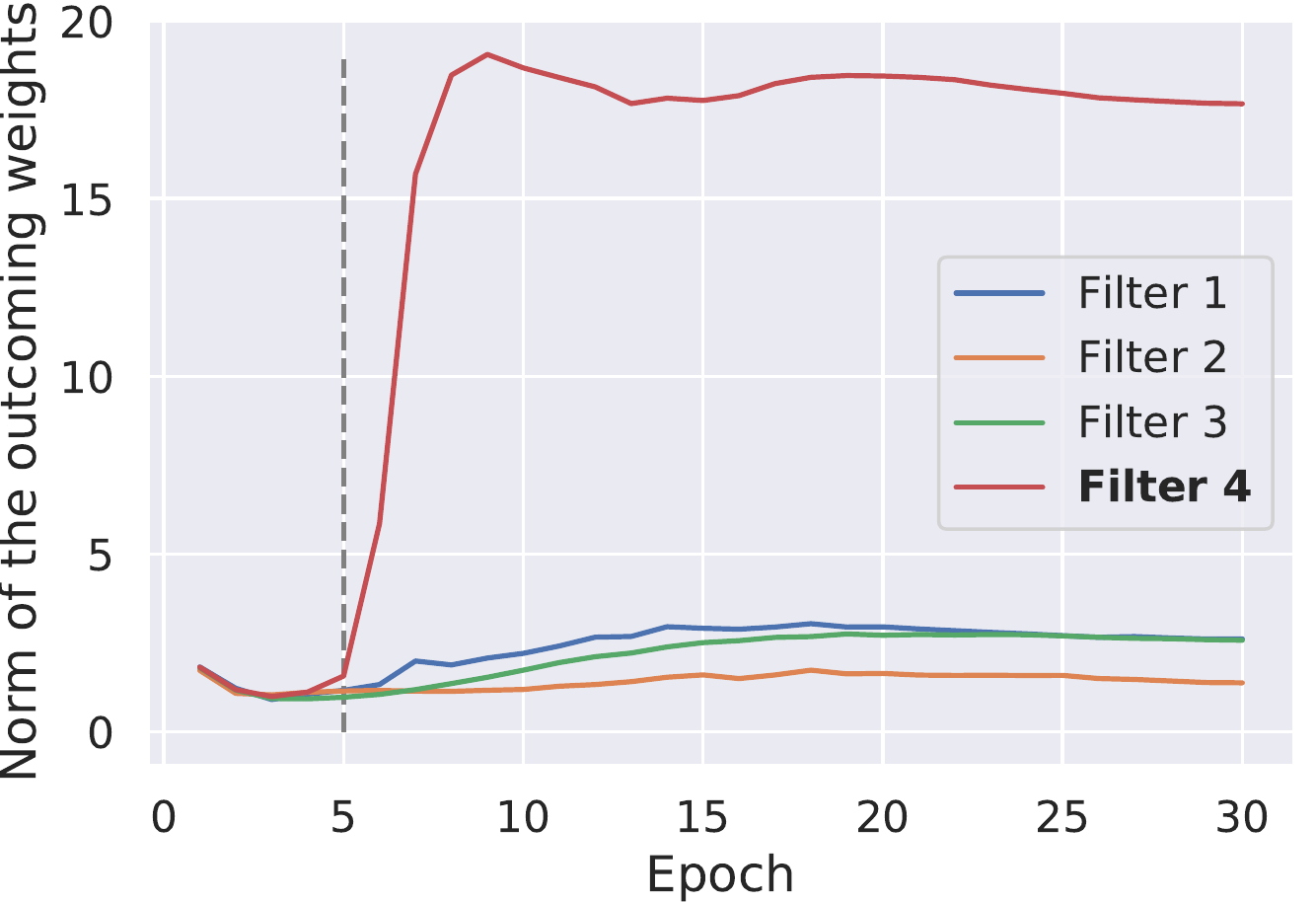}
			\caption{Evolution of the weight norms in a single-layer CNN before and after catastrophic overfitting (dashed line).
			}
			\label{fig:cnn4_filters_evolution}
		\end{minipage}
	\end{figure}

	\myparagraph{Catastrophic overfitting in a single-layer CNN.}
	We show that catastrophic overfitting is not inherent to deep and overparametrized networks, and can be observed in a very simple setup. For this we train a single-layer CNN with four filters on CIFAR-10 using FGSM AT with $\varepsilon=\nicefrac{10}{255}$ (see Sec.~\ref{app:exp_details} for details). 
	We observe that catastrophic overfitting occurs in this simple model as well, and its pattern is the same as in ResNet: a simultaneous drop of the PGD accuracy and gradient alignment (see Appendix~\ref{app:sec:cnn_cat_overfitting}).
	The advantage of considering a simple model is that we can inspect the learned filters and understand what causes the network to become highly non-linear locally.
	We observe that after catastrophic overfitting the network has learned in filter $w_4$ a variant of the Laplace filter (see Fig.~\ref{fig:cnn4_filters}), an edge-detector filter  which is well-known for \textit{amplifying high-frequency noise} such as uniform noise~\cite{gonzales2002digital}. 
	Until the end of training, filter $w_4$ preserves its direction (see Appendix~\ref{app:sec:cnn_cat_overfitting} for detailed visualizations), but grows significantly in its magnitude together with its outcoming weights, in contrast to the rest of the filters as shown in Fig.~\ref{fig:cnn4_filters}.
	Interestingly, if we set $w_4$ to zero, the network largely \textit{recovers local linearity}: the gradient alignment increases from $0.08$ to $0.71$, recovering its value before catastrophic overfitting. 
	Thus, in this extreme case, \textit{even a single convolutional filter can cause catastrophic overfitting}.
	Next we analyze formally gradient alignment in a single-layer CNN and elaborate on the connection to the noise sensitivity. 

	\myparagraph{Analysis of gradient alignment in a single-layer CNN.}
	We analyze here a single-layer CNN with ReLU-activation. 
	Let $Z \in \R^{p \times k}$ be the matrix of $k$ non-overlapping image patches extracted from the image $x = \vec(Z) \in \R^d$ such that $z_j = z_j(x) \in \R^p$. 
	The model prediction $f$ is parametrized by $(W,b,U,c)\in \R^{p \times m} \times \R^{m} \times \R^{m \times k} \times \R$, and its prediction and the input gradient are given as 
	\[
	f(x) = \sum_{i=1}^m \sum_{j=1}^k u_{ij} \max\{\inner{w_i, z_j} + b_i, 0\} + c, \quad
	\nabla_x f(x) = \vec \left( \sum_{i=1}^m \sum_{j=1}^k u_{ij} \Id_{\inner{w_i, z_j} + b_i \geq 0} w_i e_j^T \right).
	\] 
	
	We observe that catastrophic overfitting only happens at later stages of training. At the beginnning of the training, the gradient alignment is very high (see Fig.~\ref{fig:training_curves_resnet} and Fig.~\ref{fig:training_curves_cnn4}), and FGSM solves the inner maximization problem accurately enough.
	Thus, an important aspect of  FGSM training is that  the model starts training from \textit{highly aligned gradient}. This motivates us to inspect closely gradient alignment at initialization.
	\begin{lemma}{\normalfont \textbf{(Gradient alignment at initialization)}} \label{lem:grad_alignment_at_init} 
		Let $z \sim \U([0,1]^p)$ be an image patch for $p \geq 2$, $\eta \sim \U([-\varepsilon, \varepsilon]^d)$ a point inside the $\ell_\infty$-ball,
		the parameters of a single-layer CNN initialized i.i.d. as $w \sim \mathcal{N}(0, \sigma_w^2 I_p)$ for every column of $W$, $u \sim \mathcal{N}(0, \sigma_u^2 I_m)$ for every column of $U$, $b:=0$,
		then the gradient alignment is lower bounded by
		\begin{align*}
		\lim_{k,m\to \infty} \cos \left( \nabla_x \l(x, y), \nabla_x \l(x+\eta, y) \right)
		\geq
		\max\left\{ 1 - \sqrt{2} \E_{w,z} \left[e^{-\frac{1}{\varepsilon^2}\inner{\nicefrac{w}{\norm{w}_2}, z}^2} \right]^{1/2}, 0.5 \right\}.
		\end{align*}
	\end{lemma}
	\begin{wrapfigure}{r}{0.43\textwidth}
		\vspace{-15pt}
		\centering
		\tiny
		\setlength{\tabcolsep}{2pt}
		\begin{tabular}{ccc}
			$x$ & \hspace{-3.5mm}$x * w_1 + b_1$ & \hspace{-4mm}$x * w_4 + b_4$ \\
			\includegraphics[width=0.115\textwidth]{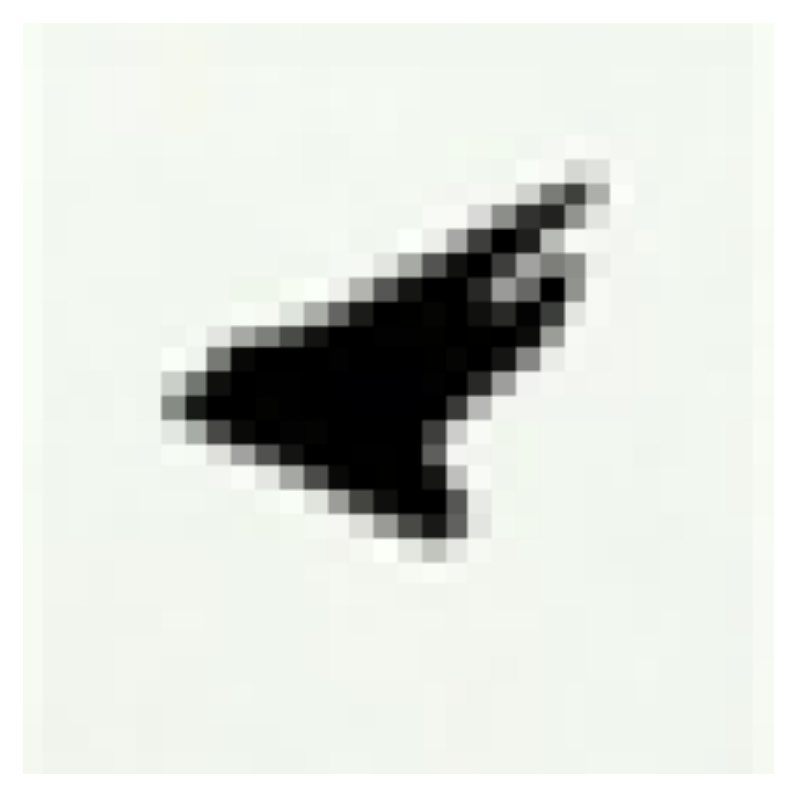} &
			\includegraphics[width=0.145\textwidth]{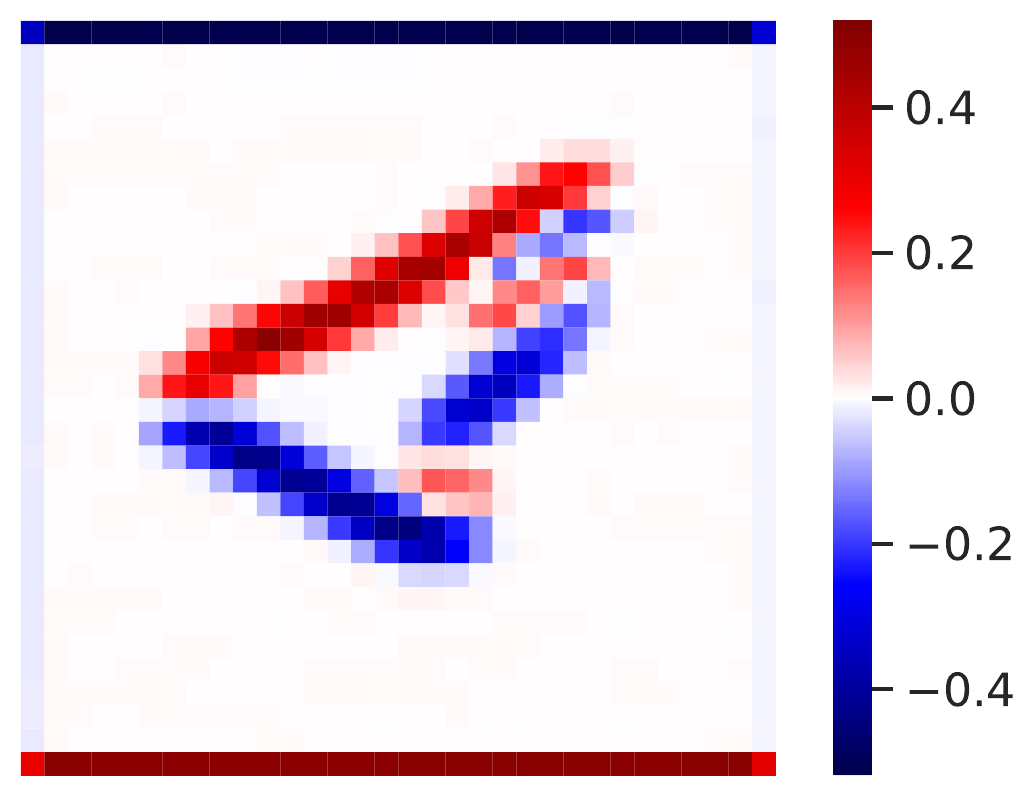} &
			\includegraphics[width=0.155\textwidth]{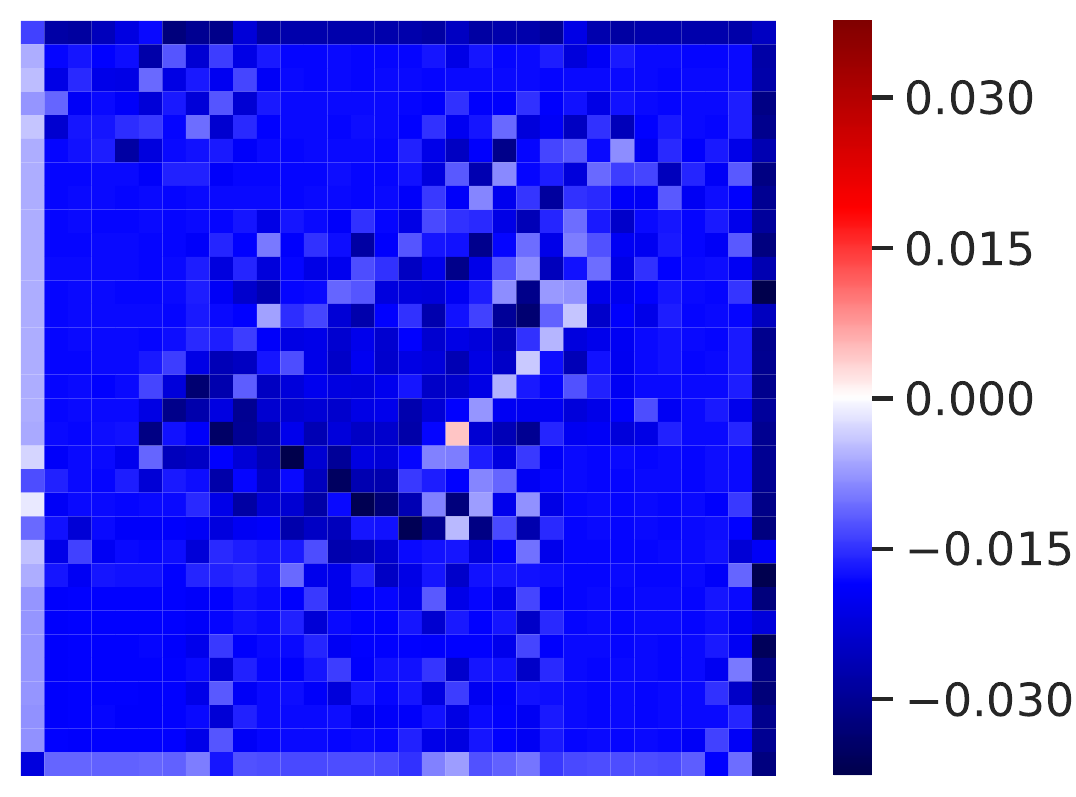} \\
			\rule{0pt}{8pt}    
			$x + \eta$ & \hspace{-1.0mm}$(x + \eta) * w_1 + b_1$ & \hspace{-1.5mm}$(x + \eta) * w_4 + b_4$ \\
			\includegraphics[width=0.115\textwidth]{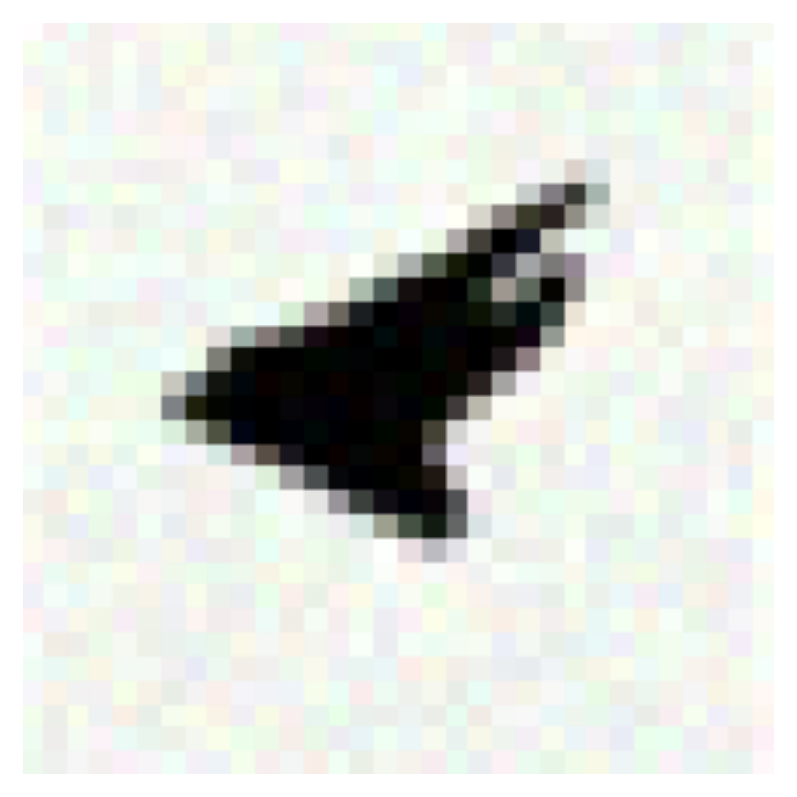} &
			\includegraphics[width=0.145\textwidth]{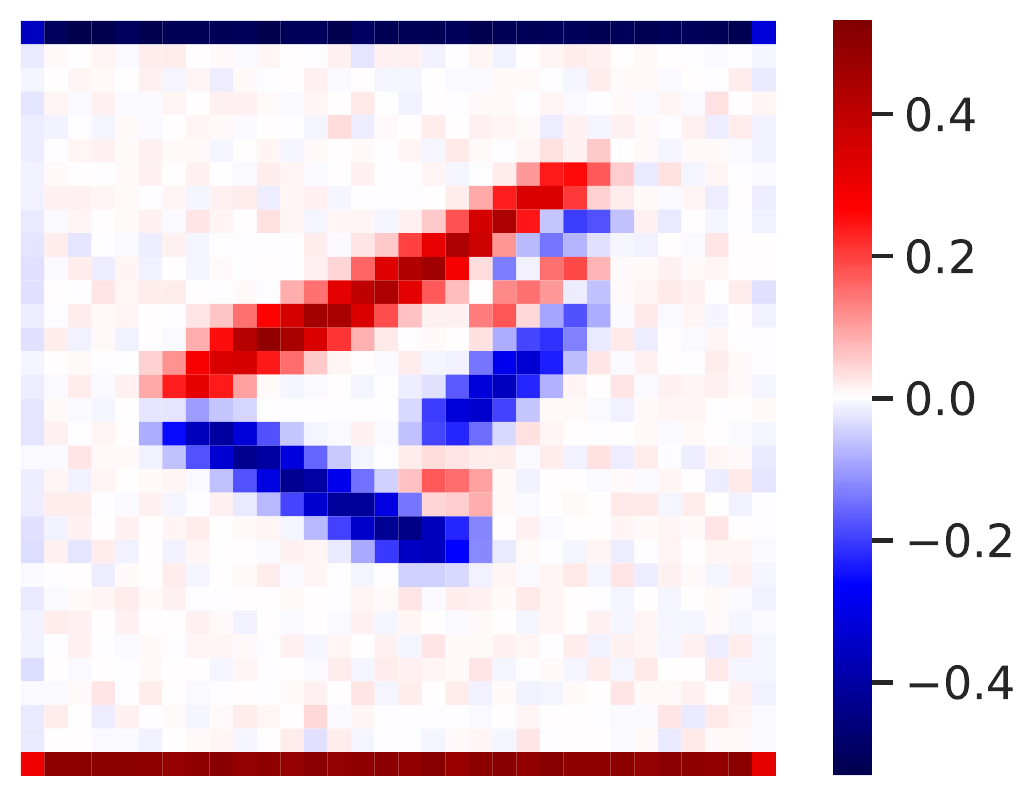} &
			\includegraphics[width=0.150\textwidth]{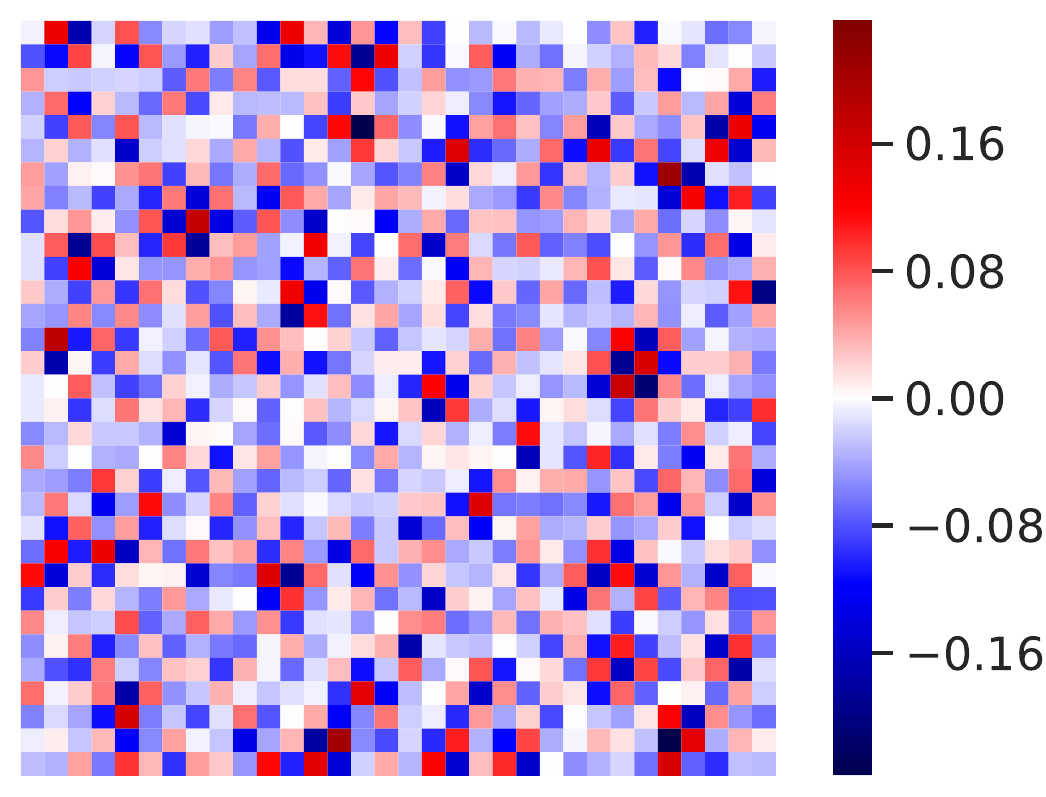}
		\end{tabular}
		\caption{Feature maps of filters $w_1$ and $w_4$ in a single-layer CNN. A small noise $\eta$ is significantly amplified by the Laplace filter $w_4$ in contrast to a regular filter $w_1$. 
		}
		\label{fig:cnn4_fms}
		\vspace{-10pt}
	\end{wrapfigure}
	The lemma implies that for randomly initialized CNNs with a large enough number of image patches $k$ and filters $m$, gradient alignment cannot be smaller than $0.5$. This is in contrast to the value of $0.12$ that we observe after catastrophic overfitting when the weights are no longer i.i.d.
	We note that the lower bound of $0.5$ is quite pessimistic since it holds for an arbitrarily large $\varepsilon$. The lower bound is close to $1$ when $\varepsilon$ is small compared to $\E\norm{z}_2$ which is typical in adversarial robustness (see Appendix~\ref{sec:grad_alignment_at_init} for the visualization of the lower bound). 
	High gradient alignment at initialization also holds empirically for 
	deep networks as well, e.g. for ResNet-18 (see Fig.~\ref{fig:training_curves_resnet}), starting from the value of $0.85$ in contrast to $0.04$ after catastrophic overfitting.
	Thus, it appears to be a general phenomenon that the standard initialization scheme of neural network weights~\cite{he2015delving} ensures the \textit{initial} success of FGSM training.
	
	In contrast, after some point during training, the network can learn parameters which lead to a significant reduction of gradient alignment. 
	For simplicity, let us consider a single-filter CNN where the gradient alignment for a filter $w$ and bias $b$ at points $x$ and $x+\eta$ has a simple expression:
	\begin{align}
	\cos \left( \nabla_x \l(x, y), \nabla_x \l(x+\eta, y) \right) &=
	\frac{\sum_{i=1}^k u_i^2 \Id_{\inner{w, z_i} + b \geq 0} \Id_{\inner{w, z_i+\eta_i} + b \geq 0}}
	{\sqrt{\sum_{i=1}^k u_i^2 \Id_{\inner{w, z_i} + b \geq 0} \sum_{i=1}^k u_i^2 \Id_{\inner{w, z_i + \eta_i} + b \geq 0}}}.
	\label{eq:cnn1_grad_alignment}
	\end{align}
	Considering a single-filter CNN is also motivated by the fact that in the single-layer CNN introduced earlier, the norms of $w_4$ and its outcoming weights are much higher than for the rest of the filters (see Fig.~\ref{fig:cnn4_filters_evolution}), and thus the contribution of $w_4$ to the predictions and gradients of the network is the most significant.
	We observe that when an image $x$ is convolved with the Laplace filter $w_4$,
	even a uniformly random noise $\eta$ of small magnitude is able to significantly affect the output of $(x + \eta) * w_4$ (see Fig.~\ref{fig:cnn4_fms}). As a consequence, the ReLU activations of the network change their signs which directly affects the gradient alignment in Eq.~\eqref{eq:cnn1_grad_alignment}.
	Namely, $x * w_4 + b_4$ has mostly negative values, and thus many values $\{\Id_{\inner{w_4, z_i} + b_4}\}_{i=1}^{k}$ are equal to $0$. On the other hand, nearly half of the values $\{\Id_{\inner{w_4, z_i+\eta_i} + b_4}\}_{i=1}^{k}$ become $1$, which significantly increases the denominator of Eq.~\eqref{eq:cnn1_grad_alignment}, and thus makes the cosine close to 0.
	At the same time, the output of a regular filter $w_1$ shown in Fig.~\ref{fig:cnn4_fms} is only slightly affected by the random noise $\eta$. 
	For deep networks, however, we could not identify \textit{particular} filters responsible for catastrophic overfitting, thus we consider next a more general solution.

	\section{Increasing gradient alignment improves fast adversarial training} \label{sec:main_exps}
	Based on the importance of gradient alignment for successful FGSM training, we propose a regularizer, \texttt{GradAlign}, that aims at increasing gradient alignment and preventing catastrophic overfitting.
	The core idea of \texttt{GradAlign} is to maximize the gradient alignment (as defined in Eq.~\eqref{eq:grad_alignment}) between the gradients at point $x$ and at a randomly perturbed point $x + \eta$ inside the $\ell_\infty$-ball around $x$:
	\begin{align} 
	\Omega(x, y, \theta) &\defeq
	\E_{(x, y) \sim D, \ \eta \sim \U([-\varepsilon, \varepsilon]^d)} \left[
	1 - \cos \left( \nabla_x \l(x, y; \theta), \nabla_x \l(x+\eta, y; \theta) \right) \right]. 
	\end{align}
	Crucially, \texttt{GradAlign} uses gradients at points $x$ and $x+\eta$ which does not require an expensive iterative procedure unlike, e.g., the LLR method of \cite{qin2019adversarial}.
	Note that the regularizer depends only on the gradient direction and it is invariant to the gradient norm which contrasts it to the gradient penalties \cite{gu2014deep,NIPS2017_6821,ross2018improving,simon2019first} or CURE \cite{moosavi2019robustness} (see the comparison in Appendix~\ref{app:alternatives_to_gradalign}).

	\myparagraph{Experimental setup.}
	We compare the following methods: 
	standard FGSM AT,
	FGSM-RS AT with $\alpha=1.25\varepsilon$ \cite{wong2020fast},
	FGSM AT + \texttt{GradAlign},
	\textit{AT for Free} with $m=8$ \cite{shafahi2019adversarial},
	PGD-2 AT with 2-step PGD using $\alpha=\nicefrac{\varepsilon}{2}$, and
	PGD-10 AT with 10-step PGD using $\alpha=\nicefrac{2\varepsilon}{10}$.
	We train these methods using PreAct ResNet-18~\cite{he2016identity} with $\ell_\infty$-radii $\varepsilon \in \{\nicefrac{1}{255}, \dots, \nicefrac{16}{255}\}$ on CIFAR-10 for 30 epochs and $\varepsilon \in \{\nicefrac{1}{255}, \dots, \nicefrac{12}{255}\}$ on SVHN for 15 epochs.
	The only exception is \textit{AT for Free} \cite{shafahi2019adversarial} which we train for $96$ epochs on CIFAR-10, and $45$ epochs on SVHN which was necessary to get comparable results to the other methods.
	Unlike \cite{qin2019adversarial} and \cite{zhang2019propagate}, with the training scheme of \cite{wong2020fast} and $\alpha=\nicefrac{\varepsilon}{2}$ we could successfully train a PGD-2 model with $\varepsilon=\nicefrac{8}{255}$ on CIFAR-10 with robustness better than that of their methods that use the same number of PGD steps (see Appendix~\ref{app:additional_exps_particular_linf_eps}). 
	This also echoes the recent finding of~\cite{rice2020overfitting} that properly tuned multi-step PGD AT outperforms more recently published methods. 
	As before, we evaluate robustness using PGD-50-10 with $50$ iterations and $10$ restarts using step size $\alpha=\nicefrac{\varepsilon}{4}$ following \cite{wong2020fast} for the same $\varepsilon$ that was used for training. We train each model with $5$ random seeds since the final robustness can have a large variance for high $\varepsilon$.
	Also, we remark that training with \texttt{GradAlign} leads on average to a $3\times$ slowdown on an NVIDA V100 GPU compared to FGSM training which is due to the use of double backpropagation (see \cite{etmann2019closer} for a detailed analysis). 
	We think that improving the runtime of \texttt{GradAlign} is possible, but we postpone it to future work.
	Additional implementation details are provided in Appendix~\ref{app:exp_details}.
	The code of our experiments is available at \url{https://github.com/tml-epfl/understanding-fast-adv-training}.
	
	\begin{figure}[b]
		\centering
		\includegraphics[width=0.48\textwidth]{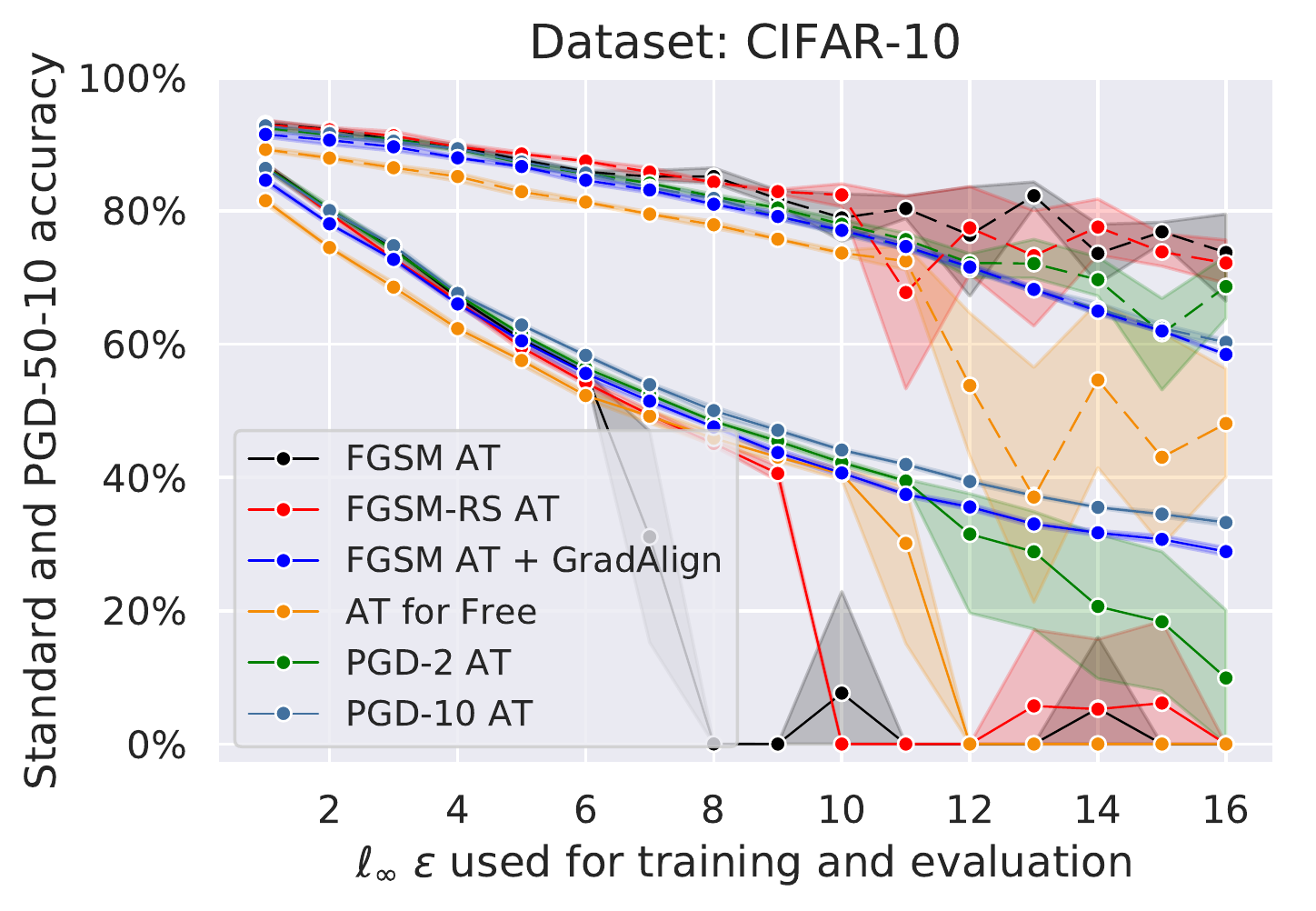} \hspace{2mm}
		\includegraphics[width=0.48\textwidth]{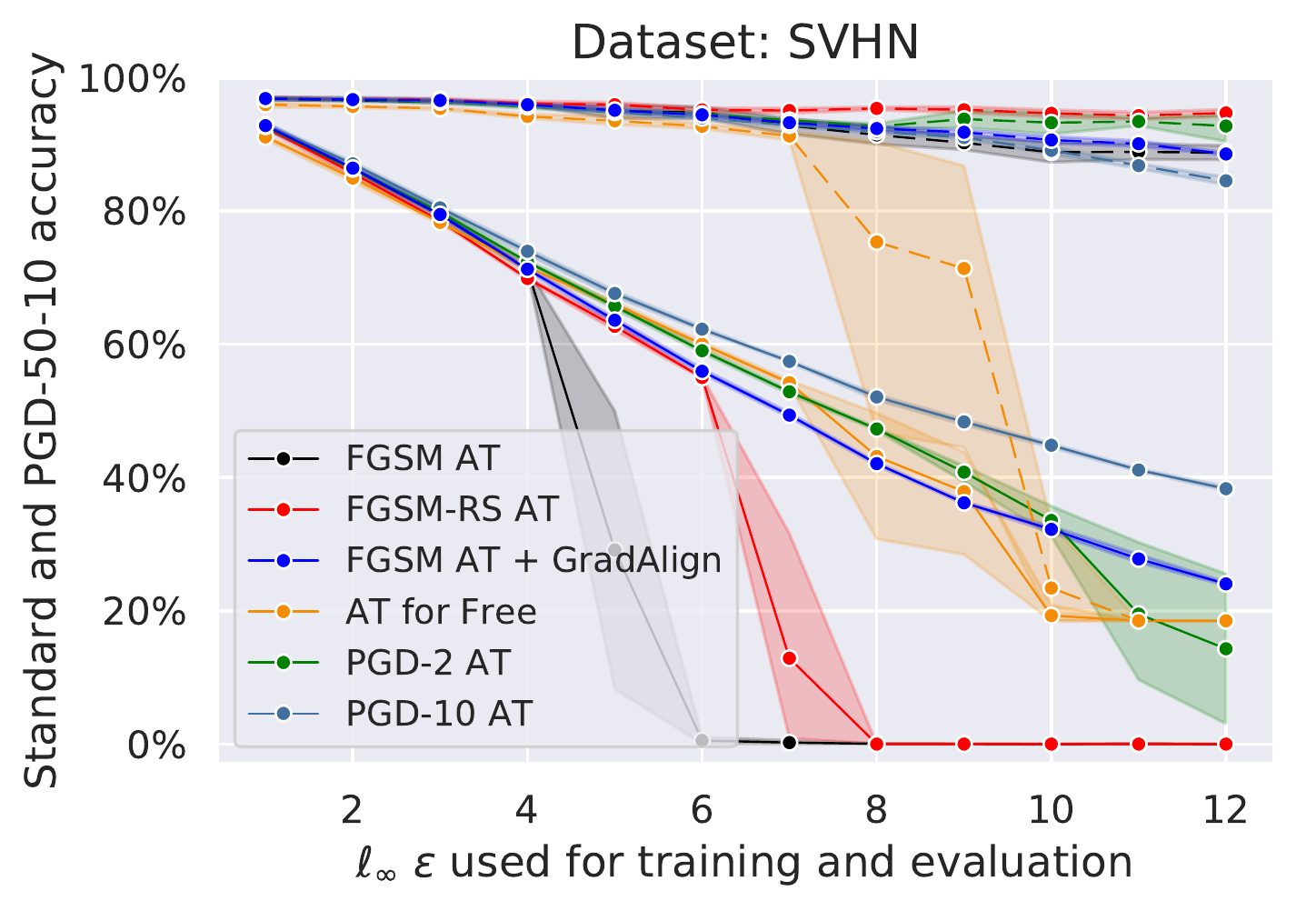}
		\caption{Accuracy (dashed line) and robustness (solid line) of different adversarial training (AT) methods on CIFAR-10 and SVHN with ResNet-18 trained and evaluated with different $l_\infty$-radii. The results are obtained without early stopping, averaged over 5 random seeds used for training and reported with the standard deviation.}
		\label{fig:main_exps}
	\end{figure}
	\myparagraph{Results on CIFAR-10 and SVHN.}
	We provide the main comparison in Fig.~\ref{fig:main_exps} and provide detailed numbers for specific values of $\varepsilon$ in Appendix~\ref{app:additional_exps_particular_linf_eps} which also includes an additional evaluation of our models with \textit{AutoAttack} \cite{croce2020reliable}.
	First, we notice that all the methods perform almost equally well for small enough~$\varepsilon$, i.e. $\varepsilon \leq \nicefrac{6}{255}$ on CIFAR-10 and $\varepsilon \leq \nicefrac{4}{255}$ on SVHN. However, the performance for larger $\varepsilon$ varies a lot depending on the method due to catastrophic overfitting.
	Importantly, \texttt{GradAlign} \textit{succesfully prevents catastrophic overfitting} in FGSM AT, thus allowing to successfully apply FGSM training also for larger $\ell_\infty$-perturbations and reduce the gap to PGD-10 training. In Appendix~\ref{app:ablation_studies}, we additionally show that FGSM~+~\texttt{GradAlign} does not suffer from catastrophic overfitting even for $\varepsilon \in \{\nicefrac{24}{255}, \nicefrac{32}{255}\}$.
	At the same time, \textit{not only} FGSM AT and FGSM-RS AT experience catastrophic overfitting, but also the recently proposed \textit{AT for Free} and PGD-2, although at higher $\varepsilon$ values than FGSM AT.
	We note that \texttt{GradAlign} is not only applicable to FGSM AT, but also to other methods that can also suffer from catastrophic overfitting. In particular, combining PGD-2 with \texttt{GradAlign} prevents catastrophic overfitting and leads to better robustness for $\varepsilon=\nicefrac{16}{255}$ on CIFAR-10 (see Appendix~\ref{app:additional_exps_particular_linf_eps}).
	Although performing early stopping can lead to non-trivial robustness, standard accuracy is often significantly sacrificed which \textit{limits the usefulness of early stopping} as we show in Appendix~\ref{app:results_with_es}.
	This is in contrast to training with \texttt{GradAlign} which leads to the same standard accuracy as PGD-10 AT. 
	
	\myparagraph{Results on ImageNet.}
	We also performed similar experiments on ImageNet in Appendix~\ref{app:additional_exps_particular_linf_eps} to illustrate that \texttt{GradAlign} can be scaled to large-scale problems despite the slowdown.
	However, we observed that even for standard FGSM training using the training schedule of \cite{wong2020fast}, catastrophic overfitting \textit{does not} occur for $\varepsilon \in \{\nicefrac{2}{255}, \nicefrac{4}{255}\}$ considered in \cite{shafahi2019adversarial,wong2020fast}, and thus there is no need to use \texttt{GradAlign} as its main role is to prevent catastrophic overfitting.
	We observe that for these $\varepsilon$ values, the gradient alignment evolves similarly to that of PGD AT from the CIFAR-10 experiments shown in Fig.~\ref{fig:training_curves_resnet}, i.e. it decreases gradually over epochs but \textit{without} a sharp drop that would indicate catastrophic overfitting.
	For $\varepsilon = \nicefrac{6}{255}$, we observe that the gradient alignment and PGD accuracy for FGSM-RS drop very early in training (after 3 epochs), but not for FGSM or FGSM~+~\texttt{GradAlign} training. This contradicts our observations on CIFAR-10 and SVHN where we observed that FGSM-RS usually helps to postpone catastrophic overfitting to higher $\varepsilon$. However, it is computationally demanding to replicate the results on ImageNet over different random seeds as we did for CIFAR-10 and SVHN. Thus, we leave a more detailed investigation of catastrophic overfitting on ImageNet for future work.
	
	\myparagraph{Robust vs. catastrophic overfitting.}
	Recently, \citet{rice2020overfitting} brought up the importance of early stopping in adversarial training to mitigate the \textit{robust overfitting} phenomenon that is characterized by a decreasing trend of test adversarial accuracy over training iterations. It is thus a natural question to ask whether robust and catastrophic overfitting are related, and whether \texttt{GradAlign} can be beneficial to mitigate robust overfitting. We observed that training FGSM~+~\texttt{GradAlign} for more than $30$ epochs also leads to slightly worse robustness on the test set (see Appendix~\ref{app:ablation_studies}), thus suggesting that \textit{catastrophic} and \textit{robust overfitting} are two distinct phenomena that have to be addressed separately. As a sidenote, we also observe that FGSM training combined with \texttt{GradAlign} does not lead to catastrophic overfitting even when trained \textit{up to 200 epochs}.

	\section{Conclusions and outlook}
	We observed that catastrophic overfitting is a fundamental problem not only for standard FGSM training, but for computationally efficient adversarial training in general.
	In particular, many recently proposed schemes such as FGSM AT enhanced by a random step or \textit{AT for free} are also prone to catastrophic overfitting, and simply using early stopping leads to suboptimal models. 
	Motivated by this, we explored the questions of \textit{when} and \textit{why} FGSM adversarial training works, and how to improve it by increasing the gradient alignment, and thus the quality of the solution of the inner maximization problem. 
	Our proposed regularizer, \texttt{GradAlign}, prevents catastrophic overfitting and improves the robustness compared to other fast adversarial training methods reducing the gap to multi-step PGD training. 
	However, \texttt{GradAlign} leads to an increased runtime due to the use of double backpropagation. We hope that the same effect of stabilizing the gradients under \textit{random} noise can be achieved in future work with other regularization methods that do not rely on double backpropagation.

	\begin{ack}
		We thank Eric Wong, Francesco Croce, and Chen Liu for many fruitful discussions. We are also very grateful to Guillermo Ortiz-Jimenez, Apostolos Modas, Ludwig Schmidt, and anonymous reviewers for the useful feedback on the paper.
	\end{ack}

	\section*{Broader Impact}
	Our work focuses on a systematic study of the failure reasons behind computationally efficient adversarial training methods. We suggest a new regularization approach which helps to overcome the shortcoming of the existing methods that is known as \textit{catastrophic overfitting}.
	
	We see primarily positive outcomes from our work since adversarial robustness is a desirable property that improves the \textit{reliability} of machine learning models. Therefore, it is crucial to be able to train robust models efficiently and without limiting efficient training only to perturbations of a small size.

	\bibliographystyle{plainnat}
	\bibliography{references}

\begin{thebibliography}{51}
\providecommand{\natexlab}[1]{#1}
\providecommand{\url}[1]{\texttt{#1}}
\expandafter\ifx\csname urlstyle\endcsname\relax
  \providecommand{\doi}[1]{doi: #1}\else
  \providecommand{\doi}{doi: \begingroup \urlstyle{rm}\Url}\fi

\bibitem[Alayrac et~al.(2019)Alayrac, Uesato, Huang, Stanforth, Fawzi, and
  Kohli]{alayrac2019arelabels}
Jean{-}Baptiste Alayrac, Jonathan Uesato, Po{-}Sen Huang, Robert Stanforth,
  Alhussein Fawzi, and Pushmeet Kohli.
\newblock Are labels required for improving adversarial robustness?
\newblock \emph{NeurIPS}, 2019.

\bibitem[Ben-Tal et~al.(2009)Ben-Tal, El~Ghaoui, and Nemirovski]{MR2546839}
Aharon Ben-Tal, Laurent El~Ghaoui, and Arkadi Nemirovski.
\newblock \emph{Robust optimization}.
\newblock Princeton Series in Applied Mathematics. Princeton University Press,
  Princeton, NJ, 2009.

\bibitem[Biggio and Roli(2018)]{biggio2018wild}
Battista Biggio and Fabio Roli.
\newblock Wild patterns: ten years after the rise of adversarial machine
  learning.
\newblock \emph{Pattern Recognition}, 2018.

\bibitem[Bochkovskiy et~al.(2020)Bochkovskiy, Wang, and
  Liao]{bochkovskiy2020yolov4}
Alexey Bochkovskiy, Chien-Yao Wang, and Hong-Yuan~Mark Liao.
\newblock Yolov4: Optimal speed and accuracy of object detection.
\newblock \emph{arXiv preprint aXiv:2004.10934}, 2020.

\bibitem[Carlini et~al.(2017)Carlini, Katz, Barrett, and
  Dill]{carlini2017provably}
Nicholas Carlini, Guy Katz, Clark Barrett, and David~L Dill.
\newblock Provably minimally-distorted adversarial examples.
\newblock \emph{arXiv preprint arXiv:1709.10207}, 2017.

\bibitem[Carmon et~al.(2019)Carmon, Raghunathan, Schmidt, Liang, and
  Duchi]{carmon2019unlabeled}
Yair Carmon, Aditi Raghunathan, Ludwig Schmidt, Percy Liang, and John~C. Duchi.
\newblock Unlabeled data improves adversarial robustness.
\newblock \emph{NeurIPS}, 2019.

\bibitem[Cohen et~al.(2019)Cohen, Rosenfeld, and Kolter]{cohen2019certified}
Jeremy~M Cohen, Elan Rosenfeld, and J~Zico Kolter.
\newblock Certified adversarial robustness via randomized smoothing.
\newblock \emph{ICML}, 2019.

\bibitem[Croce and Hein(2020)]{croce2020reliable}
Francesco Croce and Matthias Hein.
\newblock Reliable evaluation of adversarial robustness with an ensemble of
  diverse parameter-free attacks.
\newblock \emph{ICML}, 2020.

\bibitem[Etmann(2019)]{etmann2019closer}
Christian Etmann.
\newblock A closer look at double backpropagation.
\newblock \emph{arXiv preprint ArXiv:1906.06637}, 2019.

\bibitem[Globerson and Roweis(2006)]{globerson2006nightmare}
Amir Globerson and Sam Roweis.
\newblock Nightmare at test time: robust learning by feature deletion.
\newblock \emph{ICML}, 2006.

\bibitem[Gonzales and Woods(2002)]{gonzales2002digital}
Rafael~C Gonzales and Richard~E Woods.
\newblock \emph{Digital image processing (2nd edition)}.
\newblock Prentice Hall New Jersey, 2002.

\bibitem[Goodfellow et~al.(2015)Goodfellow, Shlens, and
  Szegedy]{goodfellow2014explaining}
Ian~J Goodfellow, Jonathon Shlens, and Christian Szegedy.
\newblock Explaining and harnessing adversarial examples.
\newblock \emph{ICLR}, 2015.

\bibitem[Gowal et~al.(2019)Gowal, Uesato, Qin, Huang, Mann, and
  Kohli]{gowal2019alternative}
Sven Gowal, Jonathan Uesato, Chongli Qin, Po-Sen Huang, Timothy Mann, and
  Pushmeet Kohli.
\newblock An alternative surrogate loss for pgd-based adversarial testing.
\newblock \emph{arXiv preprint arXiv:1910.09338}, 2019.

\bibitem[Gu and Rigazio(2015)]{gu2014deep}
Shixiang Gu and Luca Rigazio.
\newblock Towards deep neural network architectures robust to adversarial
  examples.
\newblock \emph{ICLR workshops}, 2015.

\bibitem[He et~al.(2015)He, Zhang, Ren, and Sun]{he2015delving}
Kaiming He, Xiangyu Zhang, Shaoqing Ren, and Jian Sun.
\newblock Delving deep into rectifiers: Surpassing human-level performance on
  imagenet classification.
\newblock \emph{ICCV}, 2015.

\bibitem[He et~al.(2016)He, Zhang, Ren, and Sun]{he2016identity}
Kaiming He, Xiangyu Zhang, Shaoqing Ren, and Jian Sun.
\newblock Identity mappings in deep residual networks.
\newblock \emph{ECCV}, 2016.

\bibitem[Hein and Andriushchenko(2017)]{NIPS2017_6821}
Matthias Hein and Maksym Andriushchenko.
\newblock Formal guarantees on the robustness of a classifier against
  adversarial manipulation.
\newblock \emph{NeurIPS}, 2017.

\bibitem[Hendrycks et~al.(2019)Hendrycks, Lee, and Mazeika]{hendrycks2019using}
Dan Hendrycks, Kimin Lee, and Mantas Mazeika.
\newblock Using pre-training can improve model robustness and uncertainty.
\newblock \emph{ICML}, 2019.

\bibitem[Huber(1981)]{MR606374}
Peter~J. Huber.
\newblock \emph{Robust statistics}.
\newblock John Wiley \& Sons, Inc., New York, 1981.

\bibitem[Juditsky et~al.(2011)Juditsky, Nemirovski, and
  Tauvel]{juditsky2011solving}
Anatoli Juditsky, Arkadi Nemirovski, and Claire Tauvel.
\newblock Solving variational inequalities with stochastic mirror-prox
  algorithm.
\newblock \emph{Stochastic Systems}, 2011.

\bibitem[Katz et~al.(2017)Katz, Barrett, Dill, Julian, and
  Kochenderfer]{katz2017reluplex}
Guy Katz, Clark Barrett, David~L Dill, Kyle Julian, and Mykel~J Kochenderfer.
\newblock Reluplex: an efficient smt solver for verifying deep neural networks.
\newblock \emph{ICCAV}, 2017.

\bibitem[LeCun et~al.(2015)LeCun, Bengio, and Hinton]{lecun2015deep}
Yann LeCun, Yoshua Bengio, and Geoffrey Hinton.
\newblock Deep learning.
\newblock \emph{Nature}, 521\penalty0 (7553), 2015.

\bibitem[Madry et~al.(2018)Madry, Makelov, Schmidt, Tsipras, and
  Vladu]{madry2018towards}
Aleksander Madry, Aleksandar Makelov, Ludwig Schmidt, Dimitris Tsipras, and
  Adrian Vladu.
\newblock Towards deep learning models resistant to adversarial attacks.
\newblock \emph{ICLR}, 2018.

\bibitem[Micikevicius et~al.(2018)Micikevicius, Narang, Alben, Diamos, Elsen,
  Garcia, Ginsburg, Houston, Kuchaiev, Venkatesh,
  et~al.]{micikevicius2018mixed}
Paulius Micikevicius, Sharan Narang, Jonah Alben, Gregory Diamos, Erich Elsen,
  David Garcia, Boris Ginsburg, Michael Houston, Oleksii Kuchaiev, Ganesh
  Venkatesh, et~al.
\newblock Mixed precision training.
\newblock \emph{ICLR}, 2018.

\bibitem[Moosavi-Dezfooli et~al.(2019)Moosavi-Dezfooli, Fawzi, Uesato, and
  Frossard]{moosavi2019robustness}
Seyed-Mohsen Moosavi-Dezfooli, Alhussein Fawzi, Jonathan Uesato, and Pascal
  Frossard.
\newblock Robustness via curvature regularization, and vice versa.
\newblock \emph{CVPR}, 2019.

\bibitem[Nakkiran et~al.(2019)Nakkiran, Kaplun, Kalimeris, Yang, Edelman,
  Zhang, and Barak]{Nakkiran2019SGD}
Preetum Nakkiran, Gal Kaplun, Dimitris Kalimeris, Tristan Yang, Benjamin~L.
  Edelman, Fred Zhang, and Boaz Barak.
\newblock {SGD} on neural networks learns functions of increasing complexity.
\newblock \emph{NeurIPS}, 2019.

\bibitem[Papernot et~al.(2017)Papernot, McDaniel, Goodfellow, Jha, Celik, and
  Swami]{papernot2017practical}
Nicolas Papernot, Patrick McDaniel, Ian Goodfellow, Somesh Jha, Z~Berkay Celik,
  and Ananthram Swami.
\newblock Practical black-box attacks against machine learning.
\newblock \emph{ASIA CCS'17}, 2017.

\bibitem[Qin et~al.(2019)Qin, Martens, Gowal, Krishnan, Dvijotham, Fawzi, De,
  Stanforth, and Kohli]{qin2019adversarial}
Chongli Qin, James Martens, Sven Gowal, Dilip Krishnan, Krishnamurthy
  Dvijotham, Alhussein Fawzi, Soham De, Robert Stanforth, and Pushmeet Kohli.
\newblock Adversarial robustness through local linearization.
\newblock \emph{NeurIPS}, 2019.

\bibitem[Raghunathan et~al.(2018)Raghunathan, Steinhardt, and
  Liang]{raghunathan2018certified}
Aditi Raghunathan, Jacob Steinhardt, and Percy Liang.
\newblock Certified defenses against adversarial examples.
\newblock \emph{ICLR}, 2018.

\bibitem[Rice et~al.(2020)Rice, Wong, and Kolter]{rice2020overfitting}
Leslie Rice, Eric Wong, and J.~Zico Kolter.
\newblock Overfitting in adversarially robust deep learning.
\newblock \emph{ICML}, 2020.

\bibitem[Ross and Doshi-Velez(2018)]{ross2018improving}
Andrew~Slavin Ross and Finale Doshi-Velez.
\newblock Improving the adversarial robustness and interpretability of deep
  neural networks by regularizing their input gradients.
\newblock \emph{AAAI}, 2018.

\bibitem[Santurkar et~al.(2019)Santurkar, Tsipras, Tran, Ilyas, Engstrom, and
  Madry]{santurkar2019image}
Shibani Santurkar, Dimitris Tsipras, Brandon Tran, Andrew Ilyas, Logan
  Engstrom, and Aleksander Madry.
\newblock Image synthesis with a single (robust) classifier.
\newblock \emph{NeurIPS}, 2019.

\bibitem[Schmidhuber(2015)]{schmidhuber2015deep}
J{\"u}rgen Schmidhuber.
\newblock Deep learning in neural networks: An overview.
\newblock \emph{Neural networks}, 61:\penalty0 85--117, 2015.

\bibitem[Shafahi et~al.(2019)Shafahi, Najibi, Ghiasi, Xu, Dickerson, Studer,
  Davis, Taylor, and Goldstein]{shafahi2019adversarial}
Ali Shafahi, Mahyar Najibi, Amin Ghiasi, Zheng Xu, John Dickerson, Christoph
  Studer, Larry~S. Davis, Gavin Taylor, and Tom Goldstein.
\newblock Adversarial training for free!
\newblock \emph{NeurIPS}, 2019.

\bibitem[Shaham et~al.(2018)Shaham, Yamada, and
  Negahban]{shaham2015understanding}
Uri Shaham, Yutaro Yamada, and Sahand Negahban.
\newblock Understanding adversarial training: Increasing local stability of
  supervised models through robust optimization.
\newblock \emph{Neurocomputing}, 2018.

\bibitem[Simon-Gabriel et~al.(2019)Simon-Gabriel, Ollivier, Bottou,
  Sch{\"o}lkopf, and Lopez-Paz]{simon2019first}
Carl-Johann Simon-Gabriel, Yann Ollivier, Leon Bottou, Bernhard Sch{\"o}lkopf,
  and David Lopez-Paz.
\newblock First-order adversarial vulnerability of neural networks and input
  dimension.
\newblock \emph{ICML}, 2019.

\bibitem[Smith(2017)]{smith2017cyclical}
Leslie~N Smith.
\newblock Cyclical learning rates for training neural networks.
\newblock \emph{WACV}, 2017.

\bibitem[Szegedy et~al.(2014)Szegedy, Zaremba, Sutskever, Bruna, Erhan,
  Goodfellow, and Fergus]{szegedy2013intriguing}
Christian Szegedy, Wojciech Zaremba, Ilya Sutskever, Joan Bruna, Dumitru Erhan,
  Ian Goodfellow, and Rob Fergus.
\newblock Intriguing properties of neural networks.
\newblock \emph{ICLR}, 2014.

\bibitem[Taori et~al.(2020)Taori, Dave, Shankar, Carlini, Recht, and
  Schmidt]{taori2020robustness}
Rohan Taori, Achal Dave, Vaishaal Shankar, Nicholas Carlini, Benjamin Recht,
  and Ludwig Schmidt.
\newblock Measuring robustness to natural distribution shifts in image
  classification.
\newblock \emph{NeurIPS}, 2020.

\bibitem[Tjeng et~al.(2019)Tjeng, Xiao, and Tedrake]{tjeng2017evaluating}
Vincent Tjeng, Kai Xiao, and Russ Tedrake.
\newblock Evaluating robustness of neural networks with mixed integer
  programming.
\newblock \emph{ICLR}, 2019.

\bibitem[Tramèr et~al.(2018)Tramèr, Kurakin, Papernot, Goodfellow, Boneh, and
  McDaniel]{tramer2018ensemble}
Florian Tramèr, Alexey Kurakin, Nicolas Papernot, Ian Goodfellow, Dan Boneh,
  and Patrick McDaniel.
\newblock Ensemble adversarial training: Attacks and defenses.
\newblock \emph{ICLR}, 2018.

\bibitem[Tsipras et~al.(2019)Tsipras, Santurkar, Engstrom, Turner, and
  Madry]{tsipras2018robustness}
Dimitris Tsipras, Shibani Santurkar, Logan Engstrom, Alexander Turner, and
  Aleksander Madry.
\newblock Robustness may be at odds with accuracy.
\newblock \emph{ICLR}, 2019.

\bibitem[Vivek and Babu(2020)]{babu2020single}
B.~S. Vivek and R.~Venkatesh Babu.
\newblock Single-step adversarial training with dropout scheduling.
\newblock \emph{CVPR}, 2020.

\bibitem[Wang et~al.(2019)Wang, Ma, Bailey, Yi, Zhou, and
  Gu]{wang2019convergence}
Yisen Wang, Xingjun Ma, James Bailey, Jinfeng Yi, Bowen Zhou, and Quanquan Gu.
\newblock On the convergence and robustness of adversarial training.
\newblock \emph{ICML}, 2019.

\bibitem[Weng et~al.(2018)Weng, Zhang, Chen, Song, Hsieh, Boning, Dhillon, and
  Daniel]{weng2018towards}
Tsui-Wei Weng, Huan Zhang, Hongge Chen, Zhao Song, Cho-Jui Hsieh, Duane Boning,
  Inderjit~S. Dhillon, and Luca Daniel.
\newblock Towards fast computation of certified robustness for relu networks.
\newblock \emph{ICML}, 2018.

\bibitem[Wong and Kolter(2018)]{wong2017provable}
Eric Wong and Zico Kolter.
\newblock Provable defenses against adversarial examples via the convex outer
  adversarial polytope.
\newblock \emph{ICML}, 2018.

\bibitem[Wong et~al.(2020)Wong, Rice, and Kolter]{wong2020fast}
Eric Wong, Leslie Rice, and J.~Zico Kolter.
\newblock Fast is better than free: Revisiting adversarial training.
\newblock \emph{ICLR}, 2020.

\bibitem[Xie et~al.(2020)Xie, Tan, Gong, Wang, Yuille, and
  Le]{xie2019adversarial}
Cihang Xie, Mingxing Tan, Boqing Gong, Jiang Wang, Alan Yuille, and Quoc~V Le.
\newblock Adversarial examples improve image recognition.
\newblock \emph{CVPR}, 2020.

\bibitem[Zhang et~al.(2019{\natexlab{a}})Zhang, Zhang, Lu, Zhu, and
  Dong]{zhang2019propagate}
Dinghuai Zhang, Tianyuan Zhang, Yiping Lu, Zhanxing Zhu, and Bin Dong.
\newblock You only propagate once: Accelerating adversarial training via
  maximal principle.
\newblock \emph{NeurIPS}, 2019{\natexlab{a}}.

\bibitem[Zhang et~al.(2019{\natexlab{b}})Zhang, Yu, Jiao, Xing, Ghaoui, and
  Jordan]{zhang19theoretically}
Hongyang Zhang, Yaodong Yu, Jiantao Jiao, Eric Xing, Laurent~El Ghaoui, and
  Michael Jordan.
\newblock Theoretically principled trade-off between robustness and accuracy.
\newblock \emph{ICML}, 2019{\natexlab{b}}.

\bibitem[Zhu et~al.(2019)Zhu, Cheng, Gan, Sun, Goldstein, and
  Liu]{zhu2019freelb}
Chen Zhu, Yu~Cheng, Zhe Gan, Siqi Sun, Tom Goldstein, and Jingjing Liu.
\newblock Freelb: Enhanced adversarial training for natural language
  understanding.
\newblock \emph{ICLR}, 2019.

\end{thebibliography}
	\clearpage

	\appendix

	\begin{center}
		\Large\textbf{Appendix}
	\end{center}

	\section{Deferred proofs}
	In this section, we show the proofs omitted from Sec.~\ref{sec:role_limitation_of_rs} and Sec.~\ref{sec:understanding_co_via_ga}.
	
	\subsection{Proof of Lemma~\ref{lem:norm_fgsm_rs}} \label{sec:proof_norm_fgsm_rs}
	We state again Lemma~\ref{lem:norm_fgsm_rs} from Sec.~\ref{sec:role_limitation_of_rs} and present the proof.
	
	\begin{customlemma}{\ref{lem:norm_fgsm_rs}}{\normalfont \textbf{(Effect of the random step)}} \nonumber
		Let $\eta \sim \U([-\varepsilon, \varepsilon]^d)$ be a random starting point, and  $\alpha \in [0, 2\varepsilon]$ be the step size of FGSM-RS defined in Eq.~\eqref{eq:fgsmrs-def}, then
		\begin{align*}
		\E_\eta \left[ \norm{\delta_{FGSM-RS}(\eta)}_2 \right] \leq 
		\sqrt{\E_\eta \left[ \norm{\delta_{FGSM-RS}(\eta)}_2^2 \right]} =
		\sqrt{d} \sqrt{-\frac{1}{6\varepsilon} \alpha^3 + \frac{1}{2} \alpha^2 + \frac{1}{3} \varepsilon^2}.
		\end{align*}
	\end{customlemma}
	\begin{proof}
		First, note that due to the Jensen's inequality, we can have a convenient upper bound which is easier to work with:
		\begin{align} \label{eq:jensen}
		\E \left[ \norm{\delta_{FGSM-RS}(\eta)}_2 \right] \leq \sqrt{ \E \left[ \norm{\delta_{FGSM-RS}(\eta)}_2^2 \right]}.
		\end{align}
		Therefore, we can focus on $\E \left[ \norm{\delta_{FGSM-RS}}_2^2 \right]$ which can be computed analytically. Let us denote by $\nabla \defeq \nabla_x \l(x+\eta, y; \theta) \in \R^d$, we then obtain:
		\begin{align*} 
		\E_\eta \left[ \norm{\delta_{FGSM-RS}}_2^2 \right] &= 
		\E_\eta \left[ \norm{\Pi_{[-\varepsilon, \varepsilon]} \left[ \eta + \alpha \sign(\nabla) \right]}_2^2  \right] =
		\sum_{i=1}^d  \E_{\eta_i} \left[ \Pi_{[-\varepsilon, \varepsilon]} \left[ \eta_i + \alpha \sign(\nabla_i) \right]^2 \right] \\ &= 
		d \E_{\eta_i} \left[ \min \{\varepsilon, |\eta_i + \alpha \sign(\nabla_i)|\}^2 \right] = 
		d \E_{\eta_i} \left[ \min \{\varepsilon^2, \left(\eta_i + \alpha \sign(\nabla_i)\right)^2\} \right] \\ &=
		d \E_{r_i} \left[ \E_{\eta_i} \left[ \min \{\varepsilon^2, \left(\eta_i + \alpha \sign(\nabla_i)\right)^2\} \ | \ \sign(\nabla_i) = r_i \right] \right],
		\end{align*}
		where in the last step we use the law of total expectation by noting that $\sign(\nabla_i)$ is also a random variable since it depends on $\eta_i$.
		
		We first consider the case when $\sign(\nabla_i) = 1$, then the inner conditional expectation is equal to:
		\begin{align*} 
		\int_{-\varepsilon}^{\varepsilon} \min \{ \varepsilon^2, \left( \eta_i + \alpha \right)^2 \} \frac{1}{2\varepsilon} d\eta_i &= 
		\frac{1}{2\varepsilon} \int_{-\varepsilon + \alpha}^{\varepsilon + \alpha} \min \{\varepsilon^2, x^2\} dx \\ \nonumber &= 
		\frac{1}{2\varepsilon} \left( \int_{\varepsilon}^{\varepsilon+\alpha} \varepsilon^2 dx  +  \int_{-\varepsilon+\alpha}^{\varepsilon} x^2 dx \right) \\ &=
		-\frac{1}{6\varepsilon} \alpha^3 + \frac{1}{2} \alpha^2 + \frac{1}{3} \varepsilon^2.
		\end{align*}
		The case when $\sign(\nabla_i) = -1$ leads to the same expression: 
		\begin{align*} 
		\int_{-\varepsilon}^{\varepsilon} \min \{ \varepsilon^2, \left( \eta_i - \alpha \right)^2 \} \frac{1}{2\varepsilon} d\eta_i &= 
		\frac{1}{2\varepsilon} \int_{-\varepsilon - \alpha}^{\varepsilon - \alpha} \min \{\varepsilon^2, x^2\} dx = 
		-\frac{1}{6\varepsilon} \alpha^3 + \frac{1}{2} \alpha^2 + \frac{1}{3} \varepsilon^2.
		\end{align*}
		
		Combining these two cases together with Eq.~\eqref{eq:jensen}, we have that:
		\begin{align*}
		\E_\eta \left[ \norm{\delta_{FGSM-RS}(\eta)}_2 \right] \leq  \sqrt{\E \left[ \norm{\delta_{FGSM-RS}(\eta)}_2^2 \right]} = \sqrt{d} \sqrt{ -\frac{1}{6\varepsilon} \alpha^3 + \frac{1}{2} \alpha^2 + \frac{1}{3} \varepsilon^2 }.
		\end{align*}
	\end{proof}

	\subsection{Proof and discussion of Lemma~\ref{lem:grad_alignment_at_init}} 
	\label{sec:grad_alignment_at_init}
	We state again Lemma~\ref{lem:grad_alignment_at_init} from Sec.~\ref{sec:understanding_co_via_ga} and present the proof.
	
	\begin{customlemma}{\ref{lem:grad_alignment_at_init}}{\normalfont \textbf{(Gradient alignment at initialization)}} 
		Let $z \sim \U([0,1]^p)$ be an image patch for $p \geq 2$, $\eta \sim \U([-\varepsilon, \varepsilon]^d)$ a point inside the $\ell_\infty$-ball,
		the parameters of a single-layer CNN initialized i.i.d. as $w \sim \mathcal{N}(0, \sigma_w^2 I_p)$ for every column of $W$, $u \sim \mathcal{N}(0, \sigma_u^2 I_m)$ for every column of $U$, $b:=0$,
		then the gradient alignment is lower bounded by
		\begin{align*}
		\lim_{k,m\to \infty} \cos \left( \nabla_x \l(x, y), \nabla_x \l(x+\eta, y) \right)
		\geq
		\max\left\{ 1 - \sqrt{2} \E_{w,z} \left[e^{-\frac{1}{\varepsilon^2}\inner{\nicefrac{w}{\norm{w}_2}, z}^2} \right]^{1/2}, 0.5 \right\}.
		\end{align*}
	\end{customlemma}
	\begin{proof}
		For $k$ and $m$ large enough, the law of large number ensures that an empirical mean of i.i.d. random variables can be approximated by its expectation with respect to random variables $z, \eta, w, u$.
		This leads to 
		\begin{align} \nonumber
		&\lim_{k,m\to \infty} \cos \left( \nabla_x \l(x, y), \nabla_x \l(x+\eta, y) \right) \\ \nonumber
		=&\lim_{k,m\to \infty} 
		\frac{\sum\limits_{r=1}^m \sum\limits_{l=1}^m \sum\limits_{i=1}^k \inner{w_r, w_l} u_{ri} u_{li} \Id_{\inner{w_r, z_i} \geq 0} \Id_{\inner{w_l, z_i+\eta_i} \geq 0}} 
		{\sqrt{\sum\limits_{r=1}^m \sum\limits_{l=1}^m \sum\limits_{i=1}^k \inner{w_r, w_l} u_{ri} u_{li} \Id_{\inner{w_r, z_i} \geq 0} \Id_{\inner{w_l, z_i} \geq 0}} \ \sqrt{\sum\limits_{r=1}^m \sum\limits_{l=1}^m \sum\limits_{i=1}^k \inner{w_r, w_l} u_{ri} u_{li} \Id_{\inner{w_r, z_i+\eta_i} \geq 0} \Id_{\inner{w_l, z_i+\eta_i} \geq 0}}} \\ 
		\nonumber
		=&\lim_{k,m\to \infty} 
		\frac{\frac{1}{km} \sum\limits_{r=1}^m \sum\limits_{i=1}^k \norm{w_r}_2^2 u_{ri}^2 \Id_{\inner{w_r, z_i} \geq 0} \Id_{\inner{w_r, z_i+\eta_i} \geq 0}} 
		{\sqrt{\frac{1}{km} \sum\limits_{r=1}^m \sum\limits_{i=1}^k \norm{w_r}_2^2 u_{ri}^2 \Id_{\inner{w_r, z_i} \geq 0} \Id_{\inner{w_r, z_i} \geq 0}} \ \sqrt{\frac{1}{km} \sum\limits_{r=1}^m \sum\limits_{i=1}^k \norm{w_r}_2^2 u_{ri}^2 \Id_{\inner{w_r, z_i+\eta_i} \geq 0} \Id_{\inner{w_r, z_i+\eta_i} \geq 0}}} \\ 
		\nonumber
		=&\frac{\E_{w,u,\eta,z}\left[\norm{w}_2^2 u^2 \Id_{\inner{w, z}  \geq 0} \Id_{\inner{w, z+\eta}   \geq 0}\right]}
		{\sqrt{\E_{w,u,z}\left[\norm{w}_2^2 u^2 \Id_{\inner{w, z} \geq 0}\right]}         \sqrt{\E_{w,u,\eta,z}\left[\norm{w}_2^2 u^2 \Id_{\inner{w, z+\eta}  \geq 0}\right]}} \\        
		=&\frac{\E_{w,z,\eta}\left[\norm{w}_2^2 \Id_{\inner{w, z}  \geq 0} \Id_{\inner{w, z+\eta}   \geq 0}\right]}
		{\sqrt{\E_{w,z}\left[\norm{w}_2^2 \Id_{\inner{w, z} \geq 0}\right]}              \sqrt{\E_{w,z,\eta}\left[\norm{w}_2^2 \Id_{\inner{w, z+\eta}  \geq 0}\right]}}.
		\label{eq:grad_alignment_proof_cos_decoupled}
		\end{align}
		We directly compute for the denominator:
		\begin{align*}
		\E_{w,z}[ \norm{w}_2^2 \Id_{\inner{w,z} \geq 0}] =\E_{w,\eta,z}[ \norm{w}_2^2 \Id_{\inner{w, z+\eta}  \geq 0}] = 0.5 p \sigma_w^2.
		\end{align*}
		For the numerator, by bounding $\mathbb{P}_{\eta}\left[ \inner{w, \eta} \geq \inner{w, z} \right] \leq e^{-\frac{\langle z,w\rangle^2}{2\varepsilon^2\| w\|_2^2}}$ via the Hoeffding's inequality, we obtain
		\clearpage
		
		\begin{align*}
		\E_{u,w,z,\eta} \left[\norm{w}_2^2 \Id_{\inner{w, z} \geq 0} \Id_{\inner{w, z+\eta}   \geq 0}\right]
		= &
		\E_{w,z,\eta} \left[\norm{w}_2^2 \Id_{\inner{w, z} \geq 0} \Id_{\inner{w, z+\eta}   \geq 0}\right]\\
		=&
		\E_{w,z} \left[\norm{w}_2^2 \Id_{\inner{w, z} \geq 0} \mathbb{P}_\eta \left( \inner{w, z+\eta} \geq 0 \right)\right] \\
		=&
		\E_{w,z} \left[\norm{w}_2^2 \Id_{\inner{w, z} \geq 0} \mathbb{P}_\eta \left( \inner{w, \eta} \geq -\inner{w, z} \right)\right] \\
		=&
		\E_{w,z} \left[\norm{w}_2^2 \Id_{\inner{w, z} \geq 0} \mathbb{P}_\eta \left( \inner{w, \eta} \leq \inner{w, z} \right)\right] \\
		=&
		\E_{w,z} \left[\norm{w}_2^2 \Id_{\inner{w, z} \geq 0} \left( 1 - \mathbb{P}_\eta \left( \inner{w, \eta} \geq \inner{w, z} \right) \right)\right] \\
		\geq &
		\E_{w,z} \left[\norm{w}_2^2 \Id_{\inner{w, z} \geq 0} \left( 1 - e^{-\frac{\inner{w, z}^2}{2\varepsilon^2\norm{w}_2^2}} \right)\right]\\
		=&
		\E_{w,z} \left[\norm{w}_2^2 \Id_{\inner{w, z} \geq 0}\right] - \E_{w,z} \left[\norm{w}_2^2 \Id_{\inner{w, z} \geq 0} \ e^{-\frac{\inner{w, z}^2}{2\varepsilon^2\norm{w}_2^2}} \right] \\
		=&
		0.5 p \sigma_w^2 - 0.5 \E_{w,z} \left[\norm{w}_2^2 e^{-\frac{\inner{w, z}^2}{2\varepsilon^2\norm{w}_2^2}} \right] \\
		\geq&
		0.5 p \sigma_w^2 - 0.5 \E_{w} \left[\norm{w}_2^4\right]^{1/2} \E_{w,z} \left[ e^{-\frac{\inner{w, z}^2}{\varepsilon^2\norm{w}_2^2}} \right]^{1/2} \\
		=& 
		0.5 p \sigma_w^2 - 0.5 \sigma_w^2 \sqrt{p^2 + 2p} \E_{w,z} \left[ e^{-\frac{\inner{w, z}^2}{\varepsilon^2\norm{w}_2^2}} \right]^{1/2},
		\end{align*}
		where the last inequality is obtained via the Cauchy-Schwarz inequality.
		On the other hand, we have: 
		\begin{align*}
		\E_{u,w,z,\eta} \left[\norm{w}_2^2 \Id_{\inner{w, z} \geq 0} \Id_{\inner{w, z+\eta}   \geq 0}\right]
		= &
		\E_{w,z} \left[\norm{w}_2^2 \Id_{\inner{w, z}  \geq 0} \mathbb{P}_\eta \left( \inner{w, \eta} \leq \inner{w, z} \right)\right] \\
		\geq &
		\E_{w,z} \left[\norm{w}_2^2 \Id_{\inner{w, z}  \geq 0} 0.5 \right] = 0.25 p \sigma_w^2.
		\end{align*}
		Now we combine both lower bounds together to establish a lower bound on Eq.~\eqref{eq:grad_alignment_proof_cos_decoupled}: 
		\begin{align}
		\nonumber
		&\frac{\E_{w,z,\eta}\left[\norm{w}_2^2 \Id_{\inner{w, z}  \geq 0}  \Id_{\inner{w, z+\eta}   \geq 0}\right]}
		{\sqrt{\E_{w,z}\left[\norm{w}_2^2 \Id_{\inner{w, z} \geq 0}\right]}              \sqrt{\E_{w,z,\eta}\left[\norm{w}_2^2 \Id_{\inner{w, z+\eta}  \geq 0}\right]}} \\
		\nonumber
		\geq &
		\frac{\max\left\{0.5 p \sigma_w^2 - 0.5 \sigma_w^2 \sqrt{p^2 + 2p} \E_{w,z} \left[ e^{-\frac{\inner{w, z}^2}{\varepsilon^2\norm{w}_2^2}} \right]^{1/2}, 0.25 p \sigma_w^2\right\}}
		{0.5 p \sigma_w^2} \\
		\nonumber
		= & 
		\max\left\{ 1 - \sqrt{1 + \frac{2}{p}} \E_{w,z} \left[e^{-\frac{\inner{\nicefrac{w}{\norm{w}_2}, z}^2}{\varepsilon^2}} \right]^{1/2}, 0.5 \right\} \\
		\geq &
		\max\left\{ 1 - \sqrt{2} \E_{w,z} \left[e^{-\frac{1}{\varepsilon^2}\inner{\nicefrac{w}{\norm{w}_2}, z}^2} \right]^{1/2}, 0.5 \right\},
		\label{eq:lemma_ga_final_expression}
		\end{align}
		where in the last step we used that $p \geq 2$.
	\end{proof}
	
\begin{figure}[t]
		\centering
		\small
		\includegraphics[width=0.5\columnwidth]{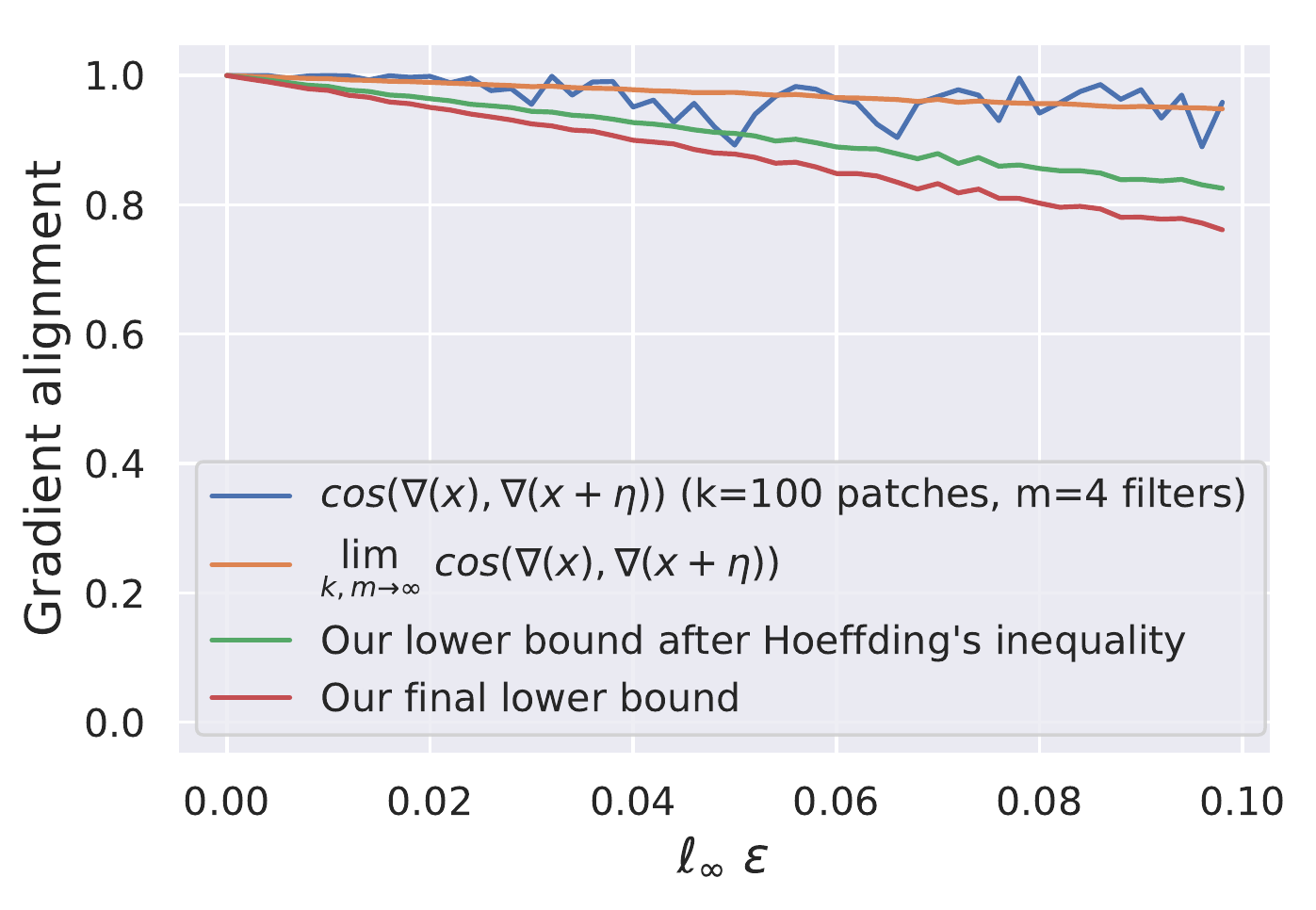} \vspace{10pt}
		\vspace{-5mm}
		\caption{Visualization of the key quantities involved in Lemma~\ref{lem:grad_alignment_at_init}.}
		\label{fig:lemma_ga_tightness}
	\end{figure}

	The main purpose of obtaining the lower bound in Lemma~\ref{lem:grad_alignment_at_init} was to get an expression that can give us an insight into the key quantities which gradient alignment at initialization depends on. Considering the limiting case $k, m \to \infty$ was necessary to obtain a ratio of expectations that allowed us to derive a simpler expression. Finally, we lower bounded the gradient alignment from Eq.~\eqref{eq:grad_alignment_proof_cos_decoupled} using the Hoeffding's and Cauchy-Schwarz inequalities and used $p\geq2$ to obtain a dimension-independent constant in front of the expectation in Eq.~\eqref{eq:lemma_ga_final_expression}.
	Now we would like to provide a better understanding about the key quantities involved in the lemma and to assess the tightness of the derived lower bound.
	For this purpose, in Fig.~\ref{fig:lemma_ga_tightness} we plot: 
	\begin{itemize}
		\item $\cos \left( \nabla_x \l(x, y), \nabla_x \l(x+\eta, y) \right)$ for $k=100$ patches and $m=4$ filters (which resembles the setting of the 4-filter CNN on CIFAR-10). We note that it is a random variable since it is a function of random variables $x, \eta, W, U$.
		\item $\lim_{k,m\to \infty} \cos \left( \nabla_x \l(x, y), \nabla_x \l(x+\eta, y) \right)$ evaluated via Eq.~\eqref{eq:grad_alignment_proof_cos_decoupled}.
		\item Our first lower bound $\max\left\{ 1 - \frac{1}{p \sigma_w^2} \E_{w,z} \left[\norm{w}_2^2 e^{-\frac{1}{2\varepsilon^2}\inner{\nicefrac{w}{\norm{w}_2}, z}^2} \right], 0.5 \right\}$ obtained via Hoeffding's inequality.
		\item Our final lower bound $\max\left\{ 1 - \sqrt{2} \E_{w,z} \left[e^{-\frac{1}{\varepsilon^2}\inner{\nicefrac{w}{\norm{w}_2}, z}^2} \right]^{1/2}, 0.5 \right\}$.
	\end{itemize}
	For the last three quantities we approximate the expectations by Monte-Carlo sampling by using $1{,}000$ samples.
	For all the quantities we use patches of size $p=3 \times 3 \times 3 = 27$ as in our CIFAR-10 experiments. We plot gradient alignment values for $\varepsilon \in [0, 0.1]$ since we are interested in small $\ell_\infty$-perturbations such as, e.g., $\varepsilon = \nicefrac{8}{255} \approx 0.03$ which is a typical value used for CIFAR-10 \cite{madry2018towards}.
	First, we can observe that all the four quantities have very high values in $[0.7, 1.0]$ for $\varepsilon \in [0, 0.1]$ which is in contrast to the gradient alignment value of $0.12$ that we observe after catastrophic overfitting for $\varepsilon = \nicefrac{10}{255} \approx 0.04$. 
	Next, we observe that $\cos \left( \nabla_x \l(x, y), \nabla_x \l(x+\eta, y) \right)$ has some noticeable variance for the chosen parameters $k=100$ patches and $m=4$ filters. However, this variance is significantly reduced when we increase the parameters $k$ and $m$, especially when considering the limiting case $k,m \to \infty$.
	Finally, we observe that both lower bounds on $\lim_{k,m\to \infty} \cos \left( \nabla_x \l(x, y), \nabla_x \l(x+\eta, y) \right)$ that we derived are empirically tight enough to properly capture the behaviour of gradient alignment for small $\varepsilon$. However, we choose to report the last one in the lemma since it is slightly more concise than the one obtained via Hoeffding's inequality.

	\section{Experimental details} \label{app:exp_details}
	We list detailed evaluation and training details below.
	
	\myparagraph{Evaluation.}
	Throughout the paper, we use PGD-50-10 for evaluation of adversarial accuracy which stands for the PGD attack with 50 iterations and 10 random restarts following \cite{wong2020fast}. We use the step size $\alpha=\nicefrac{\varepsilon}{4}$. The choice of this attack is motivated by the fact that in both public benchmarks of \cite{madry2018towards} on MNIST and CIFAR-10, the adversarial accuracy of PGD-100-50 and PGD-20-10 respectively is only 2\% away from the best entries. 
	
	Although we train our models using half precision \cite{micikevicius2018mixed}, we always perform robustness evaluation using single precision since evaluation with half precision can sometimes overestimate the robustness of the model due to limited numerical precision in the calculation of the gradients.
	
	We perform evaluation of standard accuracy using full test sets, but we evaluate adversarial accuracy using $1{,}000$ random points on each dataset.

	\myparagraph{Training details for ResNet-18.}
	We use the implementation code of \cite{wong2020fast} with the only difference that we do not use image normalization and gradient clipping on CIFAR-10 and SVHN since we found that they have no significant influence on the final results. We use cyclic learning rates and half-precision training following \cite{wong2020fast}. We do not use random initialization for PGD during adversarial training as we did not find that it leads to any improvements on the considered datasets (see the justifications in Sec.~\ref{sec:add_exps_d1} below). 
	We perform early stopping based on the PGD accuracy on the training set following \cite{wong2020fast}. We observed that such a simple model selection scheme can successfully select a model before catastrophic overfitting that has non-trivial robustness.
	
	On CIFAR-10, we train all the models for 30 epochs with the maximum learning rate $0.3$ except \textit{AT for free} \cite{shafahi2019adversarial} which we train for $96$ epochs with the maximum learning rate $0.04$ using $m=8$ minibatch replays to get comparable results to the other methods.
	
	On SVHN, we train all the models for 15 epochs with the maximum learning rate $0.05$ except \textit{AT for free} \cite{shafahi2019adversarial} which we train for $45$ epochs with the maximum learning rate $0.01$ using $m=8$ minibatch replays.
	Moreover, in order to prevent convergence to a constant classifier on SVHN, we linearly increase the perturbation radius from 0 to $\varepsilon$ during the first 5 epochs for all methods.
	
	For PGD-2 AT we use for training a 2-step PGD attack with step size $\alpha=\nicefrac{\varepsilon}{2}$, and for PGD-10 AT we use for training a 10-step PGD attack with $\alpha=\nicefrac{2\varepsilon}{10}$.
	
	For Fig.~\ref{fig:teaser_plot} and Fig.~\ref{fig:main_exps} we used the \texttt{GradAlign} $\lambda$ values obtained via a linear interpolation on the logarithmic scale between the best $\lambda$ values that we found for $\varepsilon=8$ and $\varepsilon=16$ on the test sets. 
	We perform the interpolation on the logarithmic scale since the values of $\lambda$ are non-negative, a usual linear interpolation would lead to negative values of $\lambda$.
	The resulting $\lambda$ values for $\varepsilon \in \{1, \dots, 16\}$ are given in Table~\ref{tab:app_gradalign_lambda}.
	We note that at the end we do not report the results with $\varepsilon > 12$ for SVHN since many models have trivial robustness close to that of a constant classifier.
	\begin{table}[h]
		\caption{\texttt{GradAlign} $\lambda$ values used for the experiments on CIFAR-10 and SVHN. These values are obtained via a linear interpolation on the logarithmic scale between successful $\lambda$ values at $\varepsilon=8$ and $\varepsilon=16$.}
		\label{tab:app_gradalign_lambda}
		\centering
		\setlength{\tabcolsep}{3pt}
		{\small
			\begin{tabular}{lcccccccccccccccc} 
				$\varepsilon$ ($/ 255$)       & 1 & 2 & 3 & 4 & 5 & 6 & 7 & 8 & 9 & 10 & 11 & 12 & 13 & 14 & 15 & 16 \\
				\midrule
				$\lambda_{CIFAR-10}$    & 0.03 & 0.04 & 0.05 & 0.06 & 0.08 & 0.11 & 0.15 & 0.20 & 0.27 & 0.36 & 0.47 & 0.63 & 0.84 & 1.12 & 1.50 & 2.00 \\ 
				$\lambda_{SVHN}$     & 1.66 & 1.76 & 1.86 & 1.98 & 2.10 & 2.22 & 2.36 & 2.50 & 2.65 & 2.81 & 2.98 & 3.16 & 3.35 & 3.56 & 3.77 & 4.00 \\
			\end{tabular}
		}
	\end{table}
	For the PGD-2~+~\texttt{GradAlign} experiments reported below in Table~\ref{tab:results_cifar10} and Table~\ref{tab:results_svhn}, we use $\lambda=0.1$ for the CIFAR-10 and $\lambda=0.5$ for SVHN experiments.

	\myparagraph{Training details for the single-layer CNN.}
	The single-layer CNN that we study in Sec.~\ref{sec:understanding_co_via_ga} has 4 convolutional filters, each of them of size $3\times3$. After the convolution we apply ReLU activation, and then we directly have a fully-connected layer, i.e. we do not use any pooling layer. For training we use the ADAM optimizer with learning rate $0.003$ for 30 epochs using the same cyclical learning rate schedule.

	\myparagraph{ImageNet experiments.}
	We use ResNet-50 following the training scheme of \cite{wong2020fast} which includes 3 training stages on different image resolution. 
	For \texttt{GradAlign}, we slightly reduce the batch size on the second and third stages from 224 and 128 to 180 and 100 respectively in order to reduce the memory consumption. For all $\varepsilon \in \{2, 4, 6\}$, we train FGSM models with \texttt{GradAlign} using $\lambda \in \{0.01, 0.1\}$. The final $\lambda$ we report are $\lambda \in \{0.01, 0.01, 0.1\}$ for $\varepsilon \in \{2, 4, 6\}$ respectively.

	\myparagraph{Computing infrastructure.}  
	We perform all our experiments on NVIDIA V100 GPUs with 32GB of memory.

	\section{Supporting experiments and visualizations for Sec.~\ref{sec:role_limitation_of_rs} and Sec.~\ref{sec:understanding_co_via_ga}} \label{app:support_exps}
	We describe here supporting experiments and visualizations related to Sec.~\ref{sec:role_limitation_of_rs} and Sec.~\ref{sec:understanding_co_via_ga}.
	
	\subsection{Quality of the linear approximation for ReLU networks} \label{app:sec:quality_lin_approx}
	\begin{figure}[t]
		\centering
		\small
		\begin{tabular}{c c}
			\textbf{a.)} Standard model &
			\textbf{b.)} PGD-trained model \\
			\includegraphics[width=0.38\columnwidth]{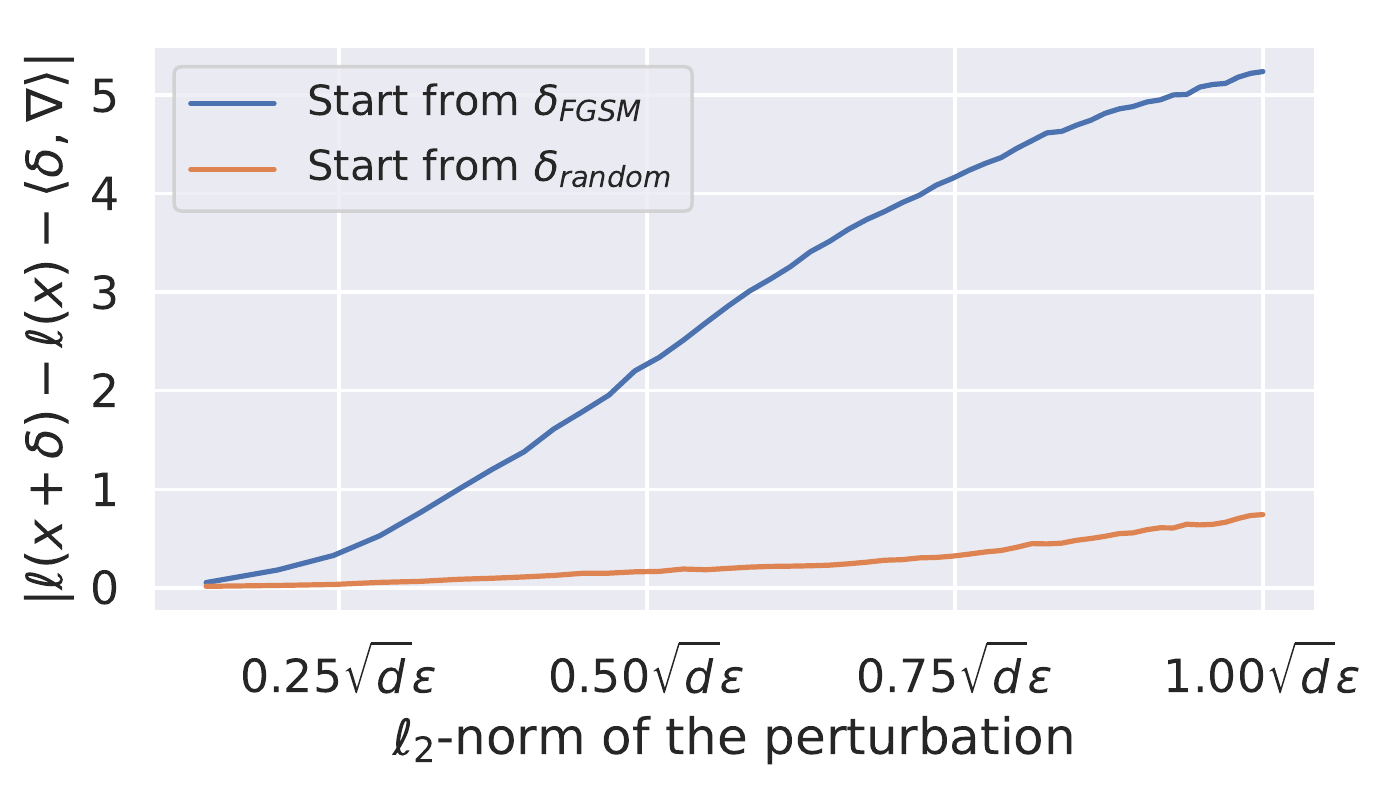} \vspace{10pt} &
			\includegraphics[width=0.4\columnwidth]{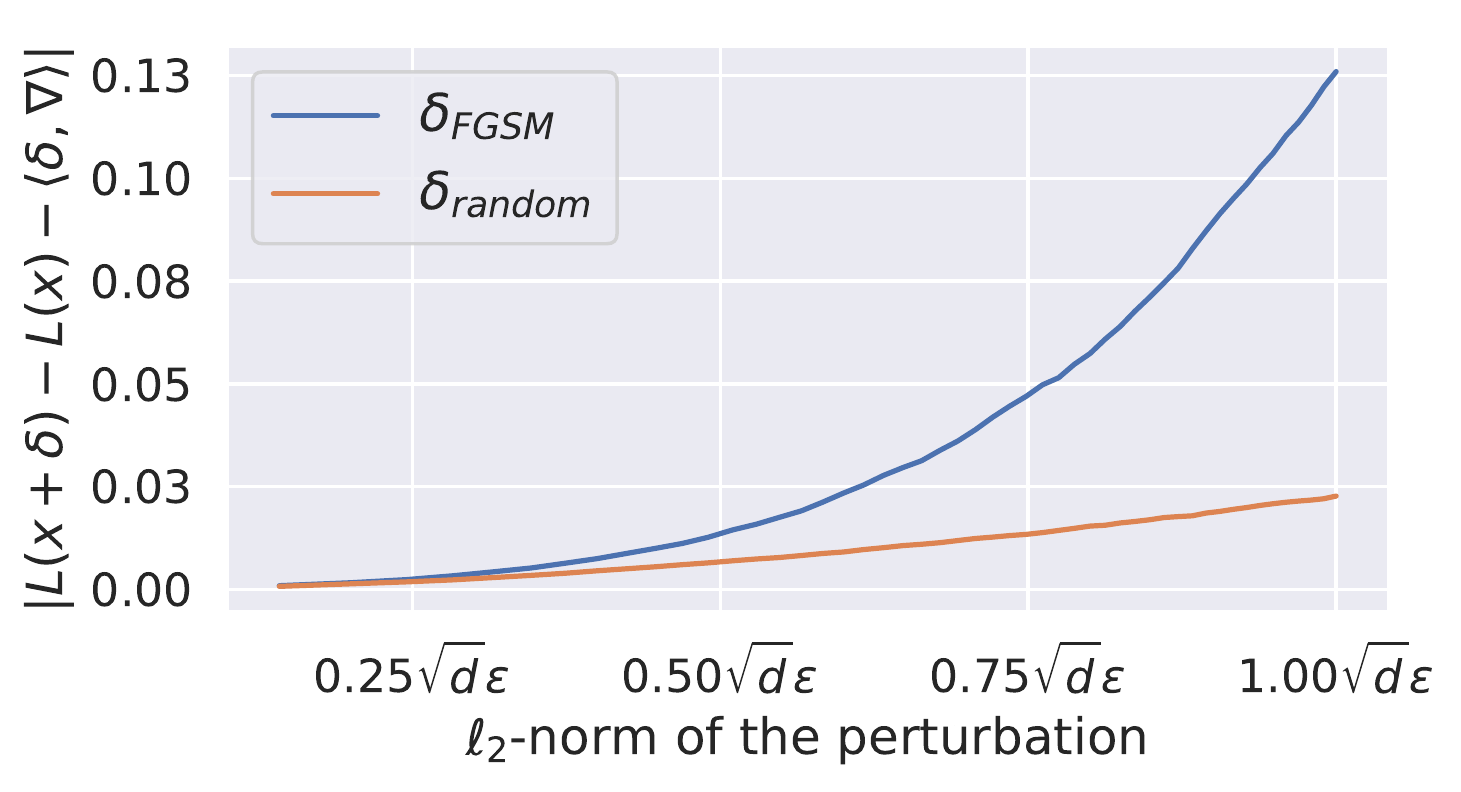}
		\end{tabular}
		\vspace{-5mm}
		\caption{The quality of the linear approximation of $\l(x+\delta)$ for $\delta$ with different $\ell_2$-norm for $\norm{\delta}_\infty$ fixed to $\varepsilon$ for a standard and PGD-trained ResNet-18 on CIFAR-10.}
		\label{fig:quality_linear_approx}
	\end{figure}
	\begin{figure}[b]
		\centering
		\includegraphics[width=0.32\textwidth]{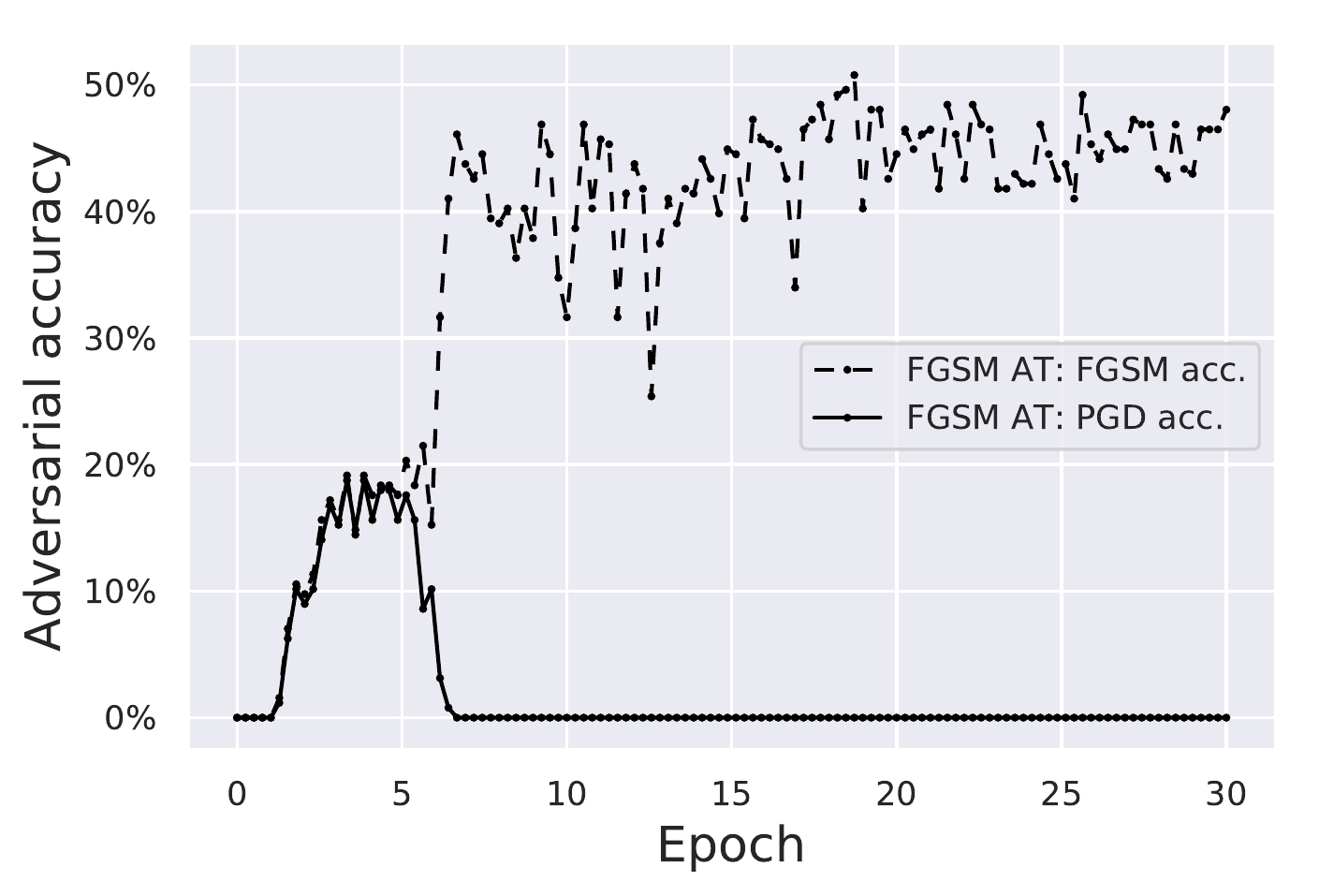}
		\includegraphics[width=0.32\textwidth]{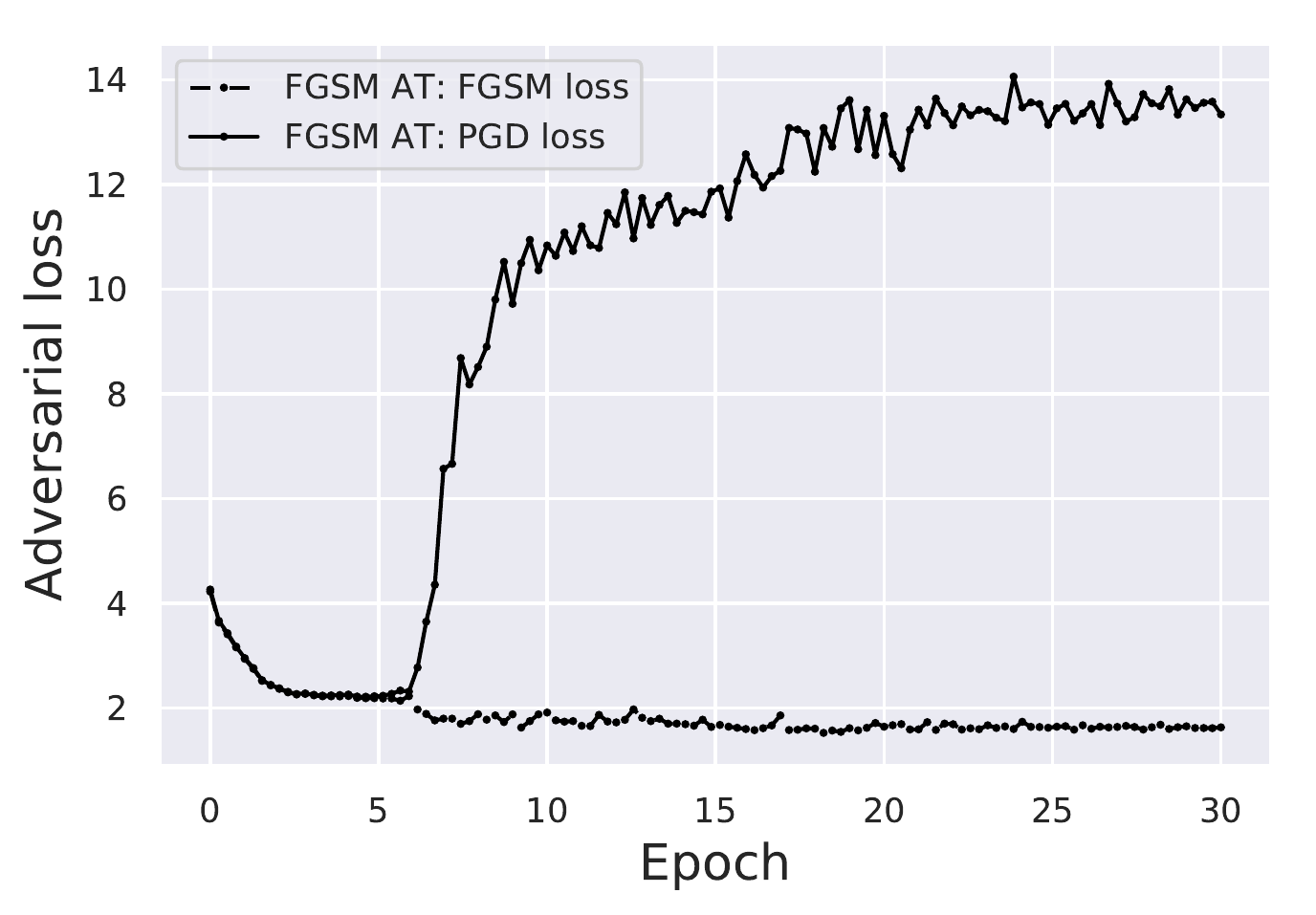}
		\includegraphics[width=0.32\textwidth]{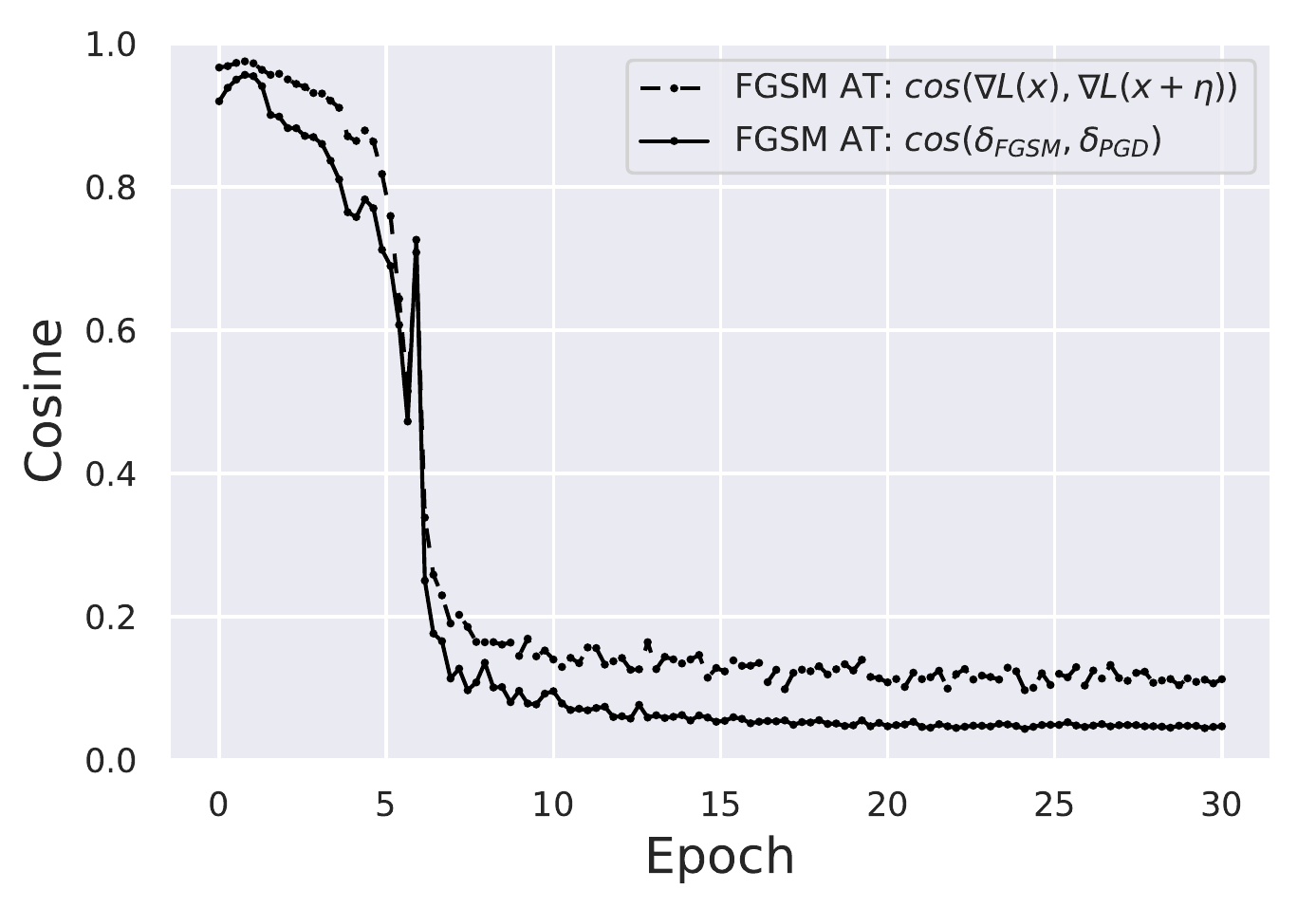}
		\caption{Visualization of the training process of an FGSM trained CNN with 4 filters with $\varepsilon = \nicefrac{10}{255}$. We can observe catastrophic overfitting around epoch 6.}
		\label{fig:training_curves_cnn4}
	\end{figure}
	For the loss function $\ell$ of a ReLU-network, we compute empirically the quality of the linear approximation defined as 
	\[ |\l(x+\delta) - \l(x) - \inner{\delta, \nabla_x \l(x)}|, \]
	where the dependency of the loss $\ell$ on the label $y$ and parameters $\theta$ are omitted for clarity.
	Then we perform the following experiment: we take a perturbation $\delta \in \{-\varepsilon, \varepsilon\}^d$, and then zero out different fractions of its coordinates, which leads to perturbations with a fixed $\norm{\delta}_\infty=\varepsilon$, but with different $\norm{\delta}_2 \in [0, \sqrt{d}\varepsilon]$. 
	As the starting $\delta$ we choose two types of perturbations: $\delta_{FGSM}$ generated by FGSM and $\delta_{random}$ sampled uniformly from the corners of the $\ell_\infty$-ball.
	We plot the results in Fig.~\ref{fig:quality_linear_approx} on CIFAR-10 for $\varepsilon=8/255$ averaged over 512 test points, and conclude that for both $\delta_{FGSM}$ and $\delta_{random}$ the validity of the linear approximation crucially depends on $\norm{\delta}_2$ even when $\norm{\delta}_\infty$ is fixed. The phenomenon is even more pronounced for FGSM perturbations as the linearization error is much higher there. Moreover, this observation is consistent across both standardly and adversarially trained ResNet-18 models.

	\subsection{Catastrophic overfitting in a single-layer CNN} \label{app:sec:cnn_cat_overfitting}
	We describe here complementary figures to Sec.~\ref{sec:understanding_co_via_ga} which are related to the single-layer CNN.
	
	\myparagraph{Training curves.}
	In Fig.~\ref{fig:training_curves_cnn4}, we show the evolution of the FGSM/PGD accuracy, FGSM/PGD loss, and gradient alignment together with $\cos(\delta_{FGSM}, \delta_{PGD})$. 
	We observe that catastrophic overfitting occurs around epoch 6 and that its pattern is the same as for the deep ResNet which was illustrated in Fig.~\ref{fig:training_curves_resnet}. Namely, we see that concurrently the following changes occur around epoch 6: 
	(a) there is a sudden drop of PGD accuracy with an increase in FGSM accuracy, 
	(b) the PGD loss grows by an order of magnitude while the FGSM loss decreases, 
	(c) both gradient alignment and $\cos(\delta_{FGSM}, \delta_{PGD})$ significantly decrease. 
	Throughout all our experiments we observe a very high correlation between $\cos(\delta_{FGSM}, \delta_{PGD})$ and gradient alignment. This motivates our proposed regularizer \texttt{GradAlign} which relies on the cosine between $\nabla_x \l(x, y; \theta)$ and $\nabla_x \l(x+\eta, y; \theta)$, where $\eta$ is a \textit{random} point. In this way, we avoid using an iterative procedure inside the regularizer unlike, for example, the approach of \cite{qin2019adversarial}.

	\myparagraph{Additional filters.}
	\begin{figure}[t]
		\centering
		\setlength{\tabcolsep}{3pt}
		\begin{tabular}{cccc|ccc}
			& Epoch 3 & Epoch 4 & Epoch 5 & Epoch 6 & Epoch 7 & Epoch 30 \\
			$w_1$-R &
			\includegraphics[width=0.144\textwidth]{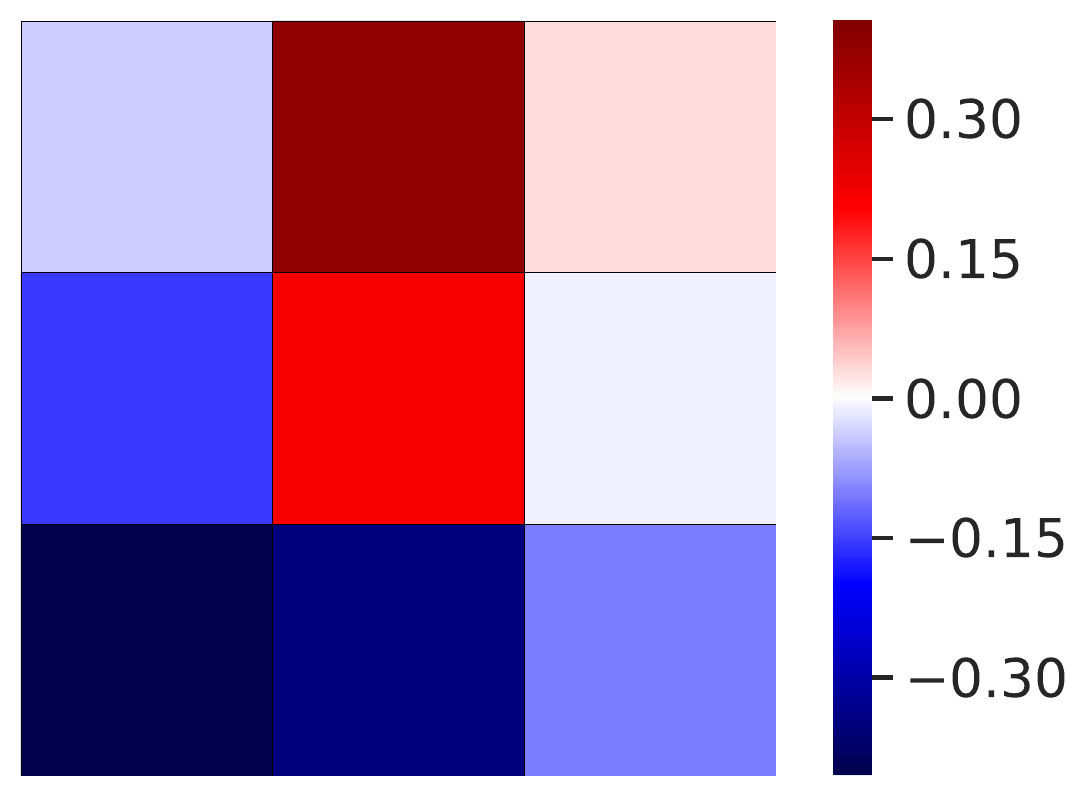} &
			\includegraphics[width=0.144\textwidth]{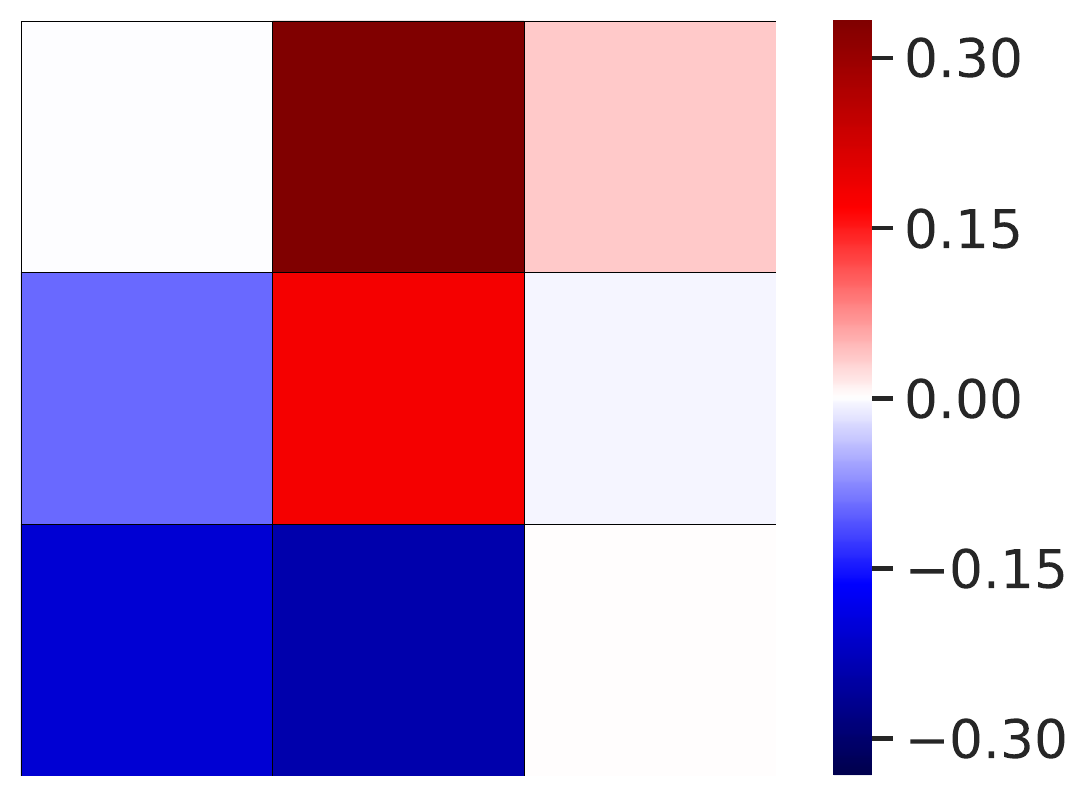} &
			\includegraphics[width=0.144\textwidth]{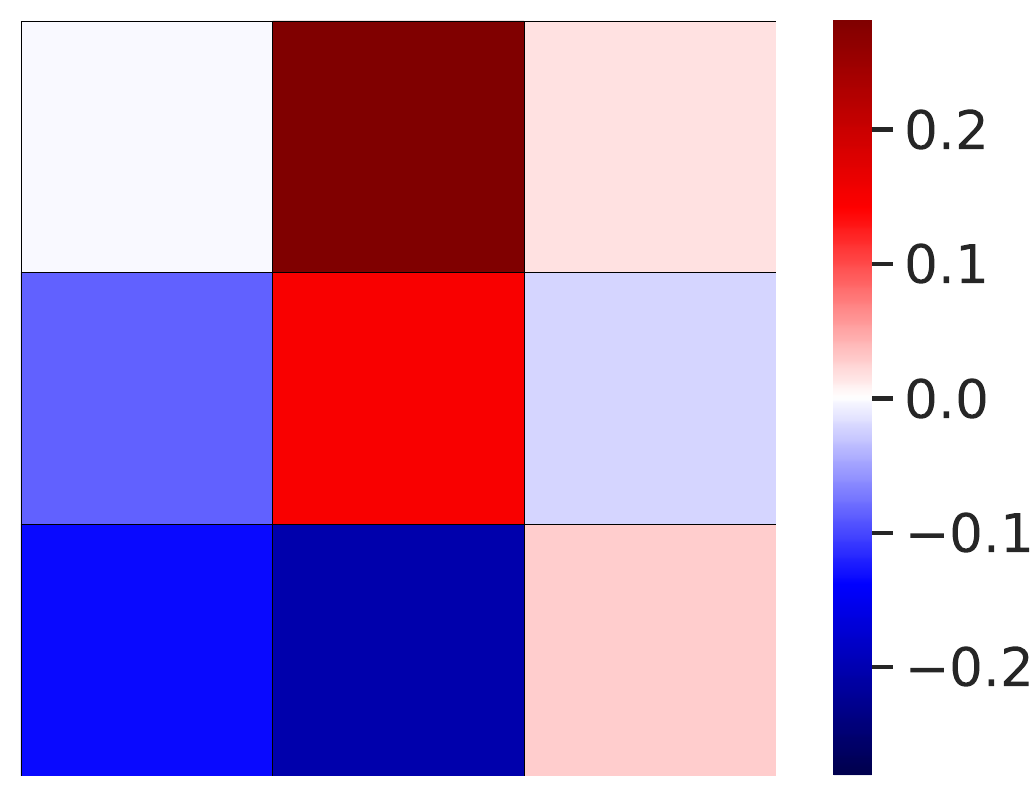} &
			\includegraphics[width=0.144\textwidth]{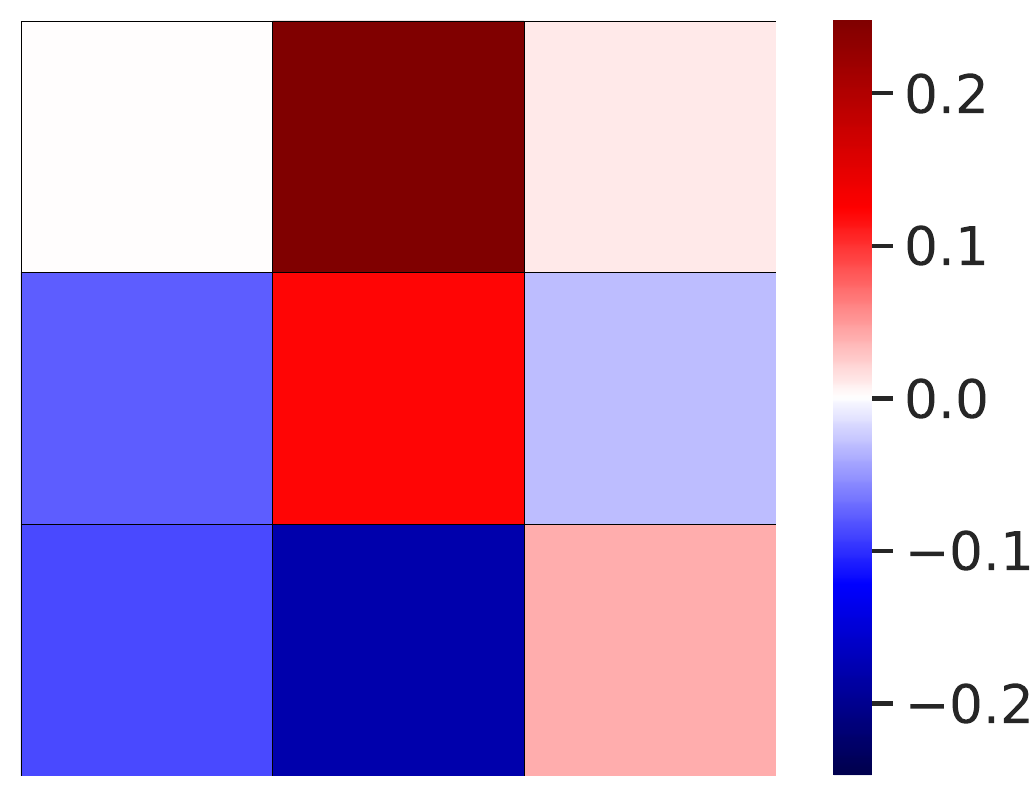} &
			\includegraphics[width=0.144\textwidth]{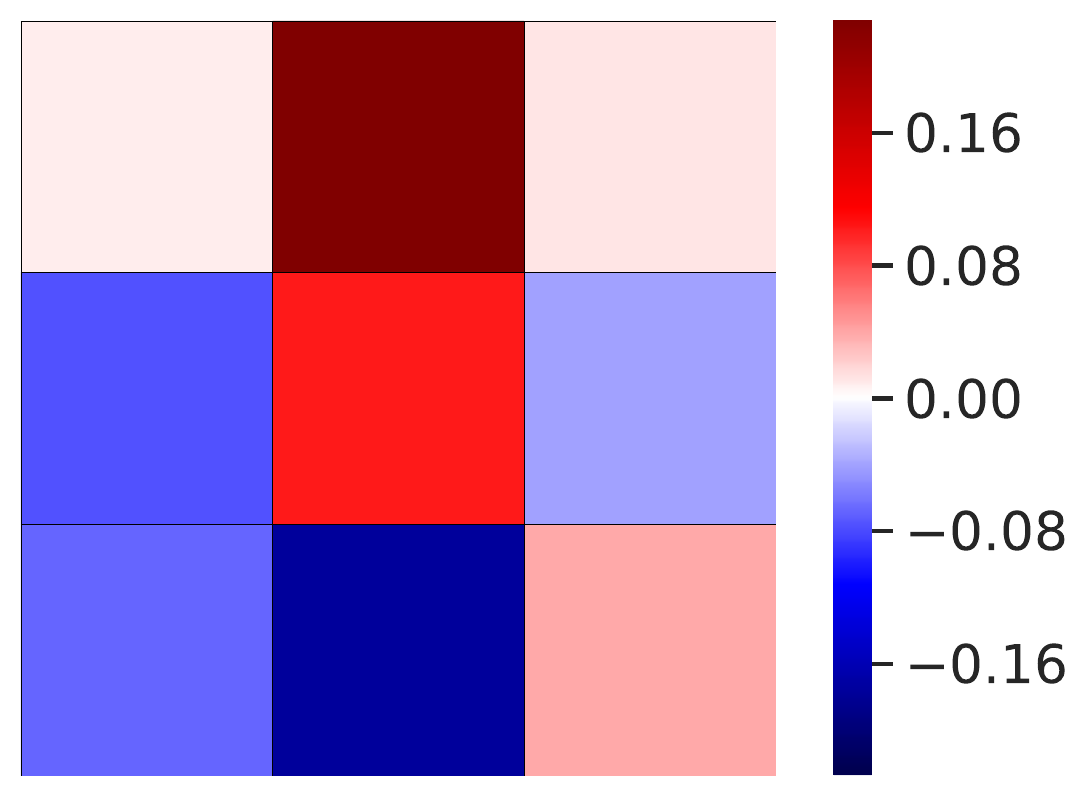} &
			\includegraphics[width=0.144\textwidth]{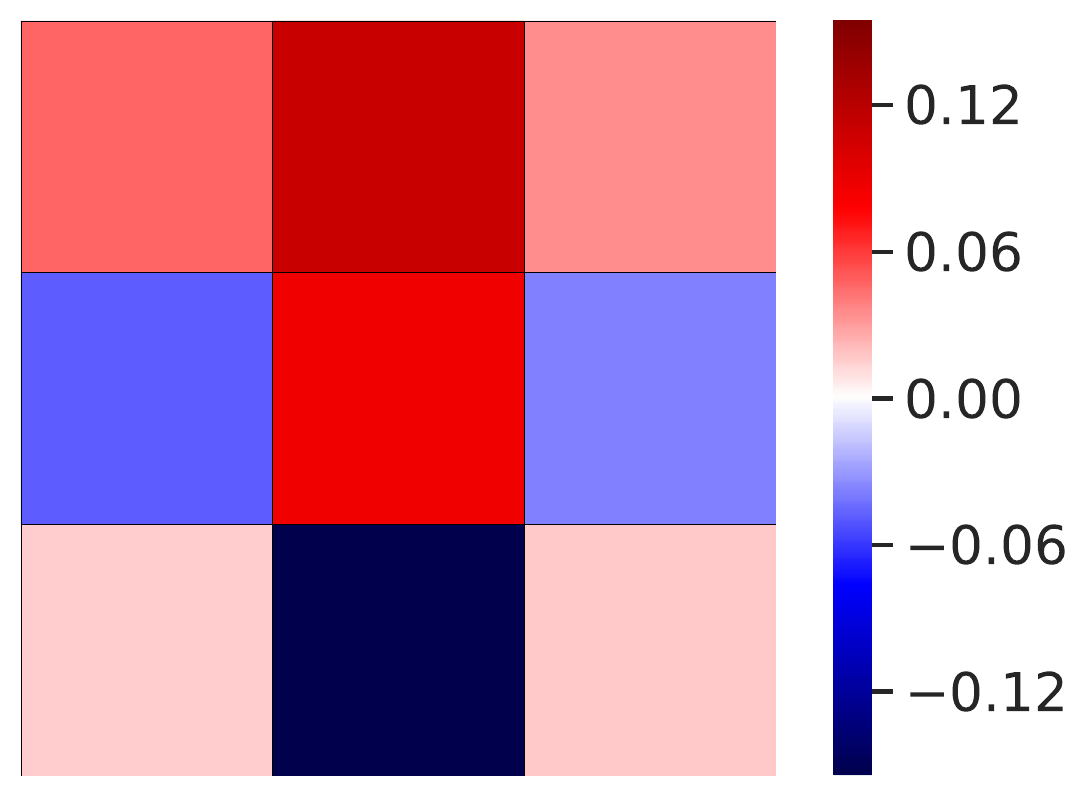} \\
			$w_1$-G &
			\includegraphics[width=0.144\textwidth]{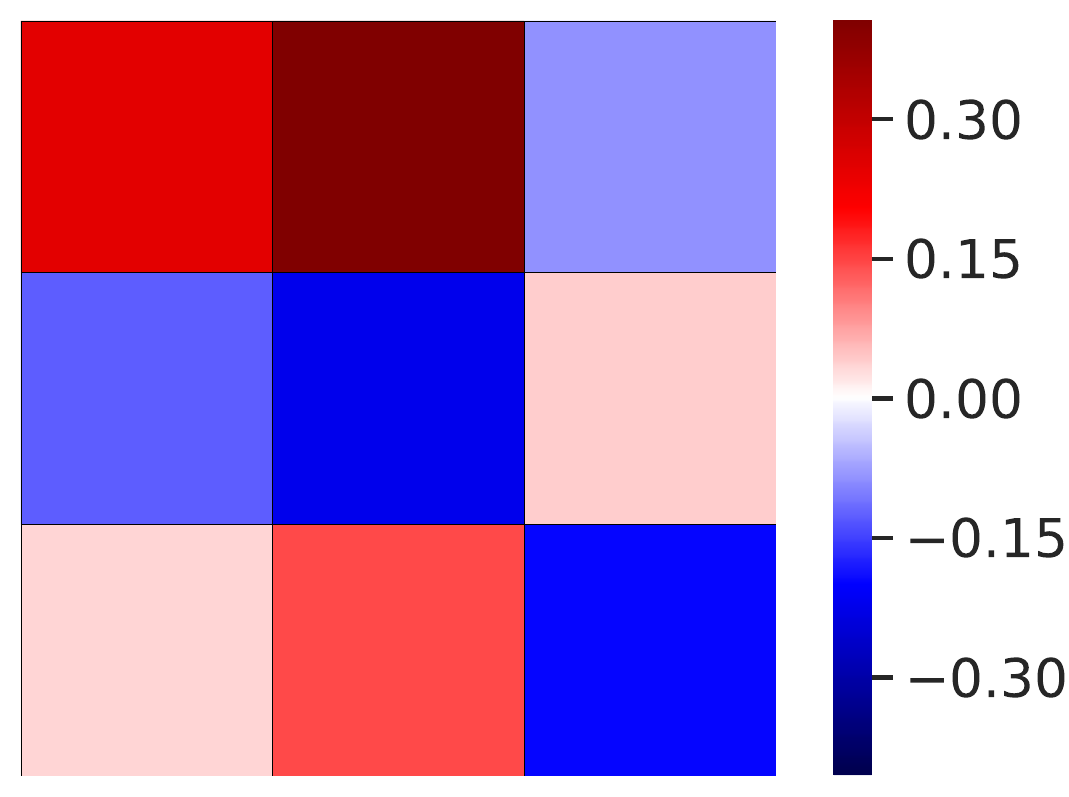} &
			\includegraphics[width=0.144\textwidth]{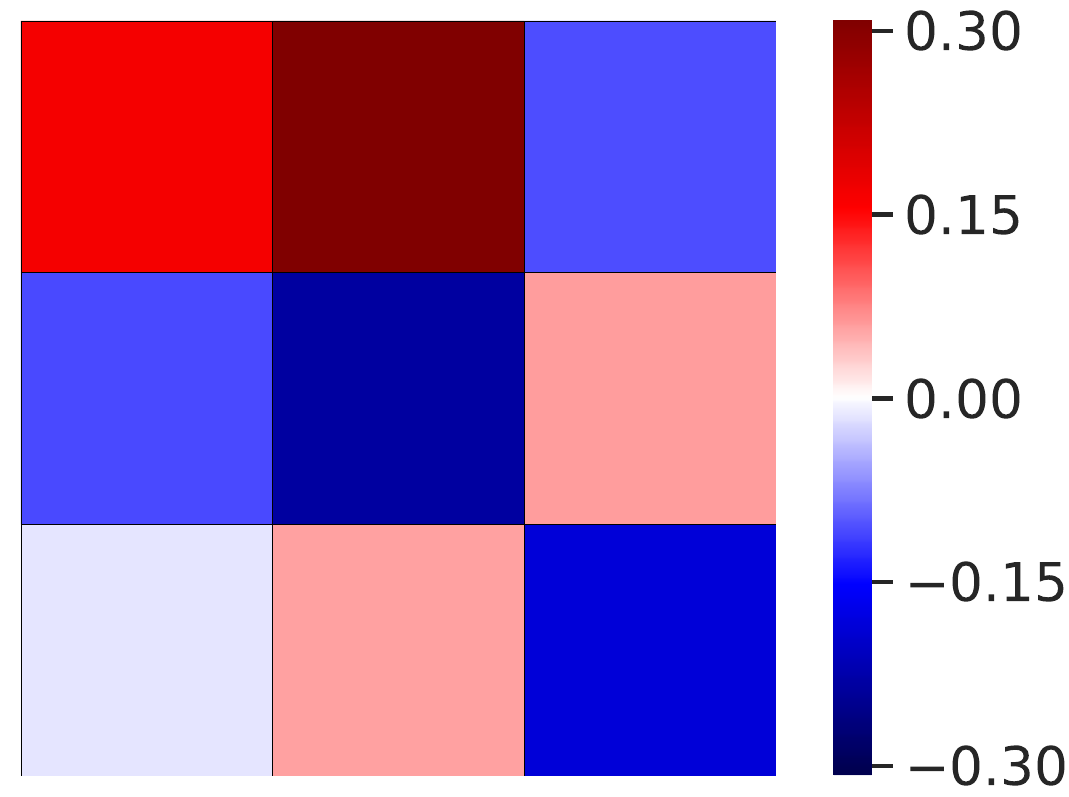} &
			\includegraphics[width=0.144\textwidth]{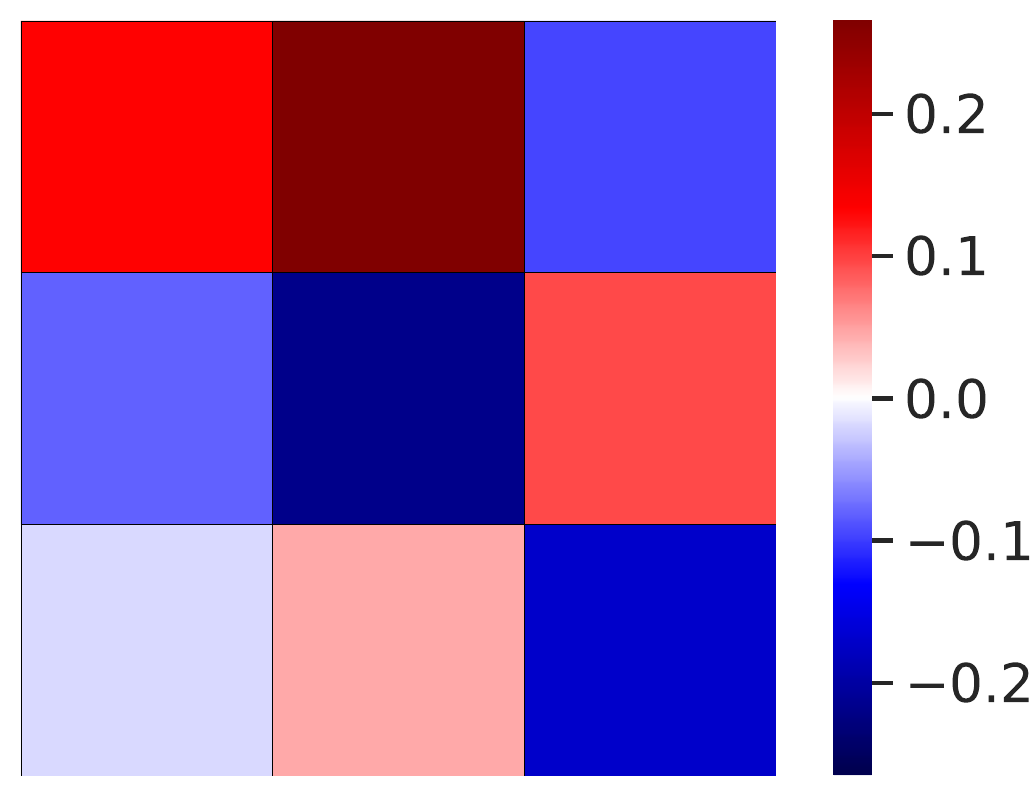} &
			\includegraphics[width=0.144\textwidth]{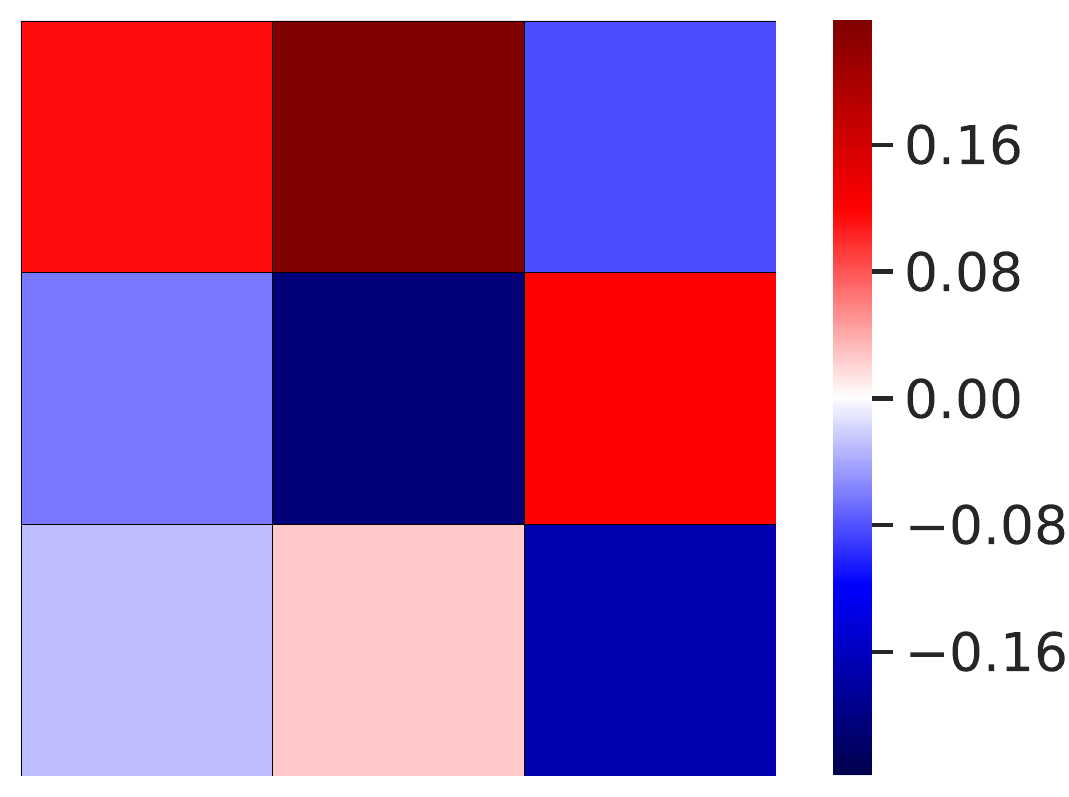} &
			\includegraphics[width=0.144\textwidth]{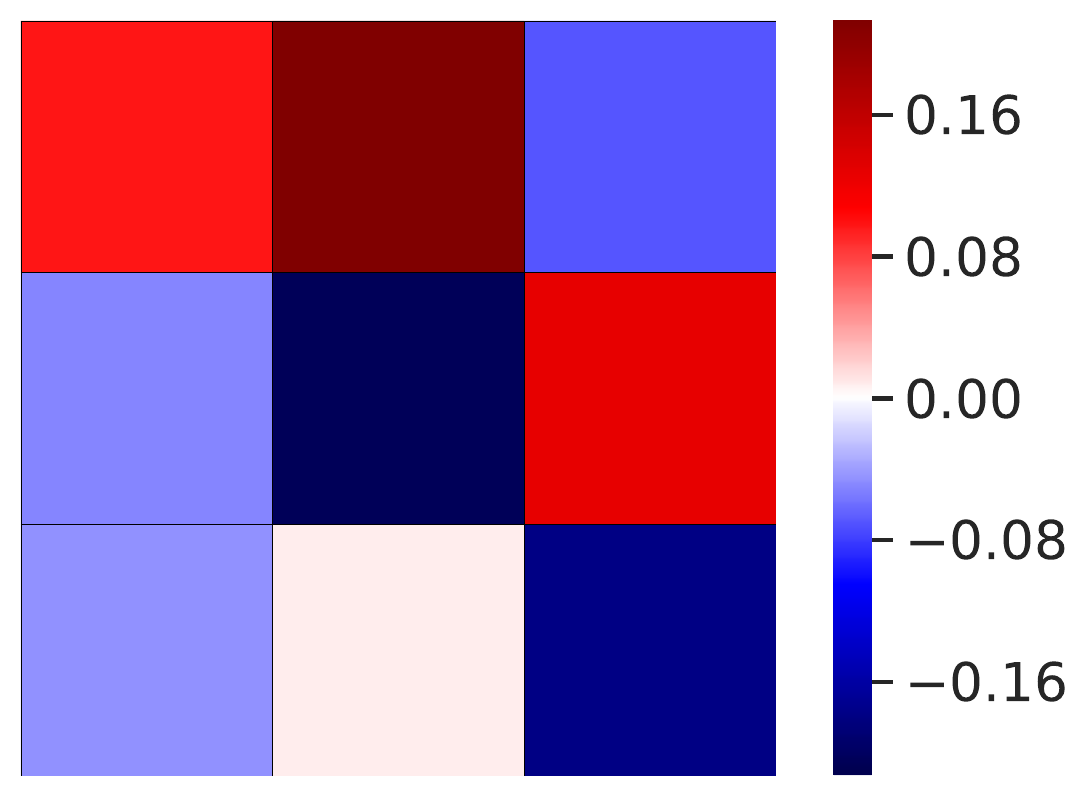} &
			\includegraphics[width=0.144\textwidth]{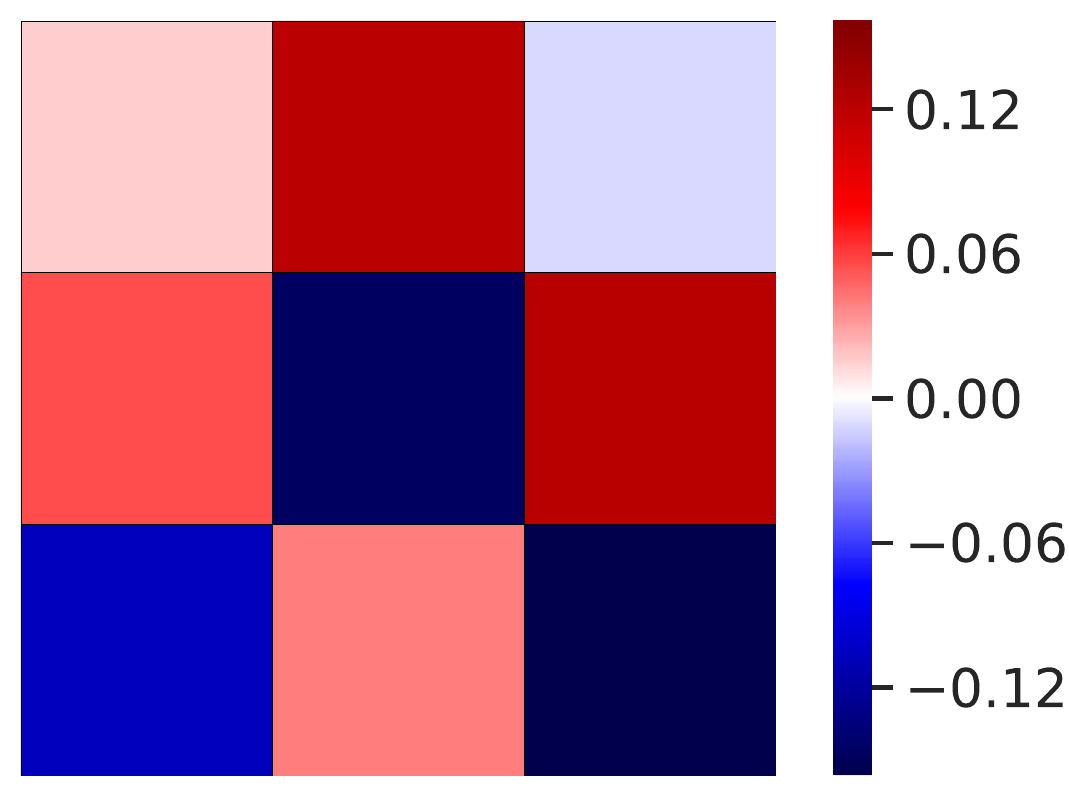} \\
			$w_1$-B &
			\includegraphics[width=0.144\textwidth]{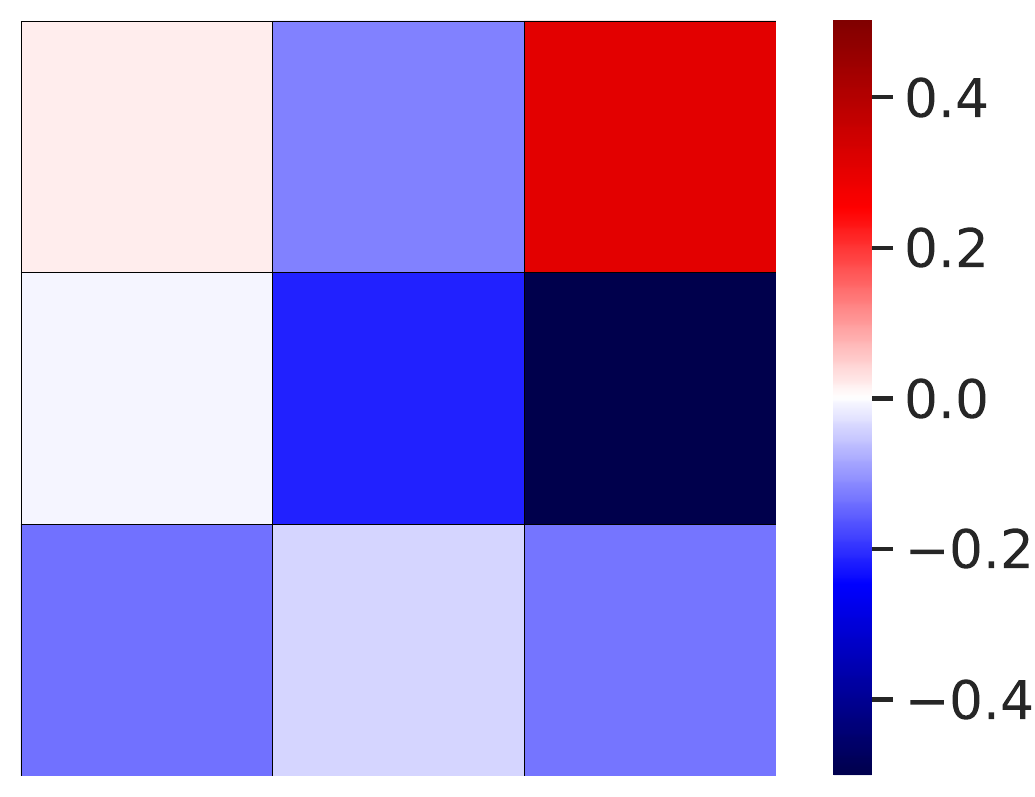} &
			\includegraphics[width=0.144\textwidth]{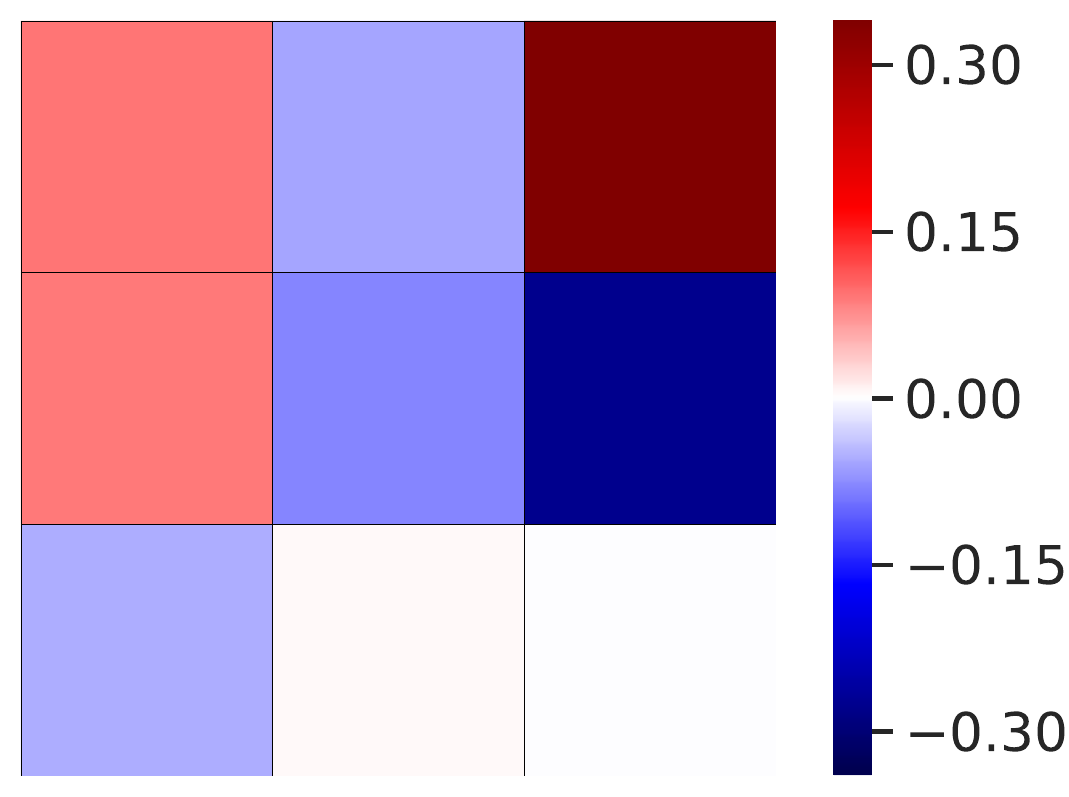} &
			\includegraphics[width=0.144\textwidth]{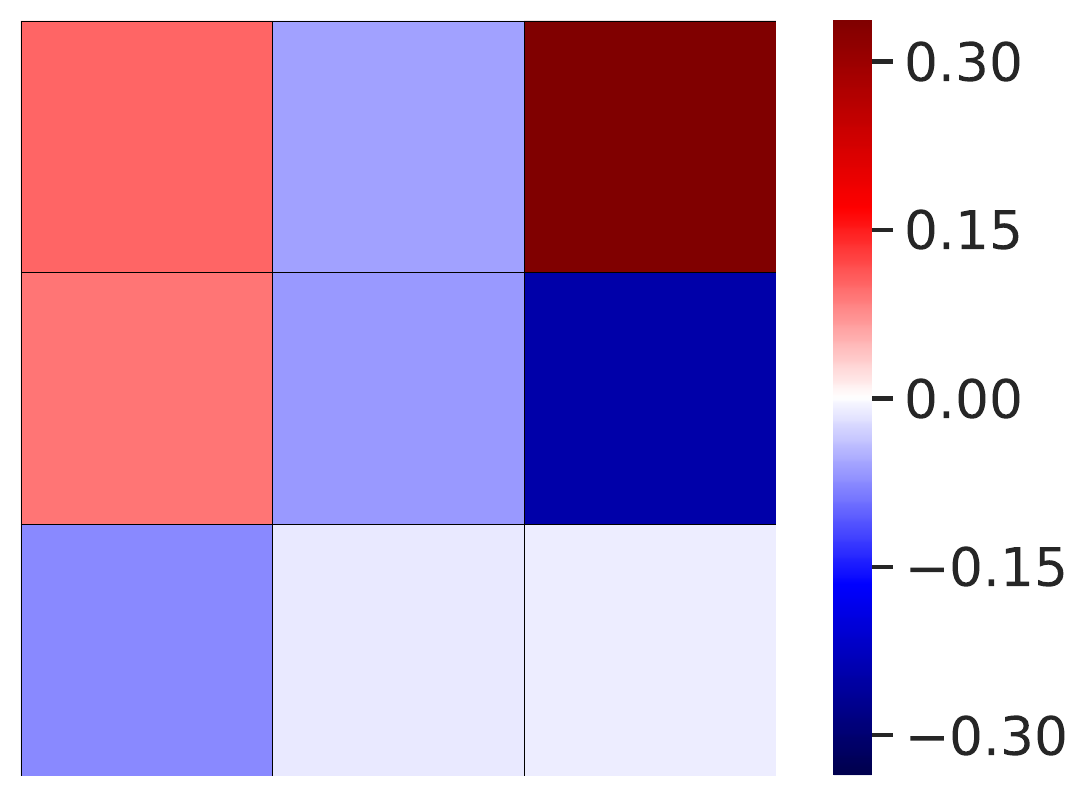} &
			\includegraphics[width=0.144\textwidth]{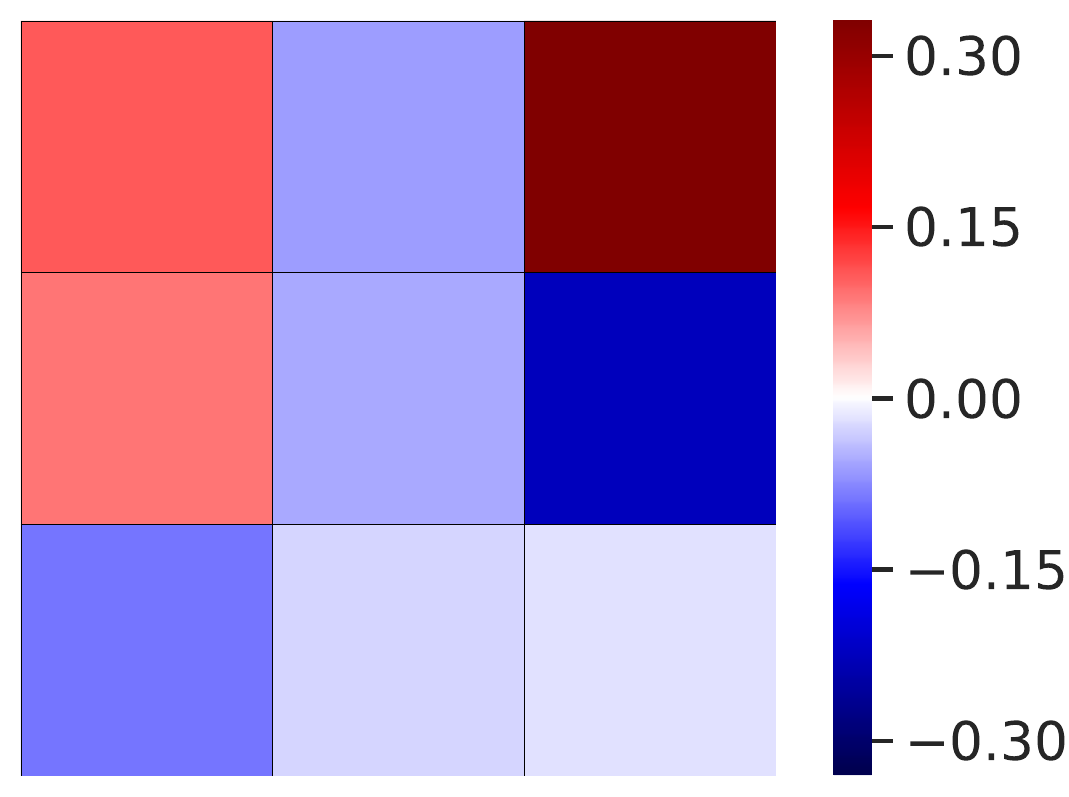} &
			\includegraphics[width=0.144\textwidth]{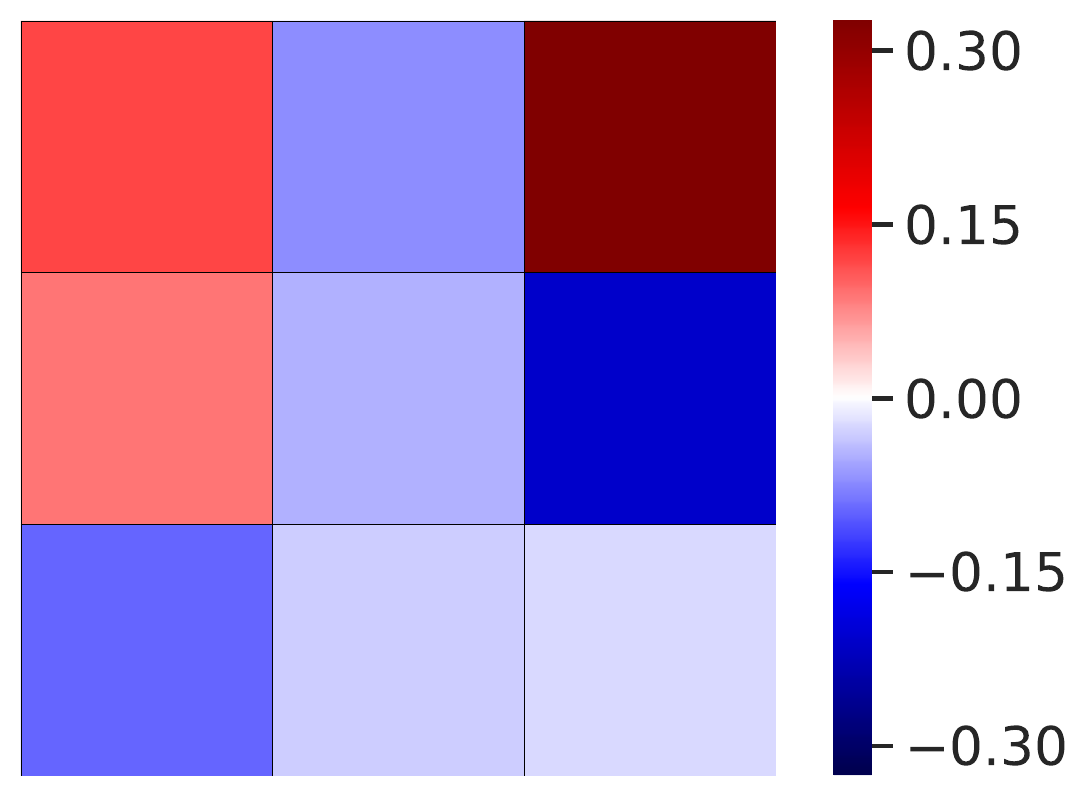} &
			\includegraphics[width=0.144\textwidth]{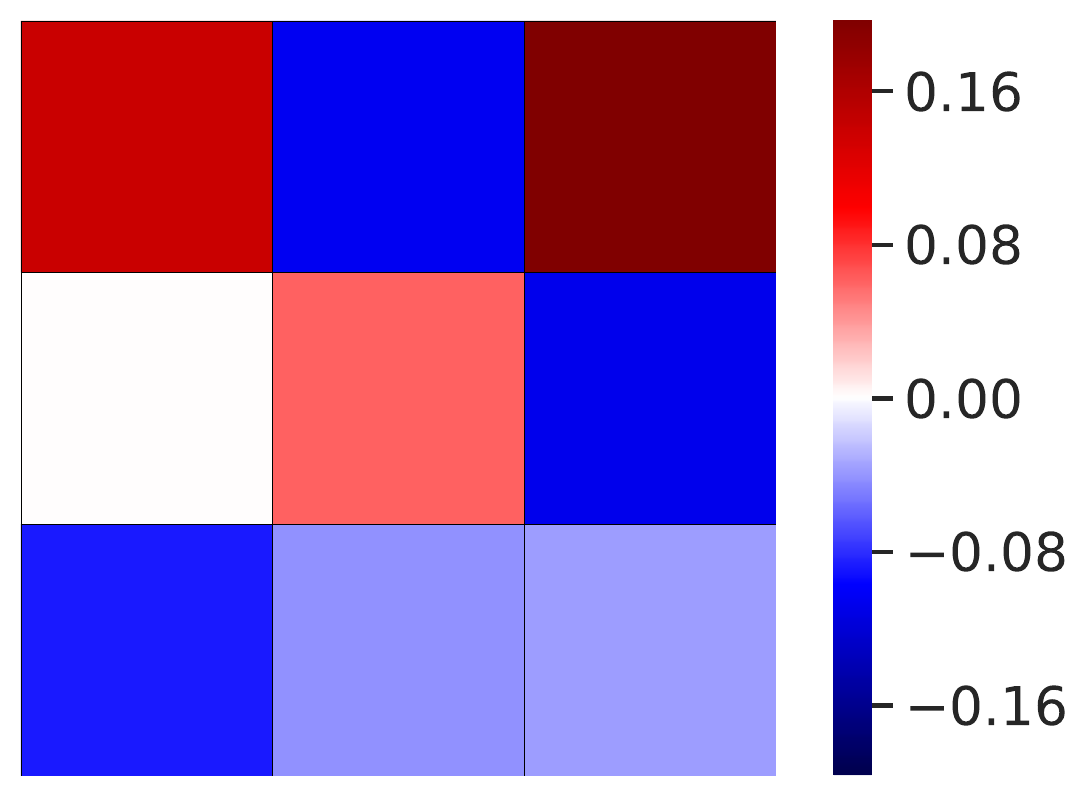} \\
			\hdashline
			\vspace{-7pt} \\
			$w_4$-R &
			\includegraphics[width=0.144\textwidth]{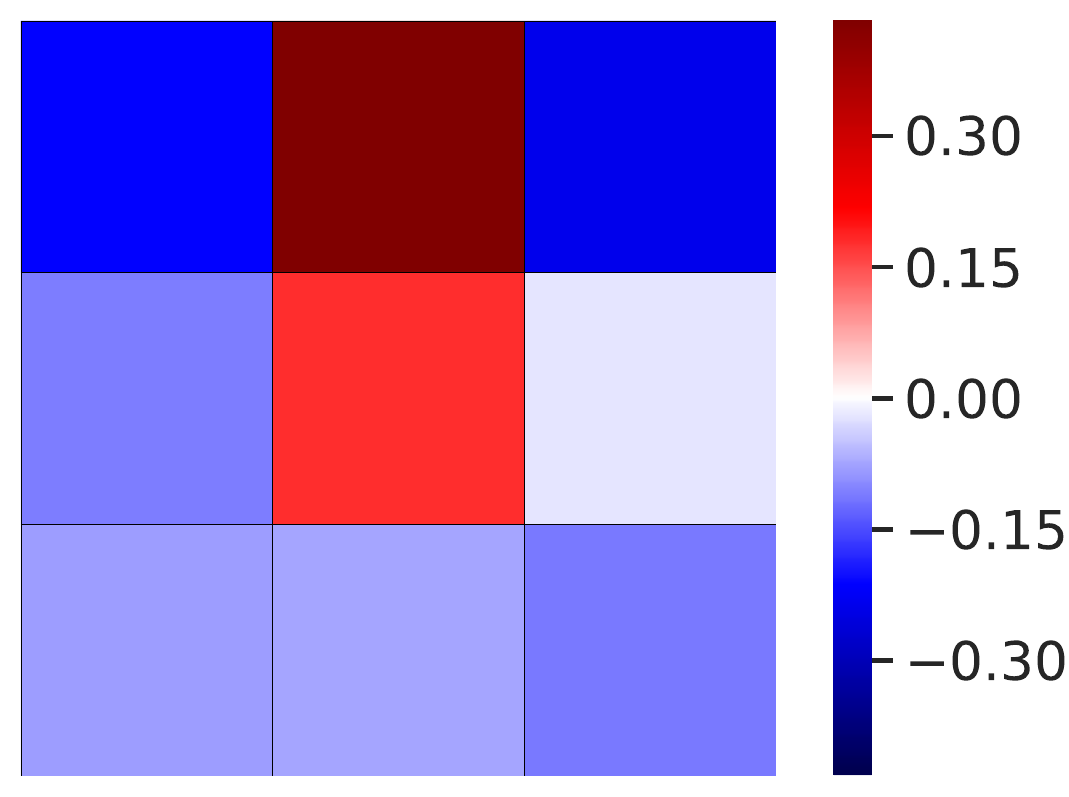} &
			\includegraphics[width=0.144\textwidth]{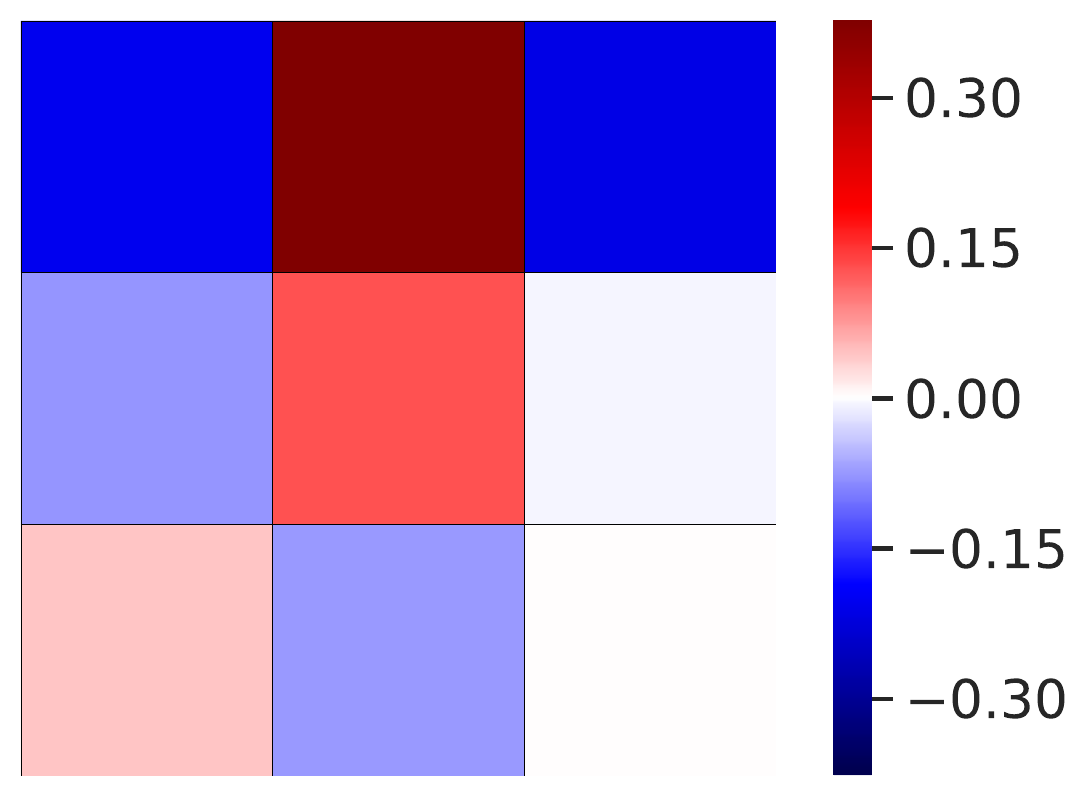} &
			\includegraphics[width=0.144\textwidth]{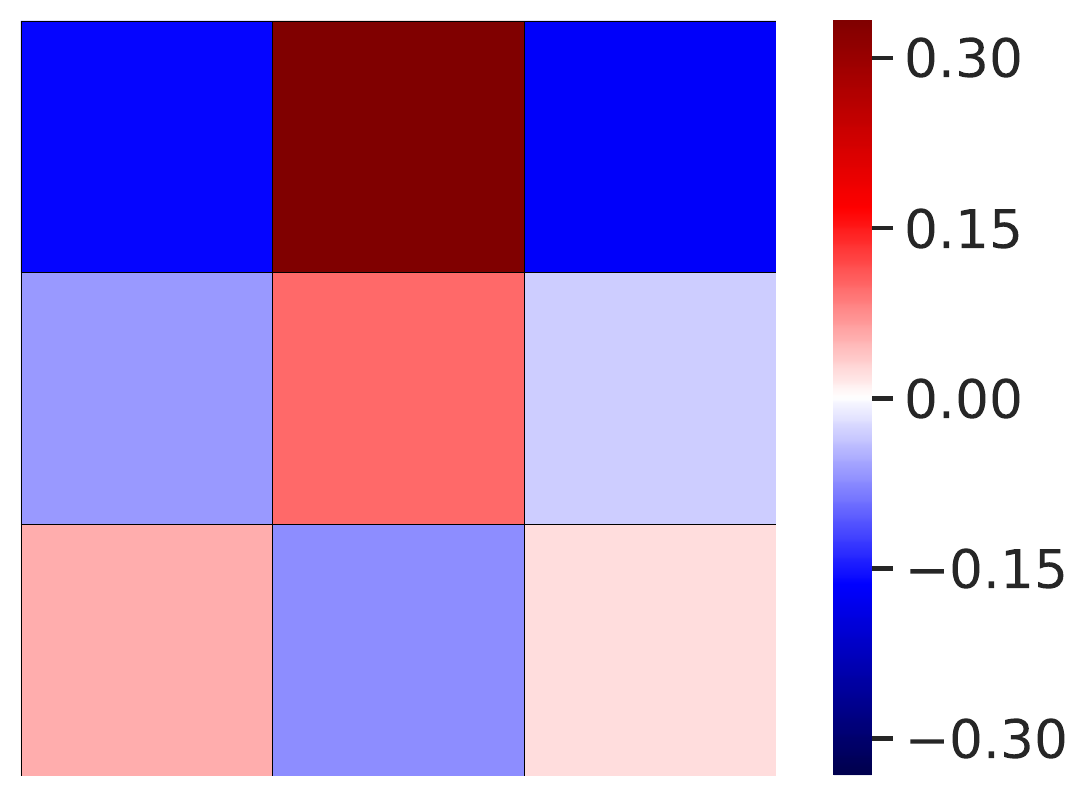} &
			\includegraphics[width=0.144\textwidth]{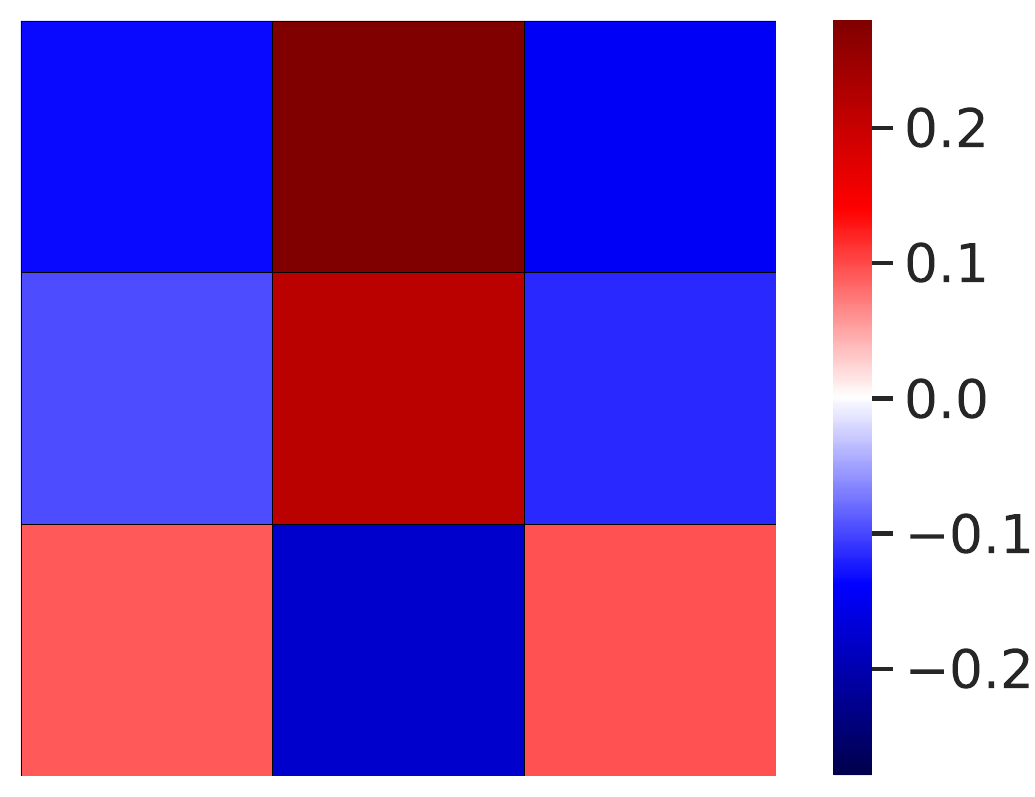} &
			\includegraphics[width=0.144\textwidth]{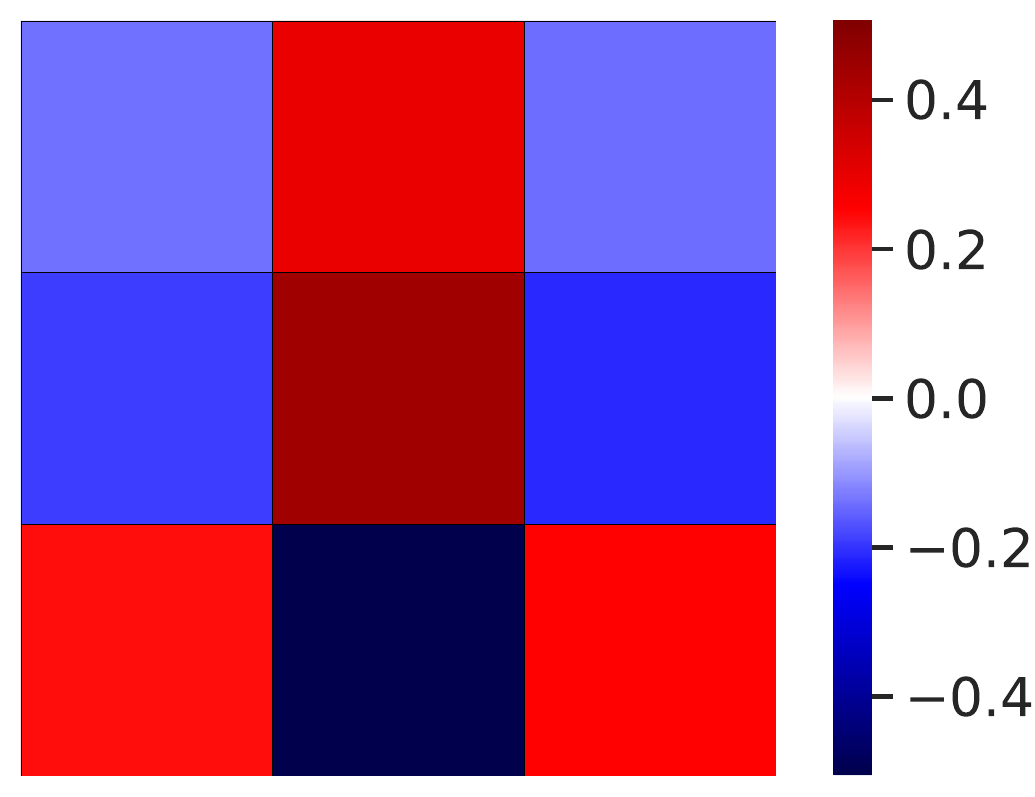} &
			\includegraphics[width=0.144\textwidth]{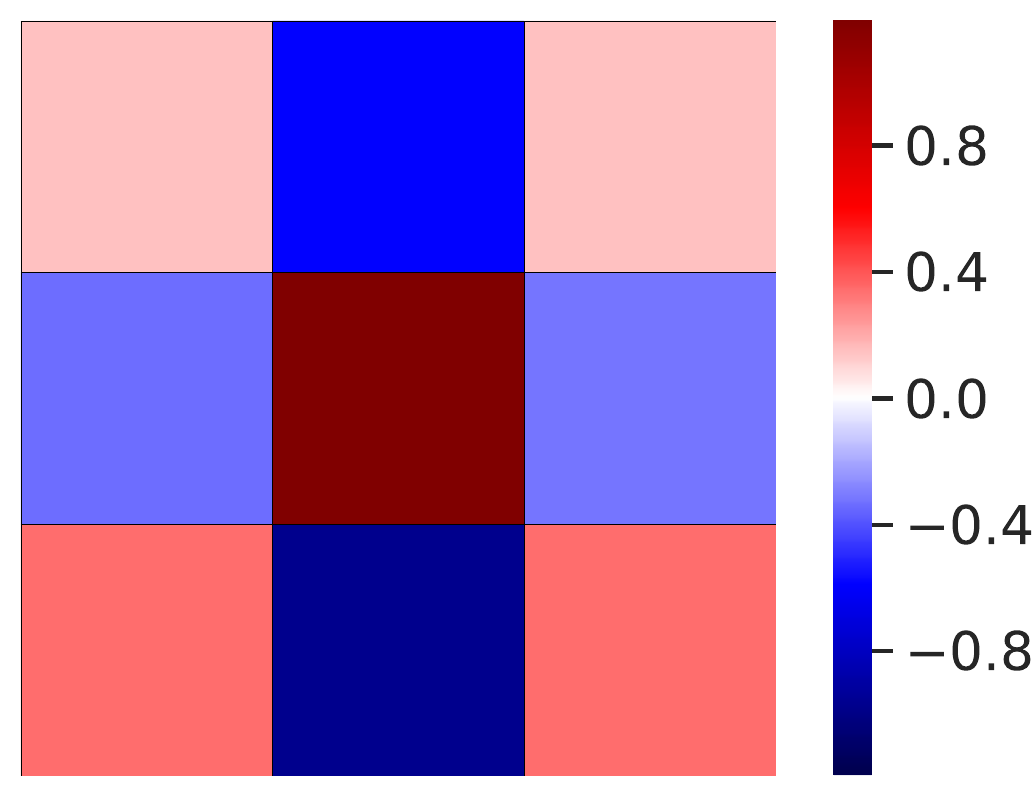} \\
			$w_4$-G &
			\includegraphics[width=0.144\textwidth]{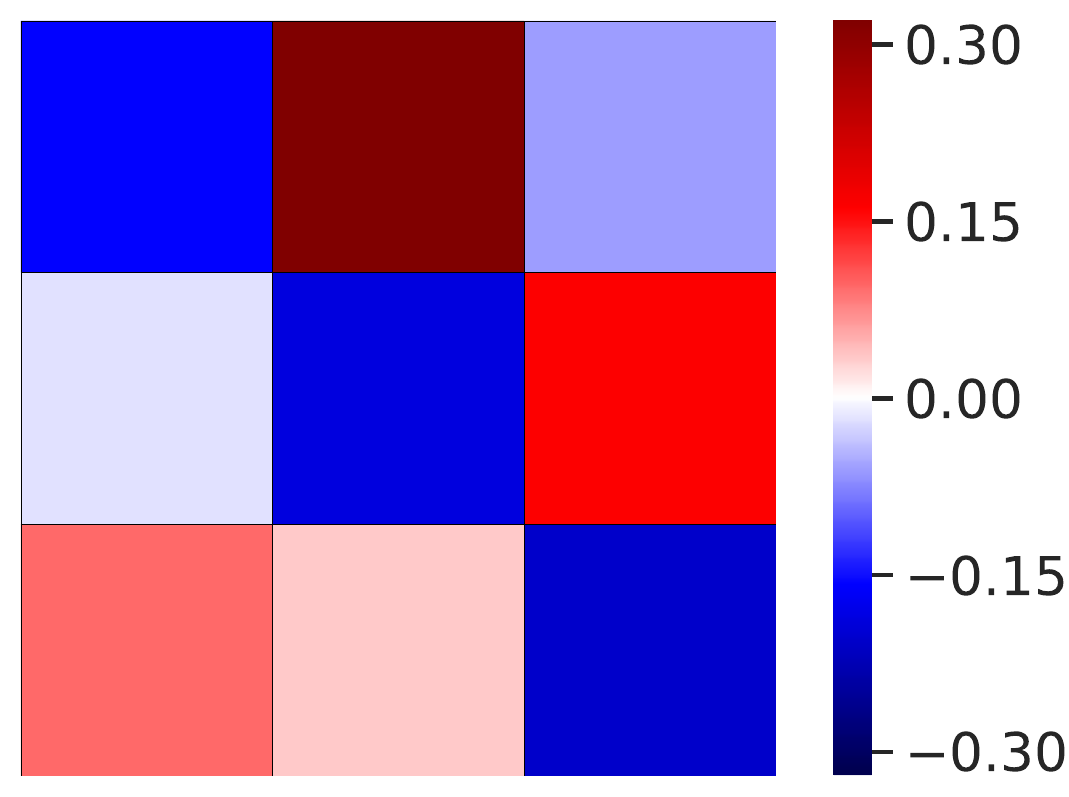} &
			\includegraphics[width=0.144\textwidth]{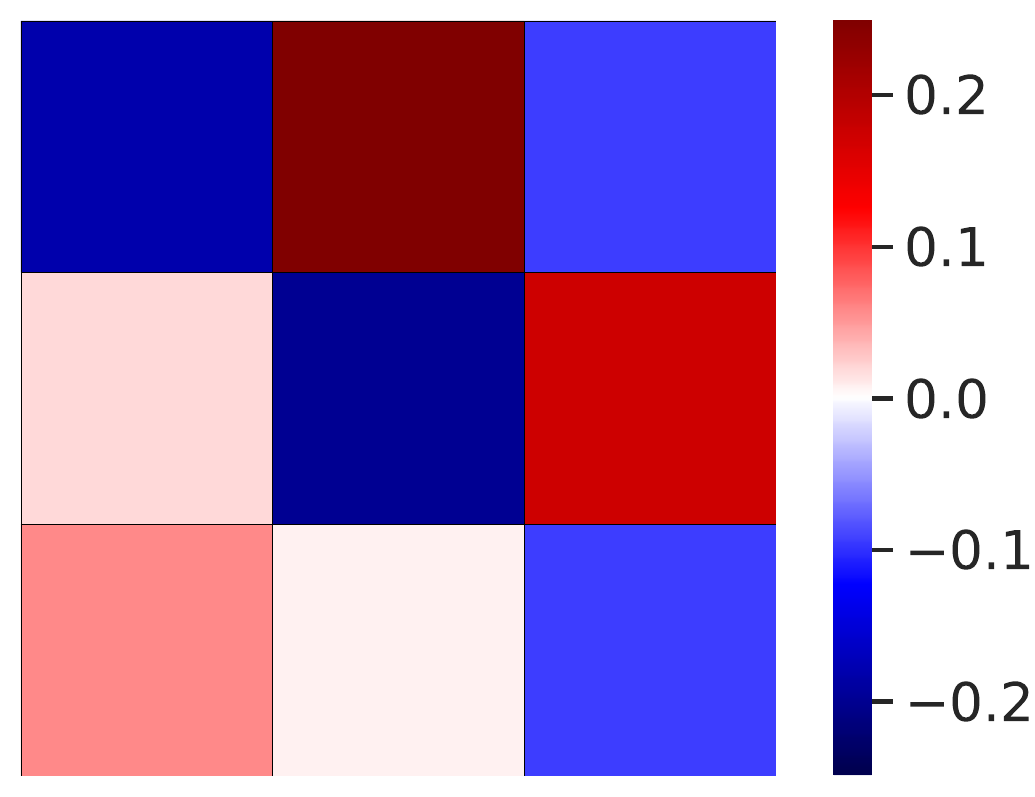} &
			\includegraphics[width=0.144\textwidth]{{conv_filters-model=fgsm_at_cnn4_new-epoch=5-color=1-filter_id=3}.pdf} &
			\includegraphics[width=0.144\textwidth]{{conv_filters-model=fgsm_at_cnn4_new-epoch=6-color=1-filter_id=3}.pdf} &
			\includegraphics[width=0.144\textwidth]{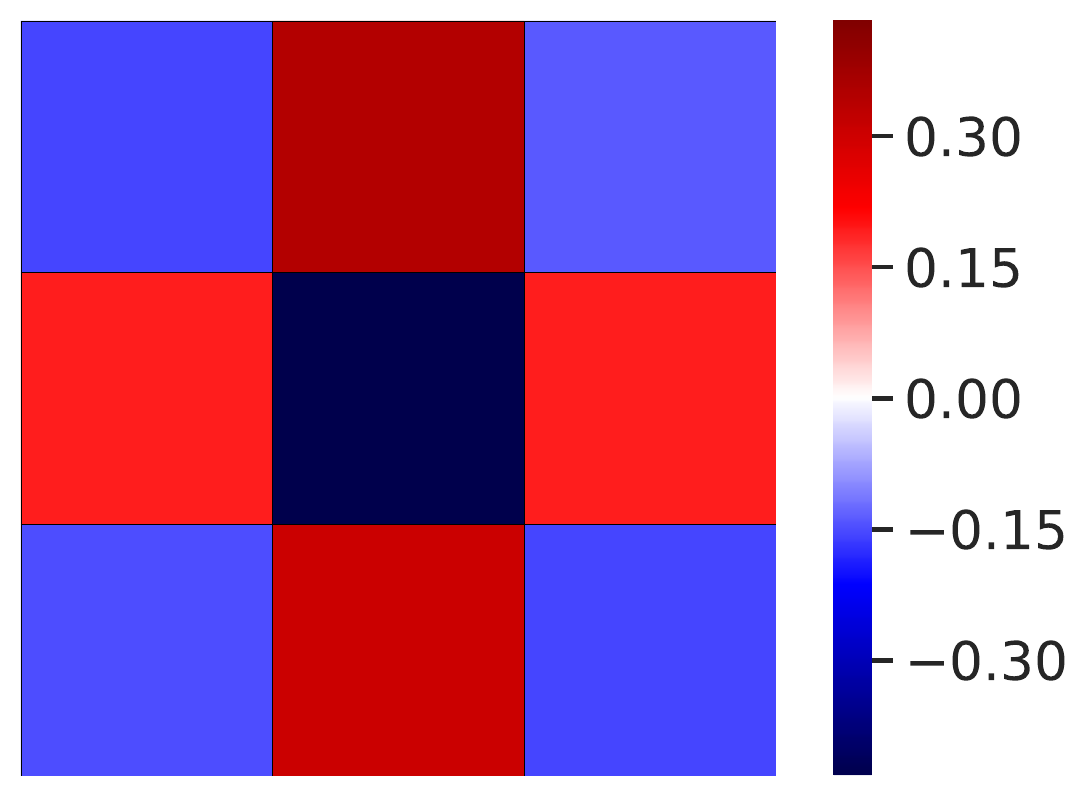} &
			\includegraphics[width=0.144\textwidth]{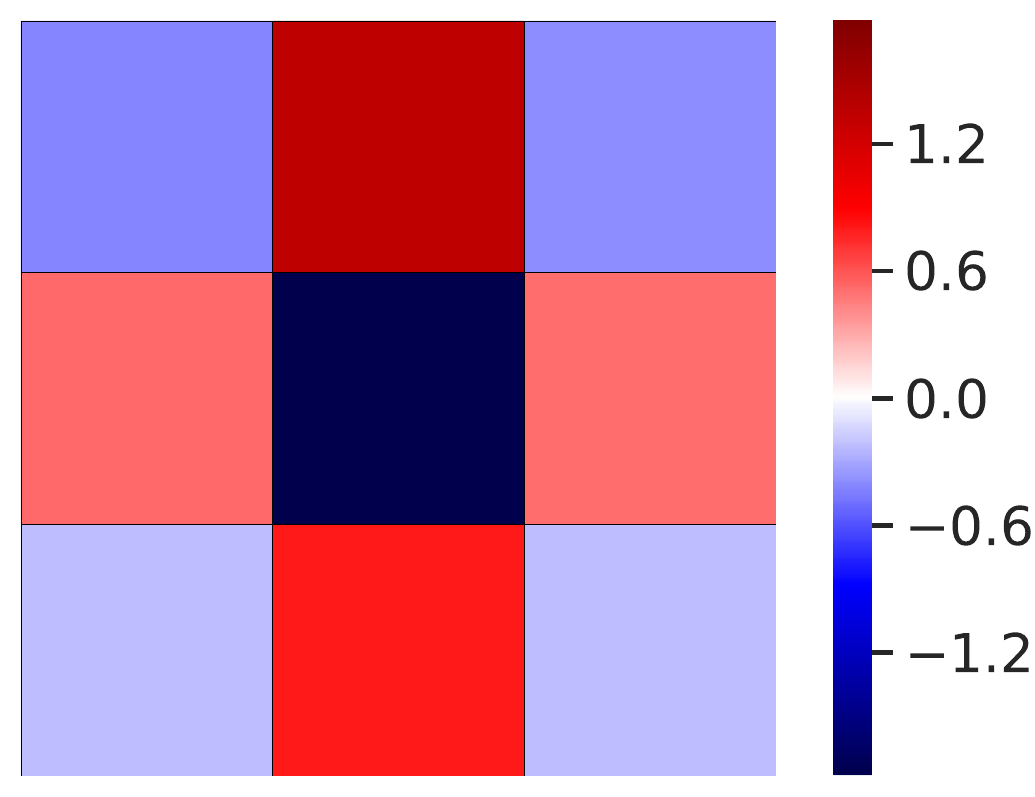} \\
			$w_4$-B &
			\includegraphics[width=0.144\textwidth]{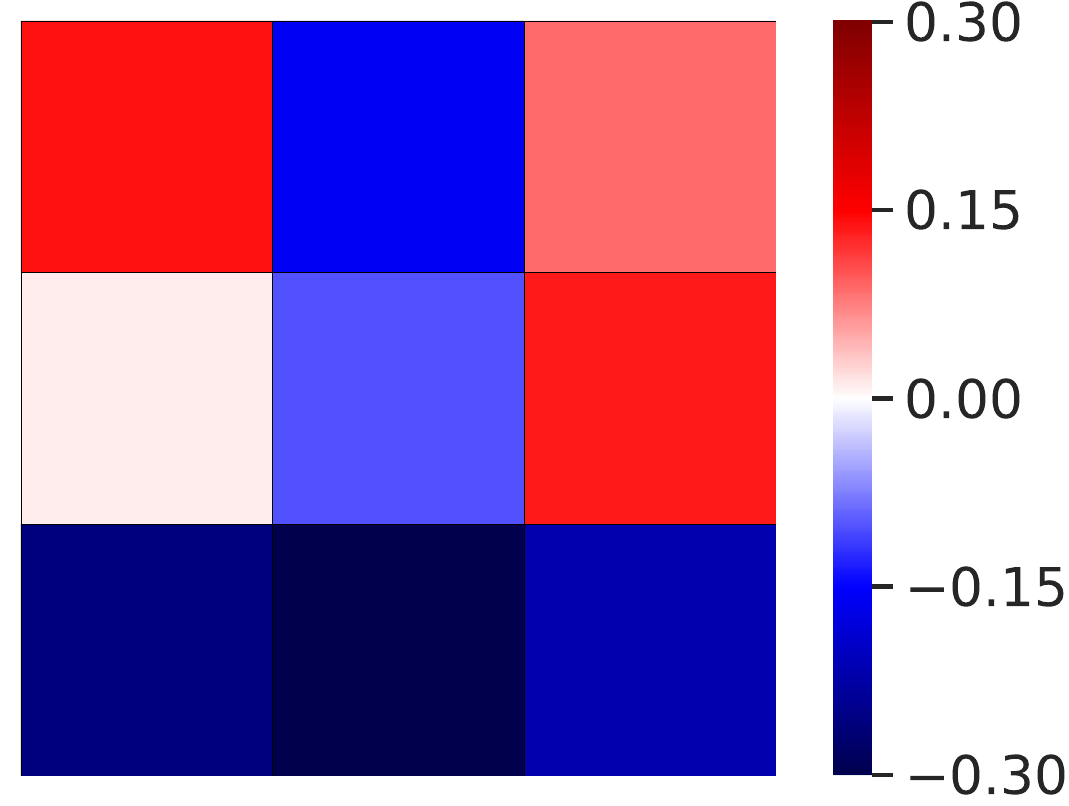} &
			\includegraphics[width=0.144\textwidth]{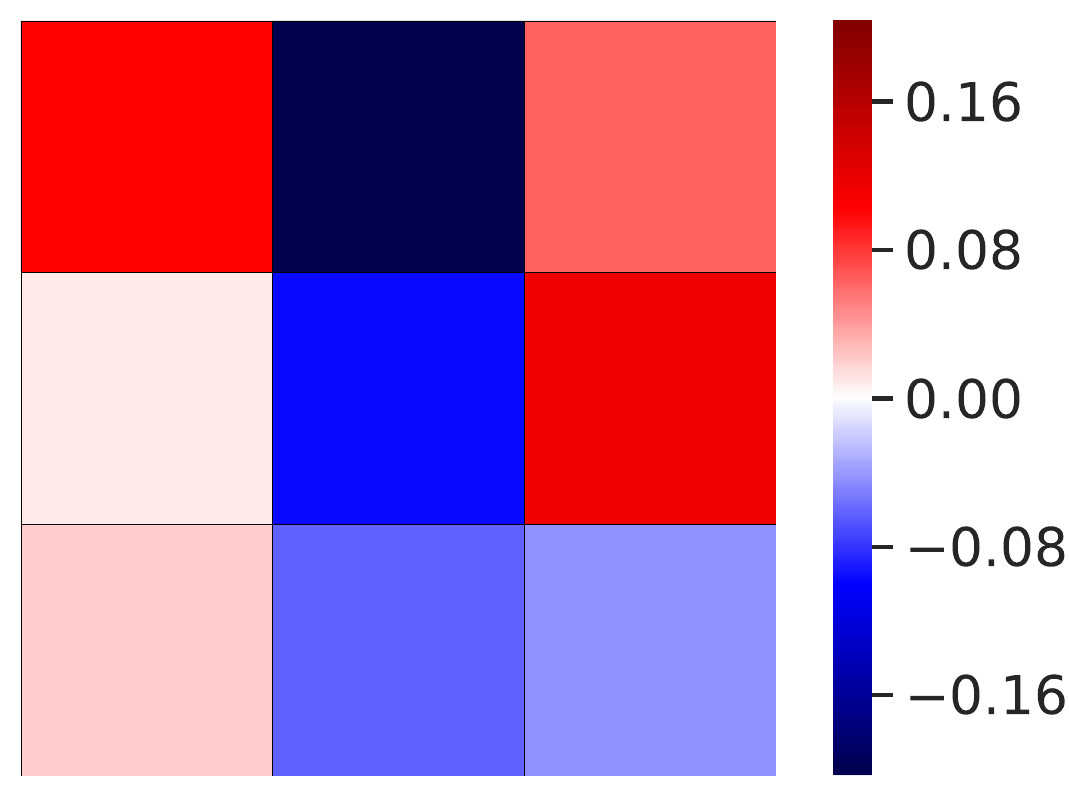} &
			\includegraphics[width=0.144\textwidth]{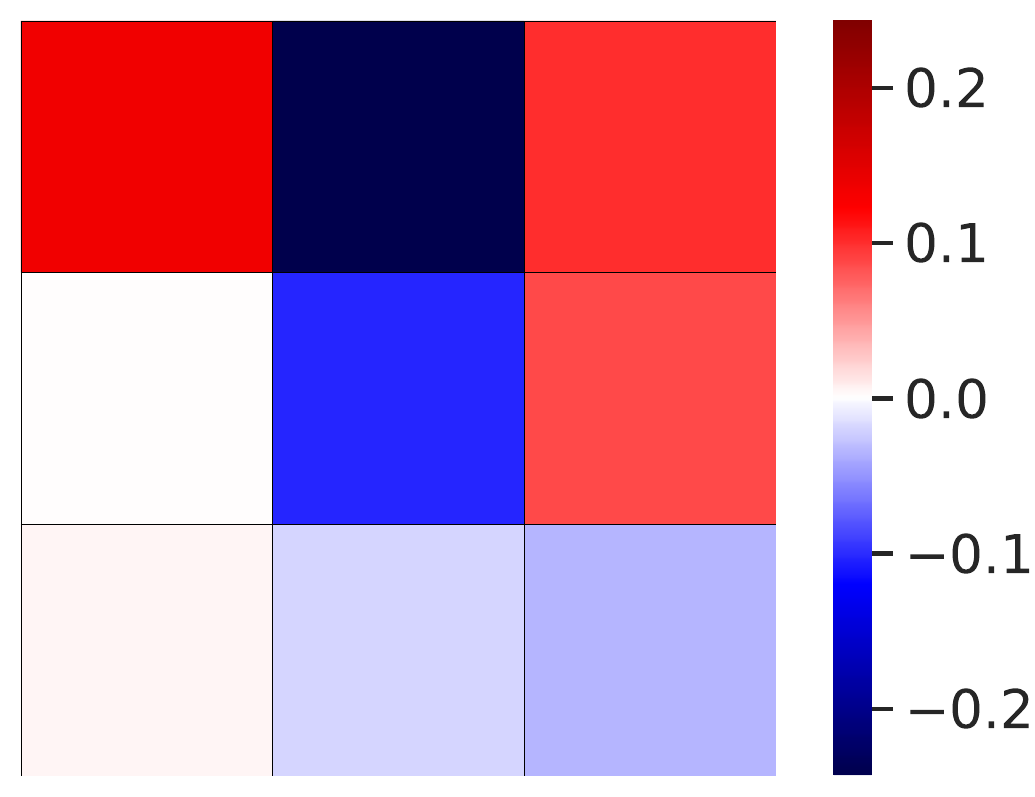} &
			\includegraphics[width=0.144\textwidth]{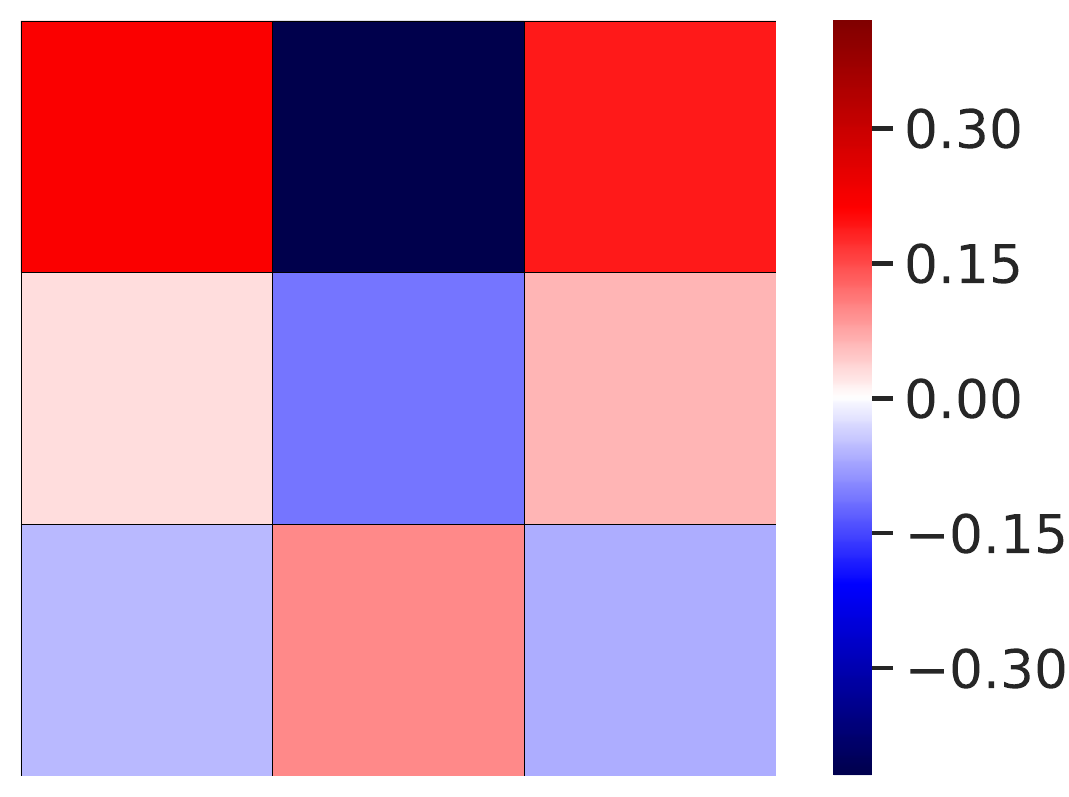} &
			\includegraphics[width=0.144\textwidth]{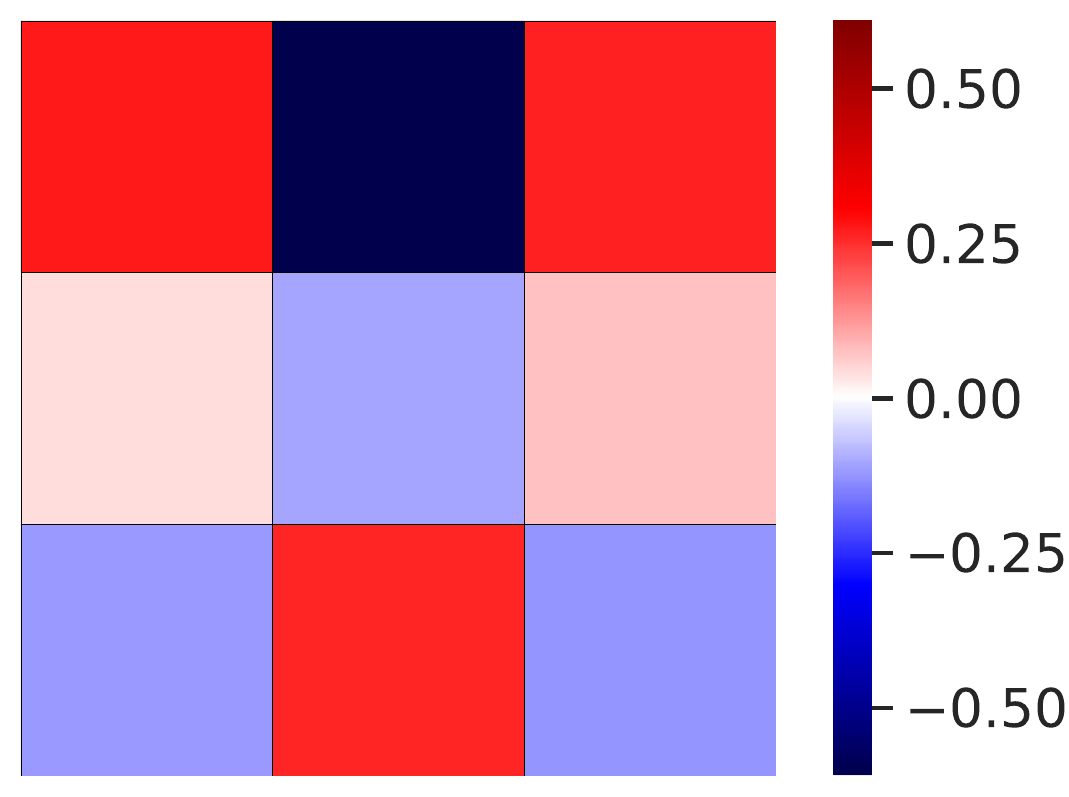} &
			\includegraphics[width=0.144\textwidth]{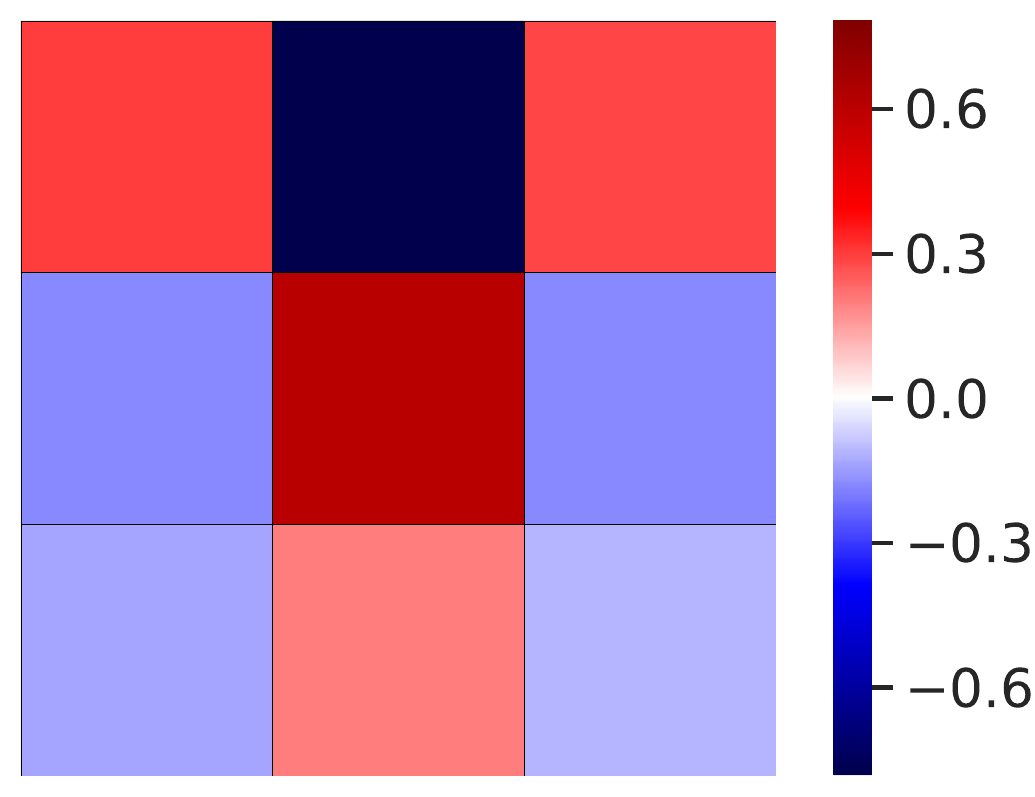}
		\end{tabular}
		\caption{Evolution of the regular filter $w_1$ and filter $w_4$ that leads to catastrophic overfitting. We plot red (R), green (G), and blue (B) channels of the filters. We can observe that in R and G channels, $w_4$ has learned a Laplace filter which is very sensitive to noise. 
		}
		\label{fig:cnn4_filters_all}
	\end{figure}
	In Fig.~\ref{fig:cnn4_filters_all}, we show the evolution of the regular filter $w_1$ and filter $w_4$ that leads to catastrophic overfitting for the three input channels (red, green, blue).
	We can observe that in the red and green channels, $w_4$ has learned a Laplace filter which is very sensitive to noise. 
	Moreover, $w_4$ significantly increases in magnitude after catastrophic overfitting contrary to $w_1$ whose magnitude only decreases (see the colorbar values in Fig.~\ref{fig:cnn4_filters_all} and the plots in Fig.~\ref{fig:cnn4_filters}).

	\myparagraph{Additional feature maps.}
	\begin{figure}[t!]
		\centering
		\includegraphics[width=\textwidth]{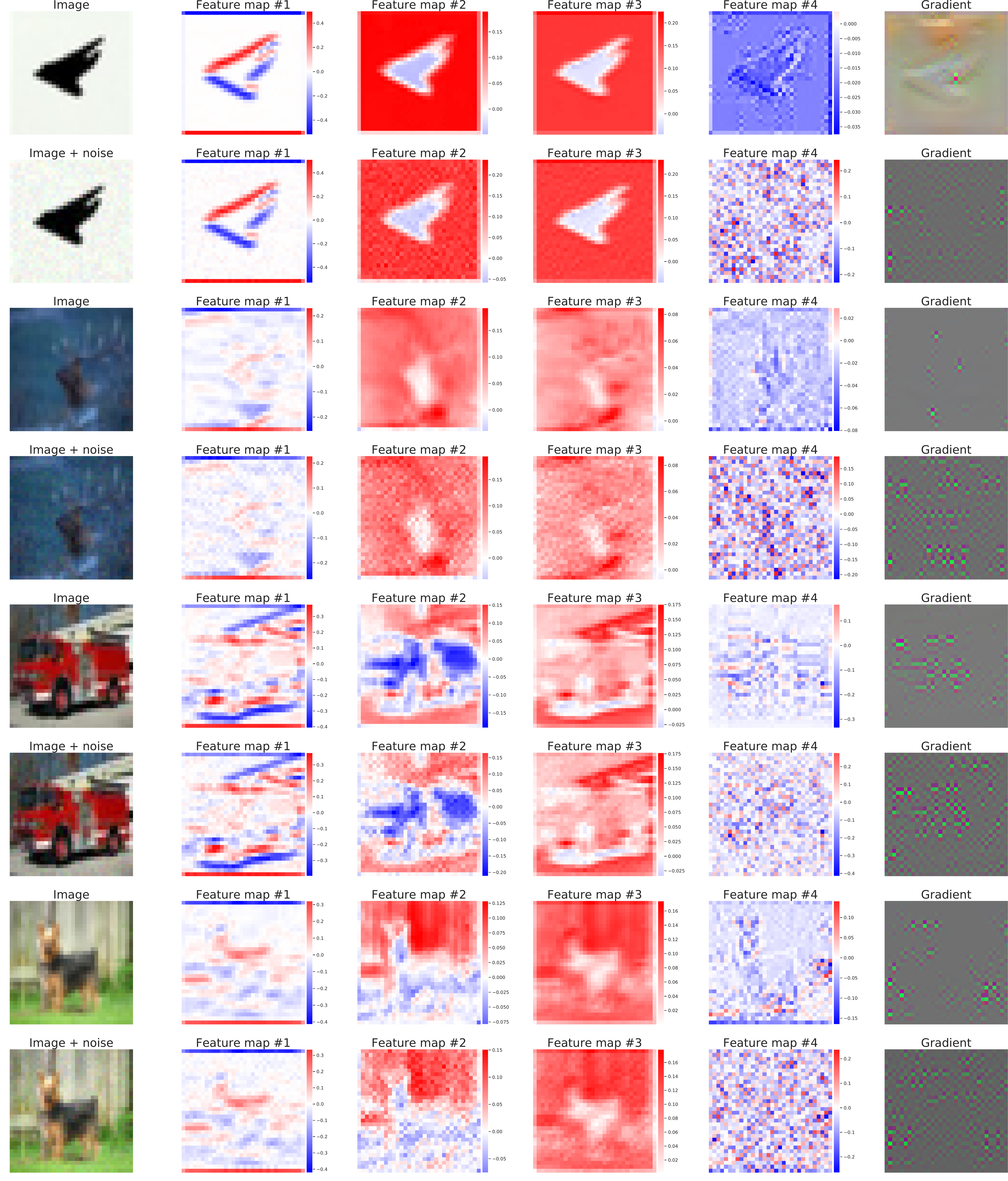}
		\caption{Input images, feature maps, and gradients of the single-layer CNN trained on CIFAR-10 at the end of training (after catastrophic overfitting). \textit{Odd row}: original images. \textit{Even row}: original image plus random noise $\U([-\nicefrac{10}{255}, \nicefrac{10}{255}]^d)$. We observe that only the last filter $w_4$ is highly sensitive to the small uniform noise since the feature maps change dramatically.}
		\label{fig:feature_maps_after_co_standard_cnn4}
	\end{figure}
	In Fig.~\ref{fig:feature_maps_after_co_standard_cnn4}, we show additional feature maps for images with and without uniform random noise $\eta \sim \U([-\nicefrac{10}{255}, \nicefrac{10}{255}]^d)$. These figures complement Fig.~\ref{fig:cnn4_fms} shown in the main part.
	We clearly see that only the last filter $w_4$ is sensitive to the noise since the feature maps change dramatically. At the same time, other filters $w_1$, $w_2$, $w_3$ are only slightly affected by the addition of the noise.
	We also show the input gradients in the last column which illustrate that after adding the noise the gradients change drammatically which leads to small gradient alignment and, in turn, to the failure of FGSM as the solution of the inner maximization problem.

	\clearpage

	\section{Additional experiments for different adversarial training schemes} \label{app:additional_exps}
	In this section, we describe additional experiments related to \texttt{GradAlign} that complement the results shown in Sec.~\ref{sec:main_exps}.

	\subsection{Stronger PGD-2 baseline}\label{sec:add_exps_d1}
	As mentioned in Sec.~\ref{sec:main_exps}, the PGD-2 training baseline that we report outperforms other similar baselines reported in the literature \cite{zhang2019propagate,qin2019adversarial}. Here we elaborate what are likely to be the most important sources of difference. 
	First, we follow the cyclical learning rate schedule of \cite{wong2020fast} which can work as implicit early stopping and thus can help to prevent catastrophic overfitting observed for PGD-2 in \cite{qin2019adversarial}. Another source of difference is that \cite{qin2019adversarial} use the ADAM optimizer while we stick to the standard PGD updates using the sign of the gradient \cite{madry2018towards}. 
	
	The second important factor is a proper step size selection. While \cite{zhang2019propagate} do not observe catastrophic overfitting, their PGD-3 baseline achieves only $32.51\%$ adversarial accuracy compared to the $48.43\%$ for our PGD-2 baseline evaluated with a stronger attack (PGD-50-10 instead of PGD-20-1). One potential explanation for this difference lies in the step size selection, where for PGD-2 we use $\alpha=\nicefrac{\varepsilon}{2}$.
	Related to the step size selection, we also found that using random initialization in PGD (we will refer to as PGD-k-RS) as suggested in \cite{madry2018towards} requires a larger step size $\alpha$. We show the results in Table~\ref{tab:pgd_2_ablation} where we can see that PGD-2-RS AT with $\alpha=\nicefrac{\varepsilon}{2}$ achieves suboptimal robustness compared to $\alpha=\varepsilon$ used for training. However, we consistently observed that PGD-2 AT with $\alpha=\nicefrac{\varepsilon}{2}$ and \textit{no random step} performs best. Thus, we use the latter as our PGD-2 baseline throughout the paper, thus always starting PGD-2 from the original point, without using any random step.
	\begin{table}[h]
		\caption{Robustness of different PGD-2 schemes for $\varepsilon=8/255$ on CIFAR-10 for ResNet-18. The results are averaged over 5 random seeds used for training. 
		}
		\label{tab:pgd_2_ablation}
		\centering
		{\small
			\begin{tabular}{lccc}
				\toprule
				\textbf{Model}                & PGD-2-RS AT, $\alpha=\nicefrac{\varepsilon}{2}$ & PGD-2-RS AT, $\alpha=\varepsilon$ & PGD-2 AT, $\alpha=\nicefrac{\varepsilon}{2}$ \\
				\midrule
				\textbf{PGD-50-10 accuracy}   & 45.06$\pm$0.44\% & 48.07$\pm$0.52\% & 48.43$\pm$0.40\% \\
				\bottomrule
			\end{tabular}
		}
	\end{table}

	\subsection{Results with early stopping}
	\label{app:results_with_es}
	\begin{figure}[b]
		\centering
		\includegraphics[width=0.48\textwidth]{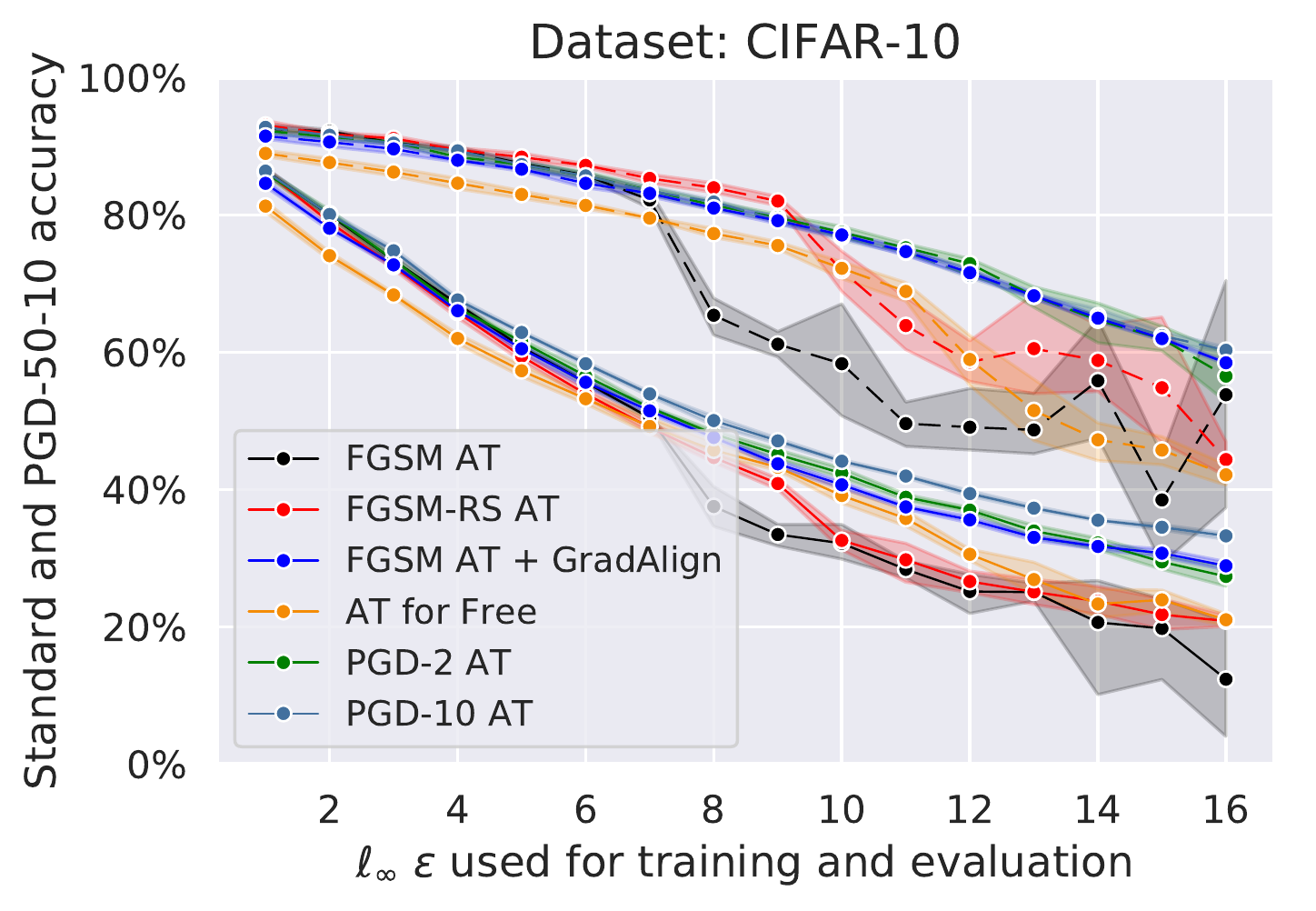} \hspace{2mm}
		\includegraphics[width=0.48\textwidth]{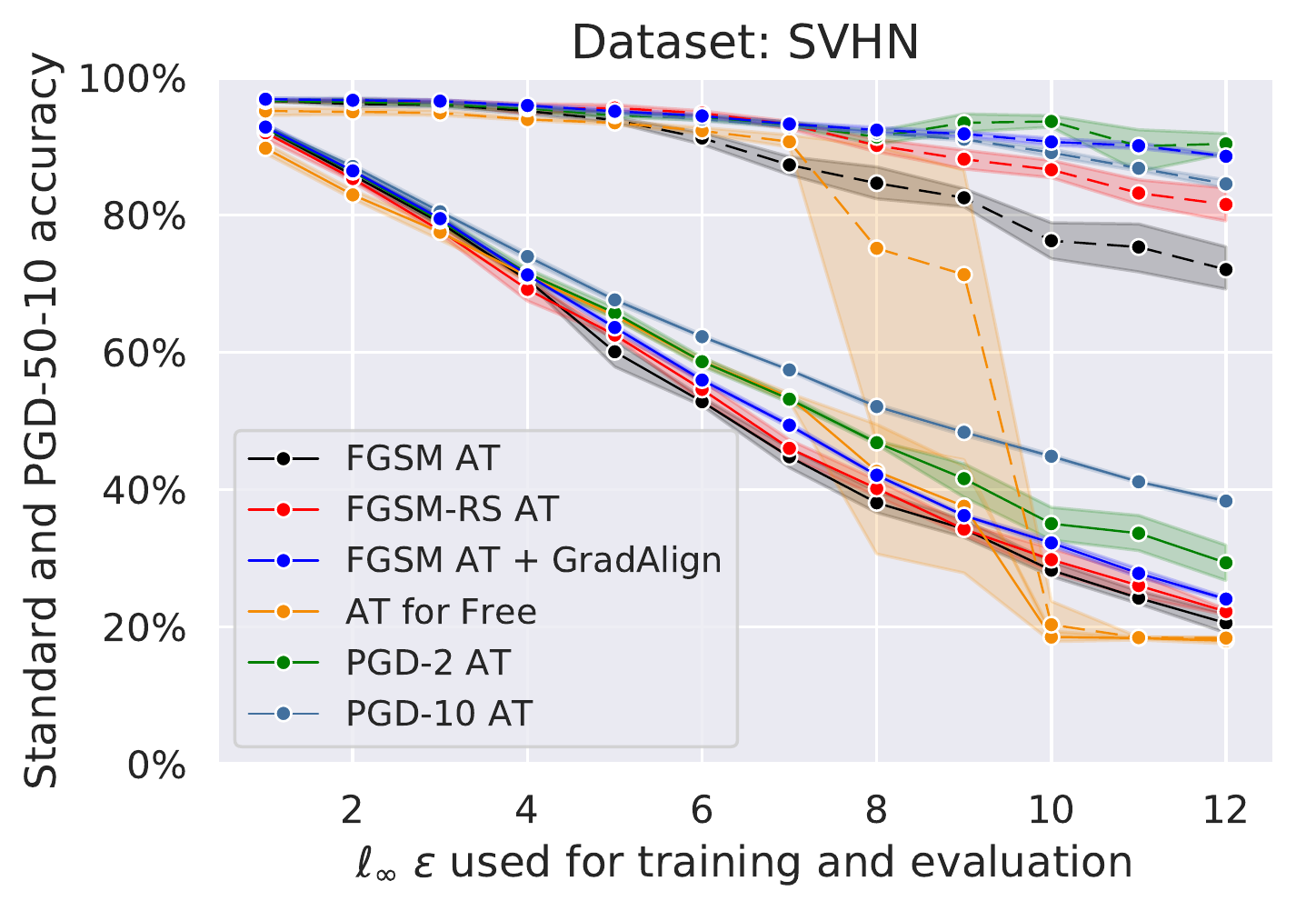}
		\caption{Accuracy (dashed line) and robustness (solid line) of different adversarial training (AT) methods on CIFAR-10 and SVHN with ResNet-18 trained and evaluated with different $l_\infty$-radii. The results are obtained \textbf{with early stopping}, averaged over 5 random seeds used for training and reported with the standard deviation.}
		\label{fig:main_exps_with_es}
	\end{figure}
	We complement the results presented in Fig.~\ref{fig:main_exps} \textit{without early stopping} with the results \textit{with early stopping} which we show in Fig.~\ref{fig:main_exps_with_es}.
	For CIFAR-10, we observe that FGSM~+~\texttt{GradAlign} leads to a good robustness and accuracy outperforming FGSM AT and FGSM-RS AT and performing similarly to PGD-2 and slightly improving for larger $\varepsilon$ close to $\nicefrac{16}{255}$. 
	For SVHN, \texttt{GradAlign} leads to better robustness than other FGSM-based methods.
	We also observe that for large $\varepsilon$ on both CIFAR-10 and SVHN, \textit{AT for Free} performs similarly to FGSM-based methods. Moreover, for $\varepsilon \geq \nicefrac{10}{255}$ on SVHN, \textit{AT for Free} converges to a constant classifier.
	
	On both CIFAR-10 and SVHN, we can see that although early stopping can lead to non-trivial robustness, standard accuracy is often significantly sacrificed which limits the usefulness of this technique. This is in contrast to training with \texttt{GradAlign} which leads to the same standard accuracy as PGD-10 training.

	\subsection{Results for specific $\ell_\infty$-radii}
	\label{app:additional_exps_particular_linf_eps}
	Here we report results from Fig.~\ref{fig:main_exps} for specific $\ell_\infty$-radii which are most often studied in the literature.
	
	\myparagraph{CIFAR-10 results.}
	\begin{table}[b!]
		\caption{
			Robustness and accuracy of different robust training methods on \textbf{CIFAR-10}.
			We report results without early stopping for ResNet-18 unless specified otherwise in parentheses. The results of all the methods reported in Fig.~\ref{fig:main_exps} are shown here with the standard deviation and averaged over 5 random seeds used for training. 
		}
		\label{tab:results_cifar10}
		\centering
		{\small
			\begin{tabular}{lccc}
				\toprule
				\textbf{Model} & \multicolumn{2}{c}{\textbf{Accuracy}} & \textbf{Attack} \\
				& Standard & Adversarial \\
				\midrule
				\multicolumn{4}{c}{$\varepsilon=8/255$} \\
				\hdashline
				Standard                                        & 94.03\% &  0.00\% & PGD-50-10 \\
				\hdashline
				CURE \cite{moosavi2019robustness}               & 81.20\% & 36.30\% & PGD-20-1 \\  
				YOPO-3-5 \cite{zhang2019propagate}              & 82.14\% & 38.18\% & PGD-20-1 \\  
				YOPO-5-3 \cite{zhang2019propagate}              & 83.99\% & 44.72\% & PGD-20-1 \\  
				LLR-2 (Wide-ResNet-28-8) \cite{qin2019adversarial}  & 90.46\% & 44.50\% & MultiTargeted \cite{qin2019adversarial} \\  
				\hdashline
				FGSM                                            & 85.16$\pm$1.3\% & 0.02$\pm$0.04\% & PGD-50-10 \\  
				FGSM-RS                                         & 84.32$\pm$0.08\% & 45.10$\pm$0.56\% & PGD-50-10 \\
				FGSM + \texttt{GradAlign}                       & 81.00$\pm$0.37\% & \textbf{47.58$\pm$0.24\%} & PGD-50-10 \\ 
				\hdashline
				AT for Free ($m=8$)                             &  77.92$\pm$0.65\% & 45.90$\pm$0.98\% & PGD-50-10  \\ 
				PGD-2 ($\alpha=4/255$)                          & 82.15$\pm$0.48\% & 48.43$\pm$0.40\% & PGD-50-10 \\ 
				PGD-2 ($\alpha=4/255$) + \texttt{GradAlign}     & 81.16$\pm$0.39\% & 47.76$\pm$0.77\% & PGD-50-10 \\
				PGD-10 ($\alpha=2\varepsilon/10$)               & 81.88$\pm$0.37\% & \textbf{50.04$\pm$0.79\%} & PGD-50-10 \\  
				\midrule
				
				\multicolumn{4}{c}{$\varepsilon=16/255$} \\
				\hdashline
				FGSM                                                         & 73.76$\pm$7.4\% & 0.00$\pm$0.00\% & PGD-50-10 \\
				FGSM-RS                                                      & 72.18$\pm$3.7\% & 0.00$\pm$0.00\% & PGD-50-10 \\
				FGSM + \texttt{GradAlign}                                    & 58.46$\pm$0.22\% & \textbf{28.88$\pm$0.70\%} & PGD-50-10 \\  
				\hdashline
				AT for Free ($m=8$)                                          & 48.10$\pm$9.83\% & 0.00$\pm$0.00\% & PGD-50-10 \\
				PGD-2 ($\alpha=\varepsilon/2$)                               & 68.65$\pm$5.83\% & \ 9.92$\pm$14.00\% & PGD-50-10 \\
				PGD-2 ($\alpha=\varepsilon/2$) + \texttt{GradAlign}          & 61.38$\pm$0.71\% & 29.80$\pm$0.42\% & PGD-50-10 \\
				PGD-10 ($\alpha=2\varepsilon/10$)                            & 60.28$\pm$0.50\% & \textbf{33.24$\pm$0.52\%} & PGD-50-10 \\   
				\bottomrule
			\end{tabular}
		}
	\end{table}
	We report robustness and accuracy in Table~\ref{tab:results_cifar10} for CIFAR-10 without using early stopping where we can clearly see which methods lead to catastrophic overfitting and thus suboptimal robustness. We compare the same methods as in Fig.~\ref{fig:main_exps}, and additionally we report the results for $\varepsilon=\nicefrac{8}{}255$ of the CURE \cite{moosavi2019robustness}, YOPO \cite{zhang2019propagate}, and LLR \cite{qin2019adversarial} approaches.
	First, for $\varepsilon=\nicefrac{8}{255}$, we see that FGSM~+~\texttt{GradAlign} outperforms \textit{AT for Free} and all methods that use FGSM training.
	Then, we also observe that the model trained with CURE \cite{moosavi2019robustness} leads to robustness that is suboptimal compared to FGSM-RS AT evaluated with a stronger attack: $36.3\%$ vs $45.1\%$.
	YOPO-3-5 and YOPO-5-3 \cite{zhang2019propagate} require 3 and 5 full steps of PGD respectively, thus they are much more expensive than FGSM-RS AT, and, however, they lead to worse adversarial accuracy: $38.18\%$ and $44.72\%$ vs $45.10\%$. \citet{qin2019adversarial} report that LLR-2, i.e. their approach with 2 steps of PGD, achieves $44.50\%$ adversarial accuracy with MultiTargeted attack \cite{gowal2019alternative} and $46.47\%$ with their untargeted PGD attack which uses a different loss function compared to our PGD attack. These two evaluations are not directly comparable to other results in Table~\ref{tab:results_cifar10} since the attacks are different and moreover they use a larger network (Wide-ResNet-28-8) which usually leads to better results \cite{madry2018towards}. 
	However, we think that the gap of $3-4\%$ adversarial accuracy of MultiTargeted evaluation compared to that of our reported FGSM~+~\texttt{GradAlign} and PGD-2 methods ($47.58\%$ and $48.43\%$ resp.) is still significant since the difference between MultiTargeted and a PGD attack with random restarts is observed to be small (e.g. around 1\% between MultiTargeted and PGD-20-10 on the CIFAR-10 challenge of~\cite{madry2018towards}). 
	
	For $\varepsilon=\nicefrac{16}{255}$, none of the one-step methods work without early stopping except FGSM~+~\texttt{GradAlign}.
	We also evaluate PGD-2~+~\texttt{GradAlign} and conclude that the benefit of combining the two comes when PGD-2 alone leads to catastrophic overfitting which occurs at $\varepsilon=\nicefrac{16}{255}$. For $\varepsilon=\nicefrac{8}{255}$, there is no benefit of combining the two approaches. This is consistent with our observation regarding catastrophic overfitting for FGSM (e.g. see Fig.~\ref{fig:main_exps} for small $\varepsilon$): if there is no catastrophic overfitting, there is no benefit of adding \texttt{GradAlign} to FGSM training.
	
	To further ensure that FGSM~+~\texttt{GradAlign} models do not benefit from gradient masking~\cite{papernot2017practical}, we additionally compare the robustness of FGSM~+~\texttt{GradAlign} and FGSM-RS models obtained via \textit{AutoAttack}~\cite{croce2020reliable}. We observe that \textit{AutoAttack} proportionally reduces the adversarial accuracy of both models: for $\varepsilon=\nicefrac{8}{255}$, FGSM~+~\texttt{GradAlign} achieves 44.54$\pm$0.24\% adversarial accuracy while FGSM-RS achieves 42.80$\pm$0.58\%. This is consistent with the evaluation results of~\cite{croce2020reliable} where they show that \textit{AutoAttack} reduces adversarial accuracy for many models by 2\%-3\% for $\varepsilon=\nicefrac{8}{255}$ compared to the originally reported results based on the standard PGD attack (see Table~2 in~\cite{croce2020reliable}). The same tendency is observed also for higher $\varepsilon$, e.g. for $\varepsilon=\nicefrac{16}{255}$ FGSM~+~\texttt{GradAlign} achieves 20.56$\pm$0.36\% adversarial accuracy when evaluated with \textit{AutoAttack}.

	\myparagraph{SVHN results.}
	\begin{table}[t!]
		\caption{
			Robustness and accuracy of different robust training methods on \textbf{SVHN}.
			We report results without early stopping for ResNet-18.
			All the results are reported with the standard deviation and averaged over 5 random seeds used for training.
		}
		\label{tab:results_svhn}
		\centering
		{\small
			\begin{tabular}{lcc}
				\toprule
				\textbf{Model} & \multicolumn{2}{c}{\textbf{Accuracy}}\\
				& Standard & PGD-50-10 \\
				\midrule
				\multicolumn{3}{c}{$\varepsilon=8/255$} \\
				\hdashline
				Standard                                        & 96.00\% & 1.00\% \\
				\hdashline
				FGSM                                            & 91.40$\pm$1.64\% & 0.04$\pm$0.05\% \\
				FGSM-RS                                         & 95.38$\pm$0.27\% & 0.00$\pm$0.00\% \\
				FGSM + \texttt{GradAlign}                       & 92.36$\pm$0.47\% & \textbf{42.08$\pm$0.25\%} \\
				\hdashline
				AT for Free ($m=8$)                             & 75.34$\pm$28.4\% & 43.16$\pm$12.3\% \\
				PGD-2 ($\alpha=\varepsilon/2$)                  & 92.68$\pm$0.45\% & 47.28$\pm$0.26\% \\
				PGD-2 + \texttt{GradAlign} ($\alpha=\varepsilon/2$)  & 92.46$\pm$0.35\% & 47.02$\pm$0.83\% \\
				PGD-10 ($\alpha=2\varepsilon/10$)               & 91.92$\pm$0.40\% & \textbf{52.08$\pm$0.49\%} \\       
				\midrule
				
				\multicolumn{3}{c}{$\varepsilon=12/255$} \\
				\hdashline
				FGSM                                            & 88.74$\pm$1.25\% & 0.00$\pm$0.00\% \\
				FGSM-RS                                         & 94.70$\pm$0.66\% & 0.00$\pm$0.00\% \\
				FGSM + \texttt{GradAlign}                       & 88.54$\pm$0.21\% & \textbf{24.04$\pm$0.31\%} \\
				\hdashline
				AT for Free ($m=8$)                             & 18.50$\pm$0.00\% & 18.50$\pm$0.00\% \\
				PGD-2 ($\alpha=\varepsilon/2$)                  & 92.74$\pm$2.26\% & 14.30$\pm$13.34\% \\
				PGD-2 + \texttt{GradAlign} ($\alpha=\varepsilon/2$)  & 87.14$\pm$0.26\% & 31.26$\pm$0.24\% \\
				PGD-10 ($\alpha=2\varepsilon/10$)                    & 84.52$\pm$0.63\% & \textbf{38.32$\pm$0.38\%} \\   
				\bottomrule
			\end{tabular}
		}
	\end{table}
	\begin{table}[t!]
		\caption{
			Robustness and accuracy of different robust training methods on \textbf{ImageNet}. We report results without early stopping for ResNet-50.
		}
		\label{tab:results_imagenet}
		\centering
		{\small
			\begin{tabular}{lccc}
				\toprule
				\textbf{Model} & \textbf{$\ell_\infty$-radius} & \textbf{Standard accuracy} & \textbf{PGD-50-10 accuracy} \\
				\midrule
				FGSM    & 2/255 & 61.7\% & 42.1\% \\
				FGSM-RS & 2/255 & 59.3\% & 41.1\% \\
				FGSM + \texttt{GradAlign} & 2/255 & 61.8\% & 41.4\% \\ 
				\hdashline
				FGSM    & 4/255 & 56.9\% & 30.6\% \\
				FGSM-RS & 4/255 & 55.3\% & 27.8\% \\
				FGSM + \texttt{GradAlign} & 4/255 & 57.8\% & 30.5\% \\ 
				\hdashline
				FGSM    & 6/255 & 51.5\% & 20.6\% \\
				FGSM-RS & 6/255 & 36.6\% &  0.1\% \\
				FGSM + \texttt{GradAlign} & 6/255 & 51.5\% & 20.3\% \\ 
				\bottomrule
			\end{tabular}
		}
	\end{table}
	We report robustness and accuracy in Table~\ref{tab:results_svhn} for SVHN without using early stopping. We can see that for both $\varepsilon=\nicefrac{8}{255}$ and $\varepsilon=\nicefrac{16}{255}$, \texttt{GradAlign} successfully prevents catastrophic overfitting in contrast to FGSM and FGSM-RS, although there is still a $5\%$ gap to PGD-2 training for $\varepsilon=\nicefrac{8}{255}$. \textit{AT for free} performs slightly better than FGSM~+~\texttt{GradAlign} for $\varepsilon=\nicefrac{8}{255}$, but it already starts to show a high variance in the robustness and accuracy depending on the random seed. For $\varepsilon=\nicefrac{12}{255}$, all the 5 models of \textit{AT for free} converge to a constant classifier.
	
	Combining PGD-2 with \texttt{GradAlign} does not lead to improved results for $\varepsilon=\nicefrac{8}{255}$ since there is no catastrophic overfitting for PGD-2. 
	However, for $\varepsilon=\nicefrac{12}{255}$, we can clearly see that PGD-2~+~\texttt{GradAlign} leads to better results than PGD-2 achieving $31.26\pm0.24\%$ instead of $14.30\pm13.34\%$ adversarial accuracy.

	\myparagraph{ImageNet results.}
	We also perform similar experiments on ImageNet in Table~\ref{tab:results_imagenet}. We observe that even for standard FGSM training, catastrophic overfitting \textit{does not} occur for $\varepsilon \in \{\nicefrac{2}{255}, \nicefrac{4}{255}\}$ considered in \cite{shafahi2019adversarial,wong2020fast}, and thus there is no additional benefit from using \texttt{GradAlign} since its main role is to prevent catastrophic overfitting. We report the results of FGSM~+~\texttt{GradAlign} for completeness to show that \texttt{GradAlign} can be applied on the ImageNet scale, although it leads to approximately $3\times$ slowdown on ImageNet compared to standard FGSM training.

	For $\varepsilon = \nicefrac{6}{255}$, we observe that catastrophic overfitting occurs for FGSM-RS very early in training (around epoch 3), but not for FGSM or FGSM~+~\texttt{GradAlign} training. This contradicts our observations on CIFAR-10 and SVHN where we observed that FGSM-RS usually helps to postpone catastrophic overfitting to higher $\varepsilon$. However, it is computationally demanding to replicate the results on ImageNet multiple times over different random seeds as we did for CIFAR-10 and SVHN. Thus, we leave a more detailed investigation of catastrophic overfitting on ImageNet for future work.

	\subsection{Ablation studies}
	\label{app:ablation_studies}
	In this section, we aim to provide more details about sensitivity of \texttt{GradAlign} to its hyperparameter $\lambda$, the total number of training epochs, and also discuss training with \texttt{GradAlign} for very high $\varepsilon$ values. 
	
	\myparagraph{Ablation study for GradAlign $\lambda$.}
	\begin{figure}[b!]
		\centering
		\begin{minipage}{0.48\textwidth}
			\includegraphics[width=1.0\textwidth]{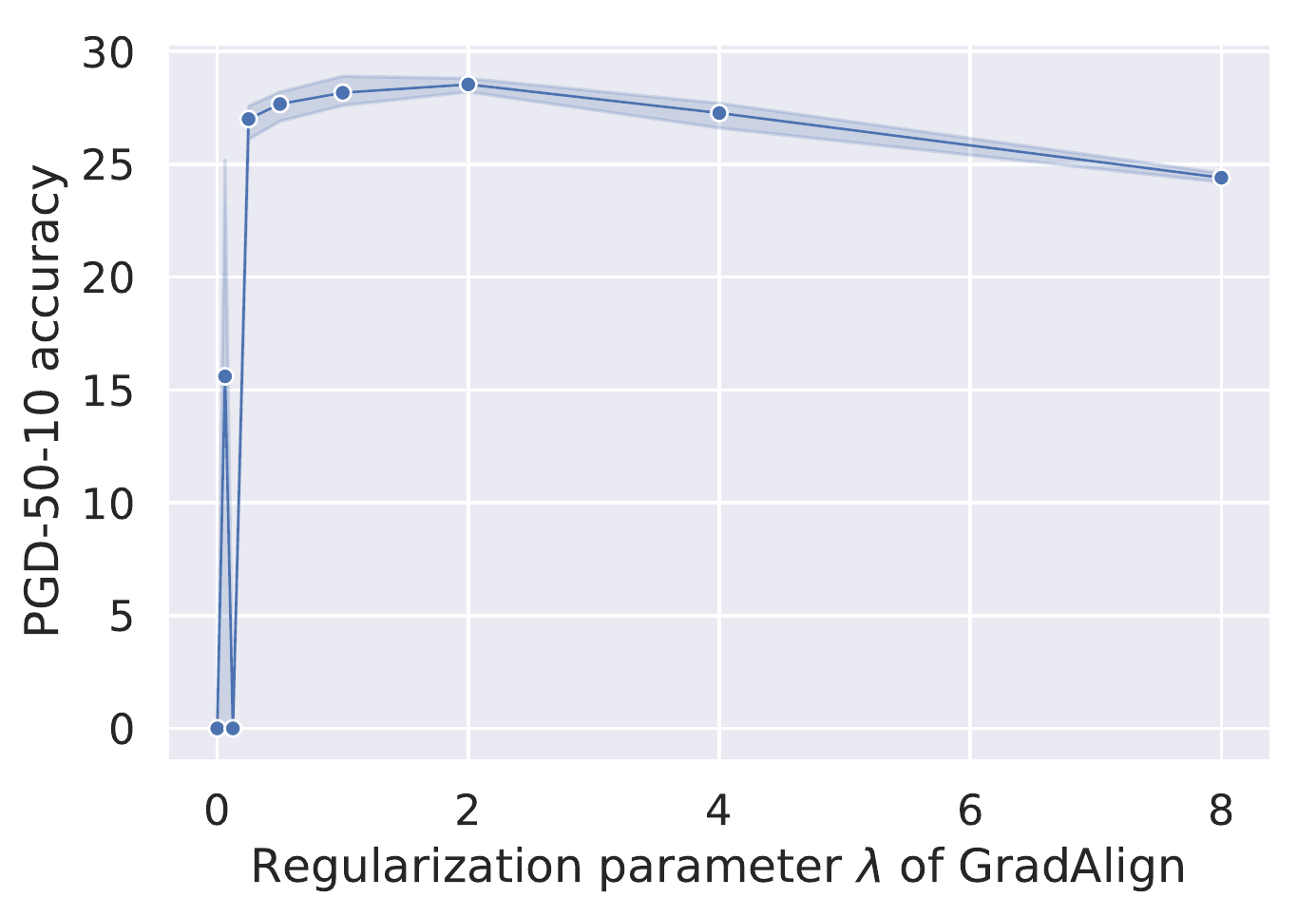} 
			\caption{Ablation study for the regularization parameter $\lambda$ for FGSM~+~\texttt{GradAlign} under $\varepsilon=\nicefrac{16}{255}$ without early stopping. We train ResNet-18 models on CIFAR-10. The results are averaged over 3 random seeds used for training and reported with the standard deviation.}
			\label{fig:ga_ablation_lambda}
		\end{minipage}
		\hspace{1mm}
		\begin{minipage}{0.48\textwidth}
			\includegraphics[width=1.0\textwidth]{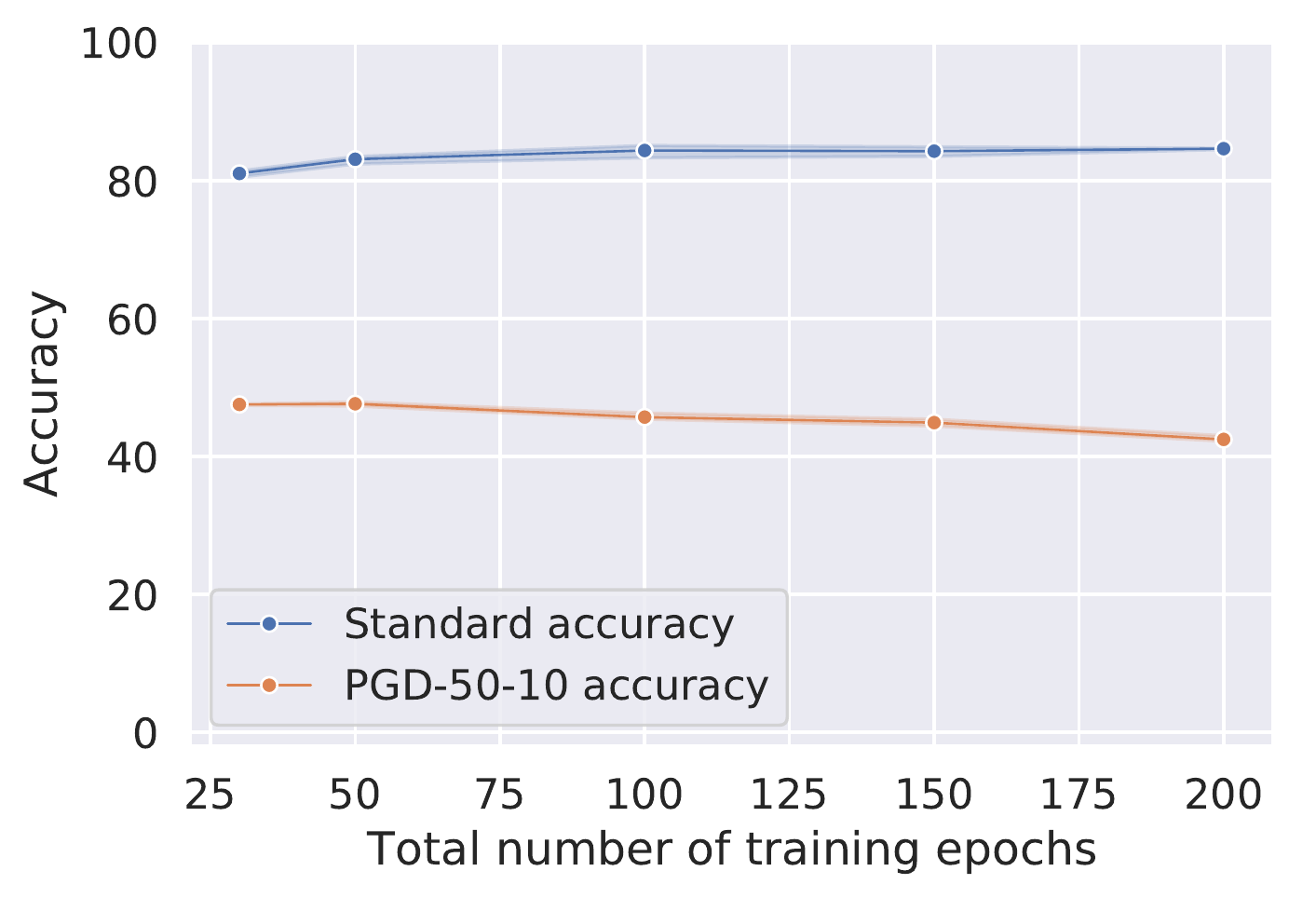}
			\caption{Ablation study for the total number of training epochs for FGSM~+~\texttt{GradAlign} under $\varepsilon=\nicefrac{8}{255}$ without early stopping. 
				We train ResNet-18 models on CIFAR-10.
				The results are averaged over 3 random seeds used for training and reported with the standard deviation.}
			\label{fig:ga_ablation_epochs}
		\end{minipage}
	\end{figure}
	We provide an ablation study for the regularization parameter $\lambda$ of \texttt{GradAlign} in Fig.~\ref{fig:ga_ablation_lambda}, where we plot the adversarial accuracy of ResNet-18 trained using FGSM~+~\texttt{GradAlign} with $\varepsilon=\nicefrac{16}{255}$ on CIFAR-10. 
	First, we observe that for small $\lambda$ catastrophic overfitting occurs so that the average PGD-50-10 accuracy is either $0\%$ or greater than $0\%$ but has a high standard deviation since only some runs are successful while other runs fail because of catastrophic overfitting. We observe that the best performance is achieved for $\lambda=2$ where catastrophic overfitting does not occur and the final adversarial accuracy is very concentrated. 
	For larger $\lambda$ values we observe a slow decrease in the adversarial accuracy since the model becomes overregularized. 
	We note that the range of $\lambda$ values which have close to the best performance ($\geq26\%$ adversarial accuracy) ranges in $[0.25, 4]$, thus we conclude that \texttt{GradAlign} is robust to the exact choice of $\lambda$. This is also confirmed by our hyperparameter selection method for Fig.~\ref{fig:main_exps}, where we performed a linear interpolation on the logarithmic scale between successful $\lambda$ values for $\varepsilon=\nicefrac{8}{255}$ and $\varepsilon=\nicefrac{16}{255}$. Even such a coarse hyperparameter selection method, could ensure that none of the FGSM~+~\texttt{GradAlign} runs reported in Fig.~\ref{fig:ga_ablation_lambda} suffered from catastrophic overfitting.

	\myparagraph{Ablation study for the total number of training epochs.}
	Recently, \citet{rice2020overfitting} brought up the importance of early stopping in adversarial training. They identify the phenomenon called \textit{robust overfitting} when training longer hurts the adversarial accuracy on the test set. Thus, we check here whether training with \texttt{GradAlign} has some influence on robust overfitting. We note that the authors of \cite{rice2020overfitting} suggest that robust and catastrophic overfitting phenomena are distinct since robust overfitting implies a gap between training and test set robustness, while catastrophic overfitting implies low robustness on \textit{both} training and test sets.
	To explore this for FGSM~+~\texttt{GradAlign}, in Fig.~\ref{fig:ga_ablation_epochs} we show the final clean and adversarial accuracies for five different models trained with $\{30, 50, 100, 150, 250\}$ epochs. We observe the same trend as \cite{rice2020overfitting} report: training longer slightly degrades adversarial accuracy (while in our case also the clean accuracy slightly improves). Thus, this experiment also suggests that robust overfitting is not directly connected to catastrophic overfitting and has to be addressed separately. Finally, we note based on Fig.~\ref{fig:ga_ablation_epochs} that when we use FGSM in combination with \texttt{GradAlign}, even training \textit{up to 200 epochs} does not lead to catastrophic overfitting.

	\myparagraph{Ablation study for very high $\varepsilon$.}
	Here we make an additional test on whether \texttt{GradAlign} prevents catastrophic overfitting for very high $\varepsilon$ values. In Fig.~\ref{fig:main_exps} and Fig.~\ref{fig:main_exps_with_es} we showed results for $\varepsilon \leq 16$ for CIFAR-10 and for $\varepsilon \leq 12$ on SVHN. For SVHN, FGSM~+~\texttt{GradAlign} achieves 24.04$\pm$0.31\% adversarial accuracy which is already close to that of a majority classifier (18.50\%). The effect of increasing the perturbations size $\varepsilon$ on SVHN even further just leads to learning a constant classifier. However, on CIFAR-10 for $\varepsilon = 16$, FGSM~+~\texttt{GradAlign} achieves 28.88$\pm$0.70\% adversarial accuracy which is sufficiently far from that of a majority classifier (10.00\%). Thus, a natural question is whether catastrophic overfitting still occurs for \texttt{GradAlign} on CIFAR-10, but just for higher $\varepsilon$ values than what we considered in the main part of the paper. To show that it is not the case, in Table~\ref{tab:ga_ablation_high_eps} we show the results of FGSM~+~\texttt{GradAlign} trained with $\varepsilon \in \{\nicefrac{24}{255}, \nicefrac{32}{255}\}$ (we use $\lambda=2.0$ and the maximum learning rate $0.1$). We observe no signs of catastrophic overfitting \textit{even for very high} $\varepsilon$ such as $\nicefrac{32}{255}$. Note that in this case the standard accuracy is very low (23.07$\pm$3.35\%), thus considering such large perturbations is not practically interesting, but it rather serves as a sanity check that our method does not suffer from catastrophic overfitting even for very high $\varepsilon$.
	\begin{table}[h]
		\caption{
			Robustness and accuracy of FGSM~+~\texttt{GradAlign} for very high $\varepsilon$ on CIFAR-10 without early stopping for ResNet-18. We report results with the standard deviation and averaged over 3 random seeds used for training. We observe no catastrophic overfitting even for very high $\varepsilon$.
		}
		\label{tab:ga_ablation_high_eps}
		\centering
		{\small
			\begin{tabular}{ccc}
				\toprule
				\textbf{$\ell_\infty$-radius} & \textbf{Standard accuracy} & \textbf{PGD-50-10 accuracy} \\
				\midrule
				24/255 & 41.80$\pm$0.36\% & 17.07$\pm$0.90\% \\
				32/255 & 23.07$\pm$3.35\% & 12.93$\pm$1.44\% \\
				\bottomrule
			\end{tabular}
		}
	\end{table}

	\subsection{Comparison of GradAlign to gradient-based penalties}
	\label{app:alternatives_to_gradalign}
	In this section, we compare \texttt{GradAlign} to other alternatives: $\ell_2$ gradient norm penalization and CURE \cite{moosavi2019robustness}. The motivation to study them comes from the fact that after catastrophic overfitting, the input gradients change dramatically inside the $\ell_\infty$-balls around input points, and thus other gradient-based regularizers may also be able to improve the stability of the input gradients and thus prevent catastrophic overfitting.
	
	In Table~\ref{tab:other_grad_penalties}, we present results of FGSM training with other gradient-based penalties studied in the literature: 
	\begin{itemize}
		\item $\ell_2$ gradient norm regularization \cite{ross2018improving,simon2019first}: $\lambda \norm{\nabla_x \ell(x, y; \theta)}_2^2$,
		\item curvature regularization (CURE) \cite{moosavi2019robustness}: $\lambda  \norm{\nabla_x \ell(x+\delta_{FGSM}, y; \theta) - \nabla_x \ell(x, y; \theta)}_2^2$.
	\end{itemize} 
	First of all, we note that the originally proposed approaches \cite{ross2018improving,simon2019first,moosavi2019robustness} \textit{do not} involve adversarial training and rely \textit{only} on these gradient penalties to achieve some degree of robustness. In contrast, we \textit{combine} the gradient penalties with FGSM training to see whether they can prevent catastrophic overfitting similarly to \texttt{GradAlign}. 
	For the gradient norm penalty, we use the regularization parameters $\lambda \in \{1{,}000, 2{,}000\}$ for $\varepsilon \in \{\nicefrac{8}{255}, \nicefrac{16}{255}\}$ respectively. For CURE, we use $\lambda \in \{700, 20{,}000\}$ for $\varepsilon \in \{\nicefrac{8}{255}, \nicefrac{16}{255}\}$ respectively. In both cases, we found the optimal hyperparameters using a grid search over $\lambda$.
	We can see that for $\varepsilon=\nicefrac{8}{255}$ all three approaches successfully prevent catastrophic overfitting, although the final robustness slightly varies between $46.69\%$ for FGSM with the $\ell_2$-gradient penalty and $47.58\%$ for FGSM with \texttt{GradAlign}. 
	
	For $\varepsilon=\nicefrac{16}{255}$, both FGSM~+~CURE and FGSM~+~\texttt{GradAlign} prevent catastrophic overfitting leading to very concentrated results with a small standard deviation (0.29\% and 0.70\% respectively). However, the average adversarial accuracy is better for FGSM~+~\texttt{GradAlign}: $28.88\%$ vs $25.38\%$.
	At the same time, FGSM with the $\ell_2$-gradient penalty leads to unstable final performance: the adversarial accuracy has a high standard deviation: $13.64\pm11.2\%$.
	
	We think that the main difference in the performance of \texttt{GradAlign} compared to the gradient penalties that we considered comes from the fact that it is invariant to the gradient norm, and it takes into account only the directions of two gradients inside the $\ell_\infty$-ball around the given input.
	
	Inspired by CURE, we also tried two additional experiments:
	\begin{enumerate}
		\item Using the FGSM point $\delta_{FGSM}$ for the gradient taken at the second input point for \texttt{GradAlign}, but we observed that it does not make a substantial difference, i.e. this version of \texttt{GradAlign} also prevents catastrophic overfitting and leads to similar results. 
		However, if we use CURE without FGSM in the cross-entropy loss, then we observe a benefit of using $\delta_{FGSM}$ in the regularizer which is consistent with the observations made in \citet{moosavi2019robustness}.
		\item Using \texttt{GradAlign} without FGSM in the cross-entropy loss. In this case, we observed that the model did not significantly improve its robustness suggesting that \texttt{GradAlign} \textit{is not} a sufficient regularizer on its own to promote robustness and has to be used \textit{with} some adversarial training method.
	\end{enumerate}
	
	We think that an interesting future direction is to explore how one can speed up \texttt{GradAlign} or to come up with other regularization methods that are also able to prevent catastrophic overfitting, but avoid relying on the input gradients which lead to a slowdown in training. We think that some potential strategies to speed up \texttt{GradAlign} can include parallelization of the computations or saving some computations by subsampling the training batches for the regularizer. We postpone a further exploration of these ideas to future work.

	\begin{table}[t]
		\caption{Additional comparison of FGSM AT with \texttt{GradAlign} to FGSM AT with other gradient penalties on CIFAR-10. 
			We report results without early stopping for ResNet-18.
			All the results are reported with the standard deviation and averaged over 5 random seeds used for training.
		}
		\label{tab:other_grad_penalties}
		\centering
		{\small
			\begin{tabular}{lccc}
				\toprule
				\textbf{Model} & \multicolumn{3}{c}{\textbf{Accuracy}}\\
				& Standard & PGD-50-10 \\
				\midrule
				\multicolumn{4}{c}{$\varepsilon=8/255$} \\
				\hdashline
				FGSM + $\norm{\nabla_x}_2^2$                                               & 77.47$\pm$0.14\% & 46.69$\pm$1.27\% & \\
				FGSM + CURE                                                                 & 80.20$\pm$0.29\% & 47.25$\pm$0.21\% & \\
				FGSM + \texttt{GradAlign}                                                   & 81.00$\pm$0.37\% & \textbf{47.58$\pm$0.24\%}  \\
				\midrule
				
				\multicolumn{4}{c}{$\varepsilon=16/255$} \\
				\hdashline
				FGSM + $\norm{\nabla_x}_2^2$                                            & 56.44$\pm$2.22\% & 13.64$\pm$11.2\% \\  
				FGSM + CURE                                                             & 62.39$\pm$0.42\% & 25.38$\pm$0.29\% \\ 
				FGSM + \texttt{GradAlign}                                               & 58.46$\pm$0.22\% & \textbf{28.88$\pm$0.70\%} & \\
				\bottomrule
			\end{tabular}
		}
	\end{table}

\end{document}